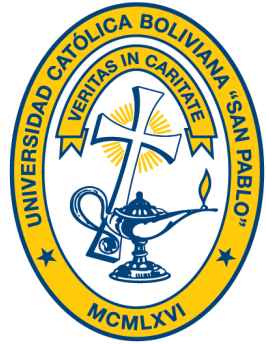

# UNIVERSIDAD
# CATÓLICA
## B O L I V I A N A
### S A N T A   C R U Z

INGENIERÍA MECATRÓNICA

**Luis José Alarcón Aneiva**

**DESARROLLO DE UN PROTOTIPO DE SENSOR DE FUERZA NEUMÁTICO**

Tesis de Grado presentado para optar al título de Licenciado en Ingeniería Mecatrónica.

Tutor: Daniel Raúl Asturizaga Hurtado de Mendoza.

Septiembre, 2020

**Luis José Alarcón Aneiva**

**DESARROLLO DE UN PROTOTIPO DE SENSOR DE FUERZA NEUMÁTICO**

La presente Tesis de Grado fue evaluada y aprobada por una banca de jurados compuesta por los siguientes miembros:

_______________________________
Job Ángel Ledezma Pérez, Dr. Ing.
**DIRECTOR DE CARRERA
INGENIERÍA MECATRÓNICA**

_______________________________
Daniel Raúl Asturizaga Hurtado de Mendoza Profesor, Ing.
**TUTOR**

_______________________________
Diego Paúl Mise Cruz, Mtr. Ing.
**RELATOR**

_______________________________
Marco Antonio Justiniano Pardo, Msc. Ing.
**INVITADO**

Santa Cruz de la Sierra, 23 de septiembre del 2020.

## AGRADECIMIENTOS

Quiero expresar mi agradecimiento con la Universidad Católica Boliviana, por abrirme las puertas y en especial con el PhD. Job Ledezma y el Ing. Daniel Asturizaga por haberme brindado su colaboración en todo el proceso de investigación y redacción de este trabajo. Por haberme orientado en todos los momentos en que necesite consejos.

Agradezco profundamente a mi familia, en especial a mi padre y mi madre que con su esfuerzo y dedicación me ayudaron a culminar mi vida universitaria y me dieron su apoyo cuando más lo necesitaba.

Finalizando, quisiera dedicar todo este esfuerzo a la memoria de mi querida abuelita Judith Arias, quien me apoyo incondicionalmente para cumplir todas mis metas y sueños.

# RESUMEN

Los robots manipuladores, lograron convertirse en partes esenciales de diferentes áreas en donde existen ambientes hostiles para los humanos como, por ejemplo: las áreas nucleares, en altas profundidades del mar o en el área espacial. En tareas criticas como en el punto anterior, los robots manipuladores implementan un sistema de control de fuerza en sus puntos de contacto con el objeto a manipular, lo que les permitiría aplicar la fuerza suficiente para manipular un objeto y no soltarlo, además de limitar la fuerza máxima aplicada para no aplastar el objeto a manipular.

Uno de los sistemas utilizados para el control de fuerza en robots manipuladores son los sistemas de actuadores elásticos en serie. Este sistema se compone principalmente por tres componentes, un elemento elástico que acumulara la fuerza entrante, un sensor de fuerza que realizara la lectura del valor de fuerza en el sistema y un actuador que se encargara de regular la posición del elemento elástico para controlar la fuerza de salida del sistema.

Una de las dificultades de aplicar un control de fuerza por SEA, es la complejidad que existe al implementar sus tres componentes principales, los cuales tienen que trabajar conjuntamente uno tras el otro. Con el objetivo de facilitar la implementación de un control de fuerza por SEA, en la presente tesis se desarrolla el sensor de fuerza neumático.

Un sensor de fuerza neumático se diferencia del resto de sensores por que tiene la capacidad de trabajar como un sensor de fuerza y como un elemento elástico. En particular, estas características facilitan la implementación de un control de fuerza mediante SEA, al reducir la cantidad de componentes necesarios. Por otra parte, el sensor de fuerza neumático cuenta con proporciones reducidas para facilitar su instalación en robots manipuladores y prótesis biomecatrónicas.

El primer paso que se realizó para el desarrollo del sensor de fuerza neumático fue la construcción del modelo matemático del sensor, para posteriormente utilizar el software *Matlab/Simulink* en la simulación del mismo. Con los datos obtenidos de la simulación del modelo matemático, se procedió al desarrollo del modelo CAD y de los planos del sensor en el software SolidWorks.

Posteriormente, se procedió a la construcción del prototipo del sensor de fuerza neumático en base de los planos realizados en el software SolidWorks. Una vez finalizado la etapa de construcción del sensor de fuerza neumático, se realizó la calibración y clasificación del sensor de fuerza en base a la normativa UNE-EN ISO 376, además se realizaron las pruebas experimentales para la validación del sensor.

Una vez obtenido la clasificación del sensor de fuerza neumático, se realizó la comparación de los resultados de la simulación del modelo matemático, con los de la prueba experimental. En la comparación se pudo evidenciar una coherencia grafica en los resultados obtenidos, validando el sistema del sensor de fuerza neumático.

**Palabras-clave:** Sensor de fuerza. Control de fuerza. Neumática.

# ABSTRACT

Manipulator robots managed to become essential parts of different areas where there are hostile environments for humans, such as: nuclear areas, in the deep sea or in the space area. In critical tasks such as in the previous point, manipulator robots implement a force control system at their points of contact with the object to be manipulated, which would allow them to apply enough force to manipulate an object and not release it, in addition to limit the maximum force applied so as not to crush the object to be manipulated.

One of the systems used for the control of force in manipulator robots, is the system of elastic actuators in series. This system mainly consists of three components, an elastic element that will accumulate the incoming force, a force sensor that will read the force value in the system, and an actuator that oversees regulating the position of the elastic element to control the force. system output.

One of the difficulties of applying a SEA force control is the complexity that exists when implementing its three main components, which must work together one after the other. To facilitate the implementation of a force control by SEA, in this thesis the pneumatic force sensor is developed.

A pneumatic force sensor differs from other force sensors in that it can work as a force sensor and as an elastic element. These features facilitate the implementation of force control by SEA, by reducing the number of components required. On the other hand, the pneumatic force sensor has reduced proportions to facilitate its installation in manipulator robots and biomechatronic prostheses.

The first step that was made for the development of the pneumatic force sensor was the construction of the mathematical model of the sensor, to later use the *MATLAB / Simulink* software to simulate it. With the data obtained from the simulation of the mathematical model, the CAD model and the sensor planes were developed in SolidWorks software.

Subsequently, the prototype of the pneumatic force sensor was built based on the plans made in the SolidWorks software. Once the stage of construction of the pneumatic force sensor was completed, the calibration and classification of the force sensor was carried out based on the UNE-EN ISO 376 standard, and the experimental tests were carried out to validate the sensor.

Once the classification of the pneumatic force sensor was obtained, the results of the simulation of the mathematical model were compared with the results of the experimental test.

In the comparison, it was possible to show a graphic coherence in the results obtained, validating the pneumatic force sensor system.

# ÍNDICE DE FIGURAS

# ÍNDICE DE TABLAS

## LISTA DE SIMBOLOS

| | | |
|---|---|---|
| $S$ | Sensibilidad de un sensor. | [-] |
| $X_N$ | Deformación correspondiente al alcance máximo. | [-] |
| $X_{0f}$ | Deformación correspondiente sin carga. | [-] |
| $i_0$ | Indicación leída en el dispositivo indicador antes de la aplicación de la carga, para el valor de carga nula. | [-] |
| $i_f$ | Indicación leída en el dispositivo indicador después de la aplicación de la carga, para el valor de carga nula. | [-] |
| $X_1$ | Deformación en la primera serie. | [-] |
| $X_2$ | Deformación en la segunda serie. | [-] |
| $\overline{X}_{wr}$ | Valor medio de las transformaciones sin rotación. | [-] |
| $X_{\max}$ | Deformación máxima en las tres series. | [-] |
| $X_{\min}$ | Deformación mínima en las tres series. | [-] |
| $\overline{X}_r$ | Valor medio de las deformaciones de rotación. | [-] |
| $c$ | Error relativo de fluencia. | [-] |
| $p_a$ | Presión de la cámara interna del sensor de fuerza neumático. | $[Pa]$ |
| $x$ | Posición del pistón del sensor de fuerza neumático. | $[m]$ |
| $\dot{x}$ | Velocidad del pistón. | $\left[\dfrac{m}{s}\right]$ |
| $R$ | Constante de gas del aire. | $\left[\dfrac{J}{kgK}\right]$ |
| $T_0$ | Temperatura del aire estándar. | $[K]$ |
| $A_a$ | Área útil de la cara del pistón. | $[m^2]$ |
| $qm_a$ | Caudal másico que entra a la cámara interna. | $[kg/s]$ |
| $V_{di}$ | Volumen muerto de la cámara interna. | $[m^3]$ |
| $F_p$ | Fuerza Neumática. | $[N]$ |
| $F_g$ | Fuerza de la gravedad. | $[N]$ |

| $F_{fr}$ | Fuerza de fricción. | [$N$] |
| $f_{cf}$ | Fricción de coulomb. | [N] |
| $f_{vf}$ | Fricción viscosa. | [N] |
| $P_a$ | Presión de la cámara interna del sensor de fuerza neumático. | [Pa] |
| $P_b$ | Presión atmosférica. | [Pa] |
| $MP$ | Masa del pistón. | [kg] |
| $B$ | Coeficiente de fricción viscosa. | [-] |
| Frc | Fricción de coulomb. | [N] |
| $w_c$ | Incertidumbre típica relativa combinada. | [-] |

## LISTA DE ABREVIATURAS Y SIGLAS

IEEE: *Institute of Electrical and Electronics Engineers* (Instituto de Ingeniería Eléctrica y Electrónica).

ASME: *The American Society of Mechanical Engineers* (Sociedad Americana de Ingenieros Mecánicos).

SEA: *Series elastic actuator* (Actuador elástico en serie).

ISO: *The International Organization for Standardization* (Organización Internacional de Normalización).

NI: *National Instruments.*

# INDICE

# CAPÍTULO I

# MARCO REFERENCIAL

## 1.1. INTRODUCCIÓN

Cuando se habla sobre sistemas de control de fuerza, en las últimas 3 décadas se observa un notable avance al desarrollarse los actuadores elásticos en serie, que a lo largo del texto serán referidos por sus siglas en ingles SEA (*Series Elastic Actuator)*. Los cuales, en aplicaciones donde se requiere un preciso control de fuerza, remplazan a los sistemas rígidos de control de fuerza. Pero en cambio, los SEA comprenden más elementos a implementar. Tienen que implementar físicamente un elemento elástico en serie a la fuente de energía mecánica en el sistema, además de un sensor de fuerza.

Un control de fuerza es requerido en tareas como el posicionamiento preciso de una carga o la actuación en una prótesis humana biomecatrónica. Inspirados en la biología, Williamson (1995) diseñó un sistema, donde posicionando un elemento elástico[1] en serie con una fuente de poder, como un motor de corriente continua con engranajes de reducción o un pistón hidráulico, obtenía un sistema con un control de fuerza más estable, sacrificando rigidez y ancho de banda de fuerza (Pratt y Williamson 1995). En la Figura 1.1 a) se observan los componentes de un SEA.

---

[1] aquellos a los que se les aplica una fuerza, que se va acumulando, pero que una vez que ha desaparecido, el elemento mecánico recupera su forma inicial devolviendo la energía acumulada.

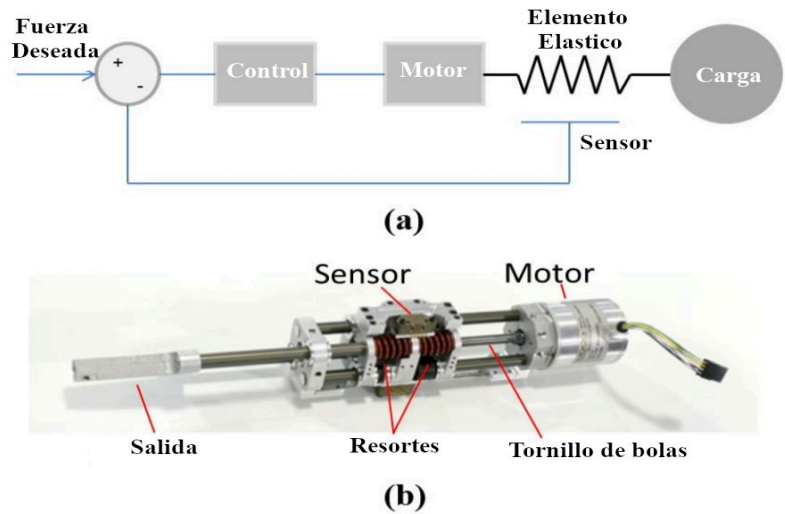

**Figura 1.1** a) Diagrama de un esquema de un SEA. b) Imagen referencial de un SEA.
**Fuente:** Adaptado de Cestari (2016)

Este sistema conocido como los SEA fue aplicado exitosamente en diversos proyectos por casi 20 años. Su configuración, al ser compuesto por un elemento elástico adherido a la fuente de energía mecánica, presenta ventajas sobre los actuadores rígidos, como tolerancia a impactos de cargas, incrementar el pico de potencia máxima, tener una baja impedancia mecánica de salida y almacenar energía mecánica pasivamente (Arnaldo Gomes Leal Junior y Additional 2016).

Para facilitar la implementación de los SEA en diferentes sistemas mecatrónicos, la presente tesis propone el diseño de un sensor de fuerza neumático. La principal ventaja de este sensor de fuerza neumático es la capacidad de trabajar como el elemento elástico en SEA y a la vez que realiza el cálculo indirecto de fuerza.

El nuevo sensor de fuerza neumático permite obtener el valor de fuerza indirectamente, del valor de presión de un compartimiento de aire comprimido interno. El comportamiento dinámico de aire comprimido dentro de la cámara principal del sensor de fuerza neumática varía su dimensión en función a la fuerza aplicada sobre el sensor, trabajando como un pistón neumático con los puertos de conexión sellados.

Este trabajo, propone el desarrollo de un nuevo tipo de sensor de fuerza neumático, el cual posibilite un control de fuerza aplicada en las terminales de agarre de un robot manipulador, trabajando como un actuador elástico en serie. La implementación de un SEA utilizando este sensor de fuerza neumático, pretende mejorar la manipulación de objetos en terminales de agarre.

### 1.2. HIPOTESIS

Es posible desarrollar un prototipo de sensor de fuerza neumático para cargas pequeñas, aplicable a sistemas robóticos, que trabaje de forma simultánea como elemento elástico y sensor de fuerza.

### 1.3. OBJETIVOS

#### 1.1.1. Objetivo General

♦   Desarrollar un prototipo de sensor de fuerza neumático para aplicaciones de control de fuerza en robótica aplicada.

#### 1.1.2. Objetivos Específicos

♦   Realizar el cálculo y dimensionamiento de un sensor de fuerza neumático.

♦   Modelar matemáticamente el sensor de fuerza neumático.

♦   Simular el comportamiento dinámico del sensor de fuerza neumático.

♦   Diseñar y fabricar el prototipo del sensor de fuerza neumático.

♦   Validar experimentalmente el sensor de fuerza desarrollado.

♦   Calibrar y clasificar el sensor de fuerza neumático en función a la norma UNE- EN ISO 376.

### 1.4. JUSTIFICACIÓN

El presente estudio presenta una alternativa de aplicación de los actuadores elásticos en serie, ampliamente utilizados en sistemas de control de fuerza. En este sentido, el hecho de usar aire comprimido encapsulado como elemento elástico y la variación de la presión interna como forma de medición indirecta de fuerza, permite el desarrollo de un transductor de fuerza de baja complejidad mecánica y de reducidas dimensiones.

### 1.5. LIMITES Y ALCANCES

#### 1.1.3. Límites

Los límites del desarrollo del sensor de fuerza neumático son los siguientes:

- El diseño mecánico del sensor de fuerza neumático debe ser construido con las herramientas y maquinaria disponible localmente.
- El rango de fuerza de lectura del sensor de fuerza neumático debe ser capaz de medir mínimamente la fuerza promedio ejercida en un agarre cilíndrico de una mano humana, la cual es de 33 N. (Pérez-González, Jurado-Tovar y Sancho-Bru 2011).

### 1.1.4. Alcances

- La presente tesis explorara los aspectos teóricos y prácticos de los sensores de fuerza neumático.

- La investigación abarcará únicamente hasta la validación, calibración y clasificación del sensor de fuerza neumático

## 1.6. ESTRUCTURA DEL DOCUMENTO

- Capítulo 2 Marco Teórico

Corresponde al estudio de las características técnicas de los sensores, el estado del arte de los sensores de fuerza y a la teoría sobre el cálculo de errores e incertidumbre para la calibración del sensor de instrumentos de fuerza.

- Capítulo 3 Modelo matemático y simulación

Se desarrolla el modelo matemático del sensor de fuerza neumático. Posteriormente, se realiza la simulación del sistema en el software *Matlab/Simulink*.

- Capítulo 4 Diseño del sensor de fuerza neumático

Se desarrolla y desglosa el diseño mecánico de todos los componentes del sensor de fuerza neumático en el software de diseño asistido por computadora *SolidWorks*.

- Capítulo 5 Prototipos de sensores de fuerza neumáticos y bancos de prueba

Se presenta el desarrollo y construcción de los diferentes prototipos de sensor de fuerza neumático desarrollados, hasta llegar al prototipo final. De igual forma abarca el desarrollo y construcción del banco de pruebas de fuerza, utilizado en las pruebas experimentales.

♦ Capítulo 6 Pruebas Experimentales y análisis de resultados

En la primera parte del Capítulo 6, se desarrolla el procedimiento de calibración y clasificación del sensor de fuerza neumático, en base a la norma UNE-EN ISO 376. En la segunda parte, se abarcan las pruebas experimentales correspondientes para realizar la comparación y validación del sistema con el modelo matemático.

♦ Capítulo 7 Marco conclusivo

Se desarrollan las conclusiones determinadas por los capítulos anteriores.

.

# CAPÍTULO II

# MARCO TEORICO

## 2.1. ESPECIFICACIONES DE LOS SENSORES

Un sensor es una parte esencial de cualquier sistema de procesamiento de información que trabaja en más de una unidad física. Ejemplos son las unidades ópticas, eléctricas, magnéticas, térmicas y mecánicas. Un transductor es parte de un sistema de medición que convierte en información las medidas que calcula de una unidad a otra, idealmente sin pérdida de información (Regtien 2012).

Un sensor tiene al menos una entrada y una salida. En instrumentos de adquisición de datos, donde el procesamiento de información se realiza mediante señales eléctricas, la salida o la entrada son eléctricas, mientras que la otra es una señal no eléctrica (desplazamiento, temperatura, elasticidad, entre otros). Un sensor con una entrada no eléctrica es un sensor de entrada, destinado a convertir una cantidad no eléctrica en una señal eléctrica para medir esa cantidad (Regtien 2012).

Las imperfecciones de los sensores generalmente se detallan en las hojas de datos proporcionadas por el fabricante. Estas especificaciones del sensor informan al usuario sobre las desviaciones del comportamiento ideal del sensor. El usuario debe aceptar imperfecciones técnicas, siempre que no superen los valores especificados.

Las características que describen el rendimiento del sensor se pueden clasificar en cuatro grupos (Regtien 2012):

- **Características estáticas**, que describen el rendimiento con respecto a lecturas muy lentas,
- **Características dinámicas**, que especifican la respuesta del sensor a las variaciones en el tiempo y en el rango de medición.

- **Características ambientales**, que relacionan el rendimiento del sensor después o durante la exposición a condiciones externas específicas (por ejemplo, presión, temperatura, vibración y radiación).
- **Características de calidad**, que describen la esperanza de vida del sensor.

### 2.1.1. Especificaciones Generales de los sensores

#### 2.1.1.1. Sensibilidad

La sensibilidad es la relación del cambio incremental en la salida del sensor ($dy$) al cambio incremental del mensurando en la entrada ($dx$). Matemáticamente, la sensibilidad se expresa como

$$S = \frac{dy}{dx} \qquad (2.1)$$

Por lo general, un sensor también es sensible a los cambios, en función a la cantidad de entrada prevista, como la temperatura ambiente o la tensión de alimentación. Estos cambios no deseados debido a factores externos también deben especificarse para una interpretación adecuada del resultado de la medición. (Fraden 2010).

#### 2.1.1.2. Linealidad

La proximidad de la curva de calibración a una línea recta especificada muestra la linealidad de un sensor. Su grado de semejanza con una línea recta describe cuán lineal es un sistema. (Kalantar-zadeh 2013)

Como se puede observar en la figura 2.1, un ejemplo de la curva linealizada de un sensor respecto a su respuesta actual, el cual permite obtener el grado de desviación lineal en el mismo sistema.

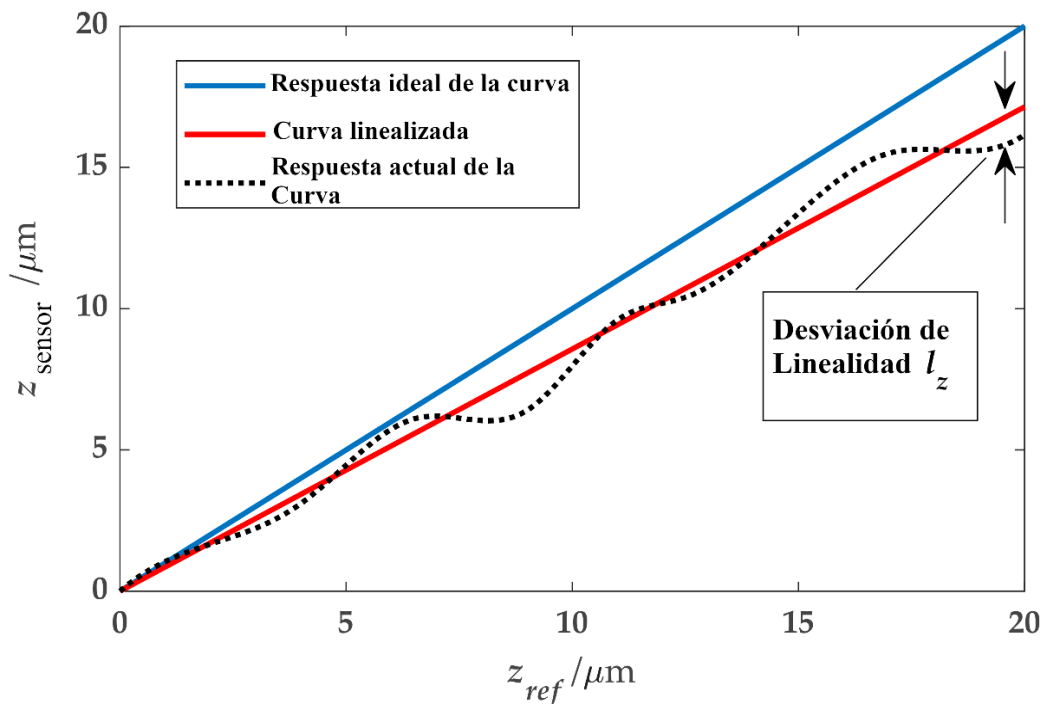

**Figura 2.1.** Ejemplo de una curva de respuesta y la derivación del factor de
amplificación y la desviación de linealidad.

**Fuente**: Adaptado de Haitjema (2020)

### 2.1.1.3. *Histéresis*

La histéresis es la diferencia entre las lecturas de salida para la misma medición,
dependiendo de la trayectoria seguida por el sensor como se puede observar en la Figura 2.2.
Las causas típicas de histéresis son la geometría del diseño, la fricción y los cambios
estructurales en los materiales del sensor (Kalantar-zadeh 2013).

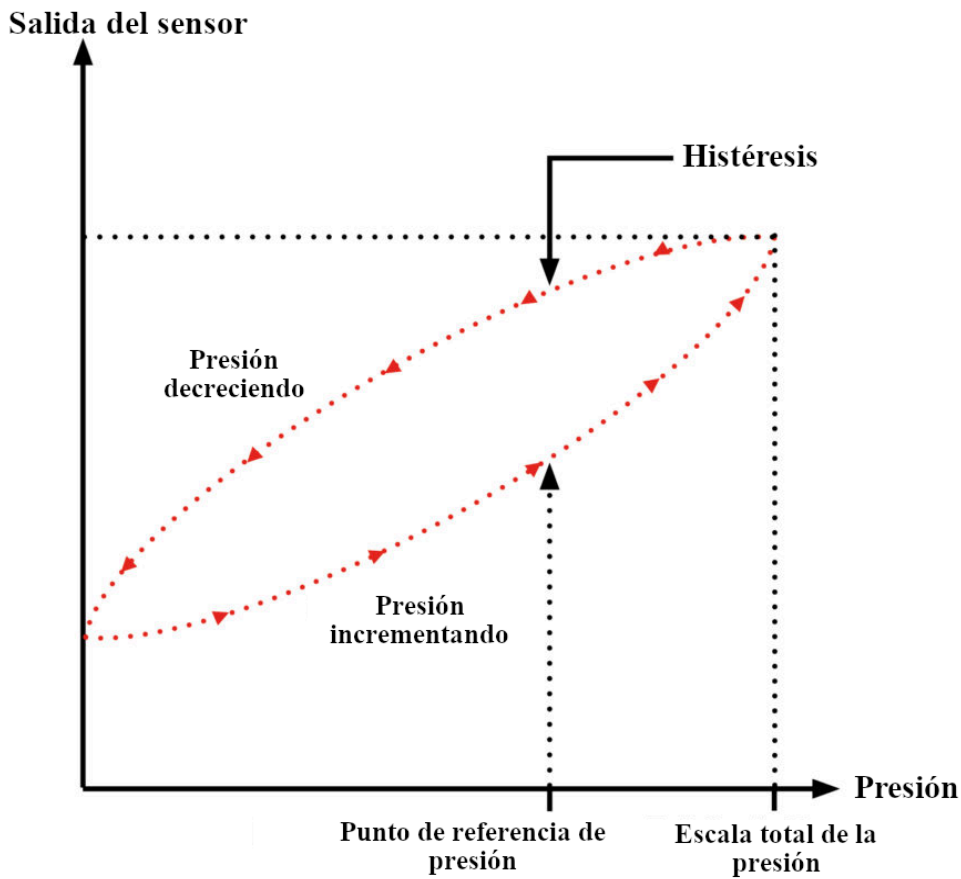

**Figura 2.2.** Representación gráfica de la Histéresis.
**Fuente:** Adaptado de Morcom (2020)

### 2.1.1.4. *Rango de medición*

Los valores máximos y mínimos que se pueden medir con los sensores se denominan rango de medición, también es llamado rango dinámico. Todos los sensores están diseñados para operar en un rango específico. Las señales fuera de este rango pueden ser inconmensurables, pueden causar inexactitudes grandes e incluso daños irreversibles al sensor. Generalmente, el rango de medición de un sensor se especifica en su ficha técnica. (Kalantar-zadeh 2013)

Por ejemplo, si se expone un sensor de temperatura a temperaturas superiores a su rango de medición especificadas en su hoja técnica, puede causar daños al sensor (Kalantar-zadeh 2013).

### *2.1.1.5. Tiempo de respuesta*

Cuando un sensor realiza una medición, el tiempo que requiere para alcanzar un valor estable es el tiempo de respuesta. Las especificaciones del tiempo de respuesta, siempre tienen que ir acompañados con las indicaciones del escalón de entrada, y el rango de salida donde el tiempo de respuesta está definido, por ejemplo, del 10% al 90% del rango de medición del sensor como se puede observar en la figura 2.3 (Regtien 2012).

**Figura 2.3.** Ejemplo gráfico del tiempo de respuesta de un sensor.

**Fuente:** Adaptado de Bossart (2020)

## 2.2. ESTADO DEL ARTE DE LOS SENSORES DE FUERZA

En la física, la magnitud de la fuerza del sistema internacional es importante, ya que su lectura se requiere en diferentes áreas de la ingeniería. Para este fin se utilizan distintos tipos de sensores que permiten medir la fuerza o peso de los objetos edificaciones, etc. (Kirkland 2007).

Los sensores de fuerza se pueden dividir en dos clases:

- cuantitativos y

- cualitativos.

El sensor de fuerza cuantitativo calcula la fuerza y representa su valor en función de una señal de salida comprendida dentro de un rango de valores. Ejemplos de esta clase de sensores son los sensores de galgas extensométricas y las celdas de carga.

Los sensores de fuerza cualitativos son los dispositivos que tienen la función de simplemente indicar si se aplica una fuerza suficientemente fuerte o no. Esto quiere decir que, la señal de salida indica cuándo la magnitud de la fuerza excede un nivel predeterminado. Un ejemplo de estos sensores son los botones de cualquier dispositivo electrónico, en los cuales solo cuando son presionados con la suficiente fuerza estos hacen contacto. Los sensores de fuerza cualitativos se utilizan con frecuencia para la detección de movimiento y posición (Regtien 2012).

Los diferentes métodos para calcular fuerza pueden ser clasificados y enumerados de la siguiente forma:

1) al equilibrar la fuerza a medir con la fuerza gravitacional de una masa estándar.
2) al medir la aceleración de una masa conocida a la que se aplica la fuerza.
3) al equilibrar la fuerza contra una fuerza desarrollada electromagnéticamente.
4) al convertir la fuerza en una presión de fluido y medir tal presión.
5) al medir la deformación producida en un miembro elástico por la fuerza desconocida.

En los sensores modernos, el método más utilizado es la numero 5 de la lista anterior, a diferencia de los métodos de los números 3 y 4 son utilizados ocasionalmente.

La mayoría de los sensores, no convierten la fuerza directamente en una señal eléctrica y por lo general, se requieren pasos intermedios. Por lo cual, muchos sensores de fuerza son considerados complejos (Fraden 2010).

### 2.2.1. Tipos de sensores de Fuerza

#### 2.2.1.1. Galgas Extensiométricas

Cuando un objeto es sometido a una tensión, el resultado en el mismo es el cambio fraccionario de sus dimensiones. El cambio de dimensión se divide por la dimensión original. Como ejemplo se puede observar en la Figura 2.4 (a) el efecto de tensión donde la barra se estira una longitud $\Delta L$ cuando se aplica una fuerza paralela a su longitud. Como en la figura 2.4(b) el efecto de compresión donde la misma varilla está comprimida por fuerzas con la misma magnitud en la dirección opuesta (Sinclair 2001).

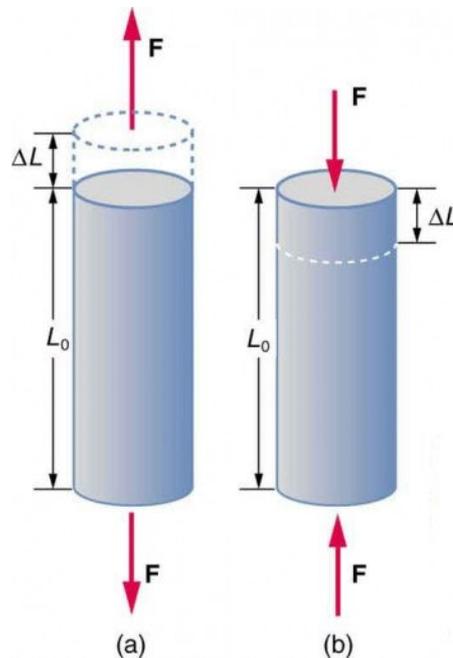

**Figura 2.4.** (a) efecto a la tensión. (b) efecto a la compresión
**Fuente:** (Elasticity: Stress and Strain, Physics 2020)

La deformación se puede definir para el área o para las mediciones de volumen de manera similar al cambio dividido por la cantidad original (Fraden 2010).

Para fluidos como líquidos o gases que se pueden comprimir de manera uniforme en todas las dimensiones, la tensión total es la fuerza por unidad de área, que es igual a la presión aplicada, y la deformación es el cambio de volumen dividido por el volumen original (Sinclair 2001).

Los transductores de deformación más comunes son para deformación mecánica por tracción. La medición de la deformación permite calcular la cantidad de presión a través del conocimiento del módulo elástico del mismo. Los módulos elásticos más comúnmente utilizados son el módulo lineal de Young, el módulo de corte (torsión), y el módulo volumétrico(presión) (Sinclair 2001).

La forma más común de calcular presión o fuerza es utilizando galgas extensiométricas por resistencia, un ejemplo de la misma es la Figura 2.5. Una galga extensiométrica por resistencia consiste en un material conductor en forma de un cable o tira delgada que se une firmemente a un soporte elástico en el que se detecta la deformación. El coeficiente de expansión térmica del soporte elástico debe coincidir con el del cable. Se pueden usar muchos metales para fabricar galgas extensométricas por resistencia. Los materiales más comunes son las aleaciones *constantán, nichrome, advance* y *karma*. Las resistencias típicas varían de 100 a varios miles de ohmios (Sinclair 2001).

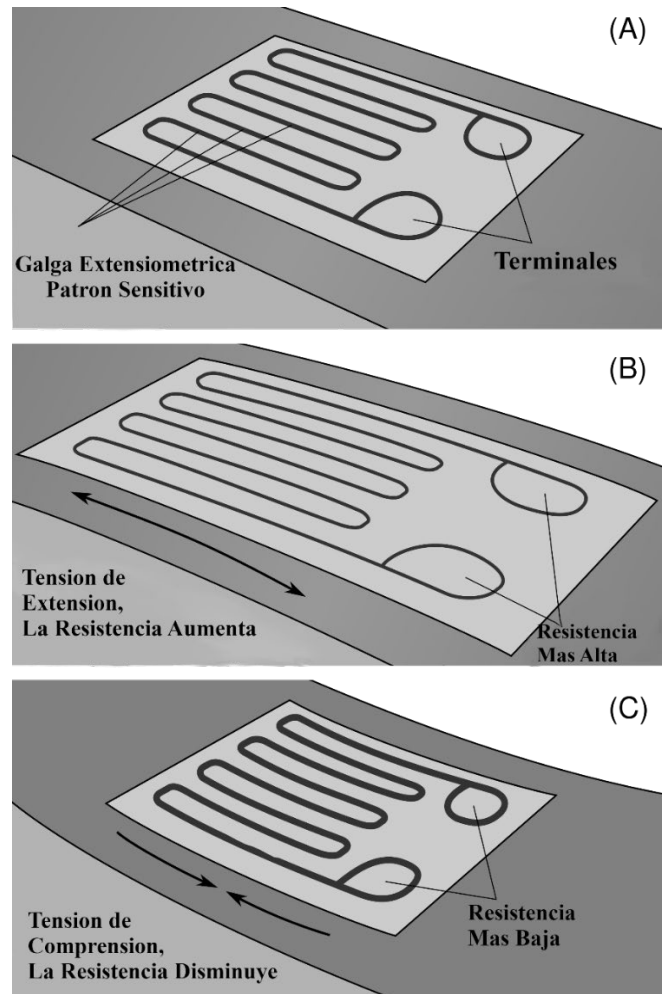

**Figura 2.5.** Visualización del concepto del trabajo de una Galga Extensiométrica.
**Fuente:** Adaptado de Izantux (2020)

Para obtener una buena sensibilidad, el sensor debe tener segmentos transversales longitudinales largos y cortos. Las galgas pueden estar dispuestos de muchas maneras para medir deformaciones en diferentes ejes. Comúnmente, están conectados al circuito de un puente *Wheatstone*. Se tiene que señalar que los semiconductores de las galgas extensiométricas son bastante sensibles a las variaciones de temperatura. Por lo tanto, los circuitos de interfaz o los medidores deben contener redes de compensación de temperatura(Fraden 2010).

### 2.2.1.2. *Sensores Táctiles*

Los sensores táctiles se pueden dividir en tres grupos:

♦   Sensores de contacto: Estos sensores detectan y miden las fuerzas de contacto en puntos en específico. Un sensor de contacto generalmente es un dispositivo que se activa cuando se le aplica una fuerza, superando un límite definido. Detecta cuando está haciendo contacto, y cuando no.

♦   Sensores de espacio: Estos sensores detectan y miden las fuerzas distribuidas en un área sensorial predeterminada. Una matriz de detección de espacio se puede considerar como un grupo coordinado de sensores de contacto.

♦   Sensores de deslizamiento: Estos sensores detectan y miden el movimiento de un objeto relativo al sensor. Esto se puede lograr mediante un sensor de deslizamiento especialmente diseñado, mediante la interpretación de los datos de un sensor de contacto o una matriz de sensores de espacio.

En general, los sensores táctiles son una clase especial de transductores de fuerza[2] o presión que se caracterizan por un grosor pequeño, como se puede observar en la Figura 2.6. Esto hace que los sensores sean útiles en las aplicaciones donde se puede desarrollar fuerza o presión entre dos superficies, que se encuentran muy próximas entre sí (Sinclair 2001).

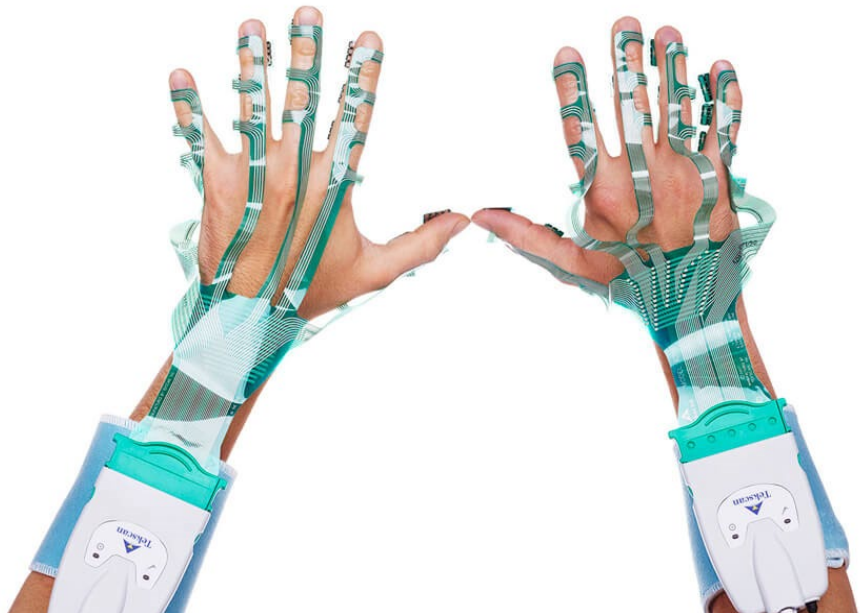

**Figura 2.6.** Ejemplo de uso de Sensores táctiles
**Fuente:** ("Tactile Sensor Solutions", 2019)

---

[2] son dispositivos que nos permiten obtener una señal eléctrica proporcional a la fuerza que se aplica sobre ellos.

Los ejemplos incluyen la robótica, donde los sensores táctiles se pueden colocar en las superficies de agarre de un actuador mecánico, para proporcionar una retroalimentación al desarrollar un contacto con un objeto, muy similar al funcionamiento de los sensores táctiles en la piel humana (Fraden 2010).

### 2.2.1.3. Tipo de Sensor Táctil: Sensor Interruptor

Se pueden usar varios métodos para fabricar sensores táctiles. Algunos requieren la formación de una capa delgada de material, que responde a la deformación(Fraden 2010).

Se puede formar un sensor táctil simple que produce una salida de encendido-apagado, con dos hojas de papel de aluminio y un material espaciador, como en la Figura 2.7. El espaciador tiene orificios redondos (o cualquier otra forma adecuada). Una hoja está conectada a tierra y la otra está conectada a una resistencia en configuración *pull-up*.

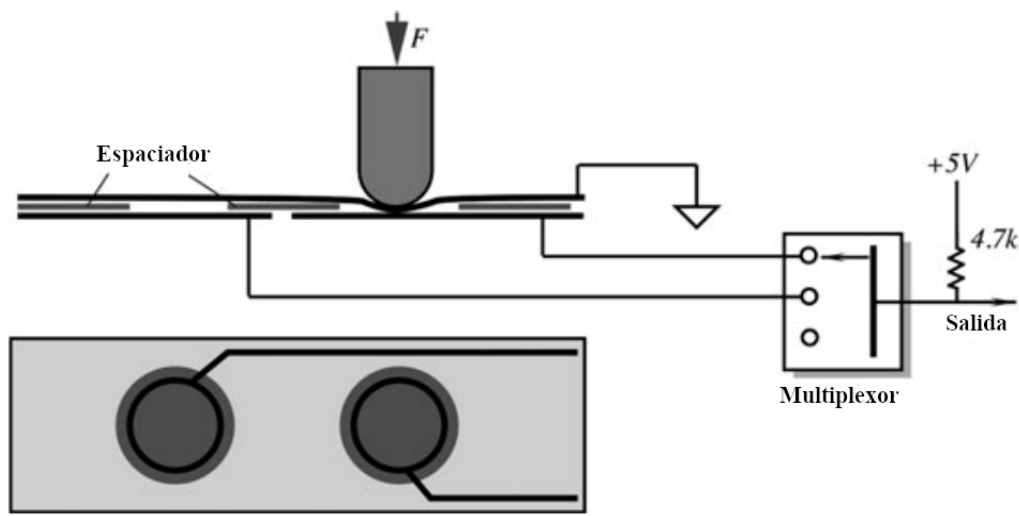

**Figura 2.7.** Interruptor de membrana como sensor táctil.
**Fuente:** Adaptado de Fraden (2010)

Se puede usar un multiplexor si se requiere más de un área de detección. Cuando se aplica una fuerza externa al conductor superior sobre el orificio en el espaciador, la hoja superior se flexiona y, al llegar al conductor inferior, hace un contacto eléctrico, conectando a tierra la resistencia *pull-up*. La señal de salida se convierte en cero, lo que indica la fuerza aplicada. Se pueden formar múltiples puntos de detección imprimiendo filas y columnas de una tinta conductora. Al tocar un área particular del sensor, la fila y la columna correspondientes se unirán, lo que indica la fuerza en una ubicación en específico.

### 2.2.1.4. Tipo de Sensor Táctil: Sensores Piezoeléctricos

El sensor piezoeléctrico, es un dispositivo que utiliza el efecto piezoeléctrico para medir los cambios de presión, aceleración, temperatura, tensión o fuerza al convertirlos en carga eléctrica. La capacidad del material piezoeléctrico para convertir la deformación mecánica en carga eléctrica se denomina efecto piezoeléctrico. La piezoelectricidad generada es proporcionalmente a la presión o fuerza aplicada a los materiales sólidos de cristal piezoeléctrico (Kumar, 2019).

El sensor no presenta la corrosión, picaduras o rebotes que normalmente están asociados con los interruptores de contacto. Puede sobrevivir más de diez millones de ciclos sin fallar. La simplicidad del diseño lo hace efectivo en aplicaciones que incluyen: contadores para líneas de ensamblaje y rotación de ejes, interruptores para procesos automatizados, detección de impacto para productos dispensados a máquina, etc. (Fraden 2010).

Los dos materiales de detección principalmente utilizados para los sensores piezoeléctricos son las cerámicas piezoeléctricas[3] y los materiales monocristalinos (como el cuarzo). La sensibilidad de los materiales cerámicos es mayor que la de los materiales naturales de un solo cristal, pero su alta sensibilidad se degrada con el tiempo.

Los materiales naturales de un solo cristal (cuarzo, fosfato de galio, turmalina, etc.) son menos sensibles, pero tienen una mayor estabilidad (Kumar, 2019).

Hay dos tipos de sensores piezoeléctricos basados en el diseño de conexión del cable: tipo guiado y tipo de pin, como se muestra en la Figura 2.8. Ambos son comúnmente comerciables (Kumar, 2019).

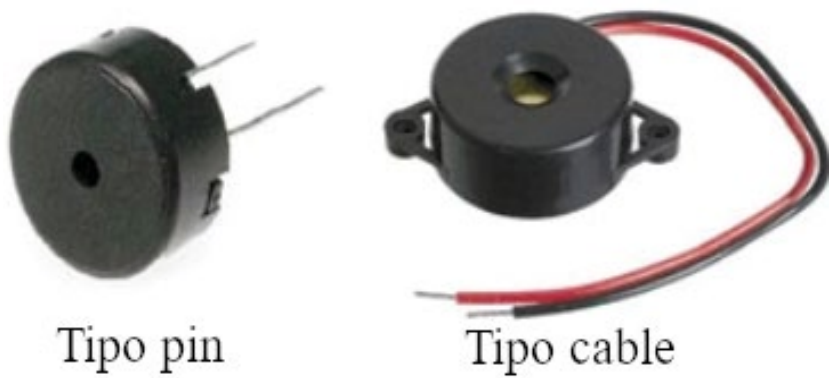

**Figura 2.8.** Sensores Piezoeléctricos.
**Fuente:** Adaptado de Kumar (2019).

---

[3] material con la capacidad para generar una carga eléctrica al ser cargados mecánicamente con presión o tensión.

### 2.2.1.5. *Tipo de Sensor Táctil: Sensores Piezoresistivos*

Otro tipo de sensor táctil es un sensor piezoresistivo. Se puede fabricar utilizando materiales cuya resistencia eléctrica están en función a su deformación. El sensor incorpora una piezoresistencia, cuya resistencia varía con la presión aplicada en la misma. Dichos materiales son elastómeros conductores o tintas sensibles a la presión.

Un elastómero conductor está fabricado de caucho de silicona, poliuretano y otros compuestos que están impregnados con partículas o fibras conductoras. Los principios de los sensores táctiles elastoméricos, se basan en variar el área de contacto, cuando el elastómero se aprieta entre dos placas conductoras como en la Figura 2.9., o al cambiar su grosor. Cuando la fuerza externa varía, el área de contacto en la interfaz entre la superficie presionada y el elastómero cambia, lo que resulta en una reducción de la resistencia eléctrica. A cierta presión, el área de contacto alcanza su máximo y la transferencia (Fraden 2010).

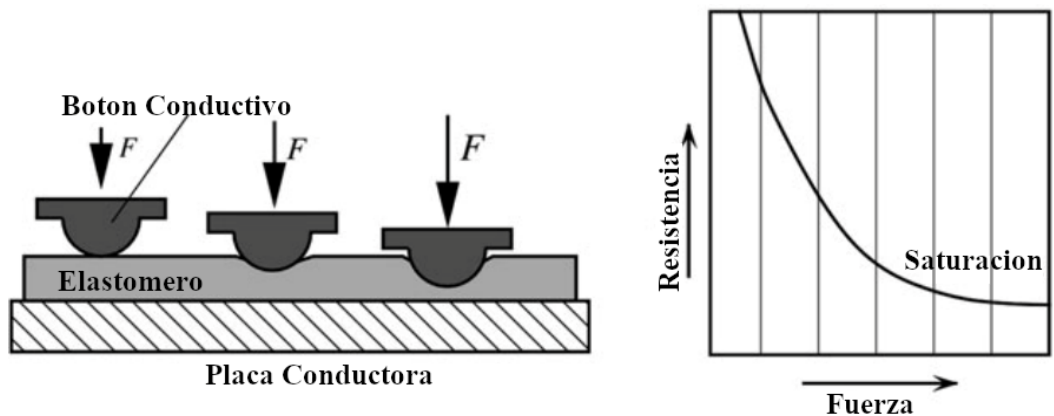

**Figura 2.9.** Variación de Fuerza aplicada sobre una resistencia sensible a la fuerza.
**Fuente:** Adaptado de Fraden (2010)

Sin embargo, la resistencia puede variar notablemente su valor cuando el polímero se somete a una presión prolongada. Se puede fabricar un sensor táctil piezoresistivo mucho más delgado con un semiconductor polímero cuya resistencia varía con la presión.

### 2.2.1.6. *Tipo de Sensor Táctil: Sensores MEM*

Los sensores MEM vienen de la definición de sensores micro electromecánicos. Los sensores táctiles en miniatura son especialmente demandados en robótica, donde se requiere una buena resolución espacial, alta sensibilidad y un amplio rango dinámico (Fraden 2010).

Los expertos en sensores estiman que la tecnología MEM se utiliza en más del 90% de los sensores de presión. La detección piezoresistiva y capacitiva se usa comúnmente en sensores de presión basados en MEM. Con su pequeño tamaño y pequeño empaque externo, los sensores MEM de todo tipo han sido utilizado en nuevas aplicaciones y han proporcionado nuevas características en muchos productos finales. Además, la fabricación de semiconductores proporciona consistencia, altos rendimientos, y bajo costo (Sinclair 2001).

Para las mediciones de presión, los sensores piezoresistivos de micro maquinado a granel. son la tecnología más utilizada debido a la facilidad del acondicionamiento de la señal, la amplia selección y el bajo costo. Estos sensores MEM se utilizan en los rangos de presión más comunes de 0 a 100 psi ("Engineering Resources Pressure Sensor Technologies", 2019).

## 2.3. CÁLCULO DE ERRORES E INCERTIDUMBRES, PARA LA CALIBRACIÓN DE INSTRUMENTOS DE FUERZA

En el procedimiento de calibración y clasificación de instrumentos de fuerza ME-002 del Centro Español de Metrología basada en la norma UNE-EN ISO 376, tiene como objetivo calibrar y clasificar aquellos instrumentos de medida de fuerza que se basan en métodos indirectos, utilizando como magnitud física de medida, la deformación elástica de un elemento sensor o una magnitud proporcional a esta, por ejemplo, la variación de presión del compartimiento interno del sensor de fuerza neumático.(Centro Español de Metrología 2019)

Los cálculos de errores e incertidumbres especificados en el procedimiento, a partir de los datos registrados en las pruebas experimentales, son los parámetros que sirven como base para la clasificación del instrumento de medida de fuerza según UNE-EN ISO 376.

### 2.3.1. Cálculo de errores

#### 2.3.1.1. Error relativo de cero, $f_0$

El error relativo de cero, es el error calculado cuando no se le aplica una carga al sensor, se calcula para cada serie con ayuda de la Ecuación (2.3) :(Centro Español de Metrología 2019)

$$f_0 = \frac{X_{0f}}{X_N} * 100 \qquad (2.2)$$

$$X_{0f} = i_f - i_0 \qquad (2.3)$$

donde:

$X_N$ = deformación correspondiente al alcance máximo.

$X_{0f}$ = deformación correspondiente sin carga.

$i_0$ = indicación leída en el dispositivo indicador antes de la aplicación de la carga, para el valor de carga nula.

$i_f$ = indicación leída en el dispositivo indicador después de la aplicación de la carga, para el valor de carga nula.

### 2.3.1.2. *Error relativo de reproducibilidad y repetibilidad,* $b, b'$

El error relativo de reproducibilidad y repetibilidad es el error que se calcula a partir de una serie de repeticiones de un mismo valor de fuerza.

<u>Sin rotación,</u> $b'$

Se calcula para cada valor de fuerza ensayada, se utilizan las medidas de las dos series $X_1$ y $X_2$ en las que el instrumento de medida de fuerza no ha sido cambiado de posición, con ayuda de la ecuación:

$$b' = \frac{|X_2 - X_1|}{\overline{X}_{wr}} * 100 \tag{2.4.}$$

$$\overline{X}_{wr} = \frac{X_1 + X_2}{2} \tag{2.5.}$$

donde:

$X_1$ = deformación en la primera serie.

$X_2$ = deformación en la segunda serie.

$\overline{X}_{wr}$ = valor medio de las transformaciones sin rotación.

<u>Con rotación,</u> $b$

Se obtiene para cada valor de fuerza ensayado, usando las medidas de las series $X_1$, $X_3$ y $X_5$, series a $0°$, $180°$ y $360°$ de giro del instrumento de medida de fuerza, con la siguiente ecuación:

$$b = \frac{\left| X_{\max} - X_{\min} \right|}{\overline{X}_r} * 100 \qquad (2.6.)$$

$$\overline{X}_r = \frac{X_1 + X_3 + X_5}{3} \qquad (2.7.)$$

donde:

$X_{\max}$ = deformación máxima en las tres series.

$X_{\min}$ = deformación mínima en las tres series.

$X_1$ = deformación de la primera serie.

$X_2$ = deformación para las cargas crecientes de la tercera serie.

$X_5$ = deformación para las cargas crecientes de la cuarta serie.

$\overline{X}_r$ = valor medio de las deformaciones de rotación.

### 2.3.1.3.    Error relativo de reversibilidad, $v$

El error relativo de reversibilidad es determinado, para cada una de las fuerzas de calibración mediante la diferencia entre los valores encontrados en el sentido creciente y en el sentido decreciente con ayuda de la siguiente ecuación:

$$v = \left| \frac{X_4 - X_3}{X_3} \right| \left( \frac{100}{2} \right) + \left| \frac{X_6 - X_5}{X_5} \right| \left( \frac{100}{2} \right) \qquad (2.8.)$$

donde:

$X_3$ = deformación para las cargas crecientes de la tercera serie.

$X_4$ = deformación para las cargas decrecientes de la tercera serie.

$X_5$ = deformación para las cargas crecientes de la cuarta serie.

$X_6$ = deformación para las cargas decrecientes de la cuarta serie.

### 2.3.1.4.    Error relativo de interpolación, $f_c$

El error relativo de interpolación es el error determinado de las variaciones de lectura por la variación de posición del sensor en cada una de las fuerzas.

Se calcula el polinomio de interpolación, por ejemplo, por el método de los mínimos cuadrados (ajuste lineal, cuadrático o cubico), utilizando los valores de $\overline{X}_r$ obtenidos, y los valores de fuerza de referencia aplicados durante la calibración, para obtener una ecuación de la forma:

$$X_a = f(F),$$ (2.9.)

La curva de interpolación no se hará pasar por fuerza nula.

El error relativo de interpolación se determina para cada una de las fuerzas de calibración con la ayuda de la ecuación siguiente:

$$f_c = \frac{\overline{X}_r - X_a}{X_a} * 100$$ (2.10.)

$\overline{X}_r$ = valor medio de las deformaciones con rotación.

$X_a$ = valor de la deformación calculado haciendo uso de una ecuación de regresión de primero, segundo o tercer grado, que proporciona el valor de deformación en función de la fuerza de calibración.

En los instrumentos de medida de fuerza mecánicos que cuentan con un comparador para la medida de la deformación, no se recomienda la utilización del polinomio de interpolación. Sin embargo, puede utilizarse la interpolación con la condición de que las características del comparador hayan sido determinadas previamente y que su error periódico sea despreciable respecto al error de interpolación del instrumento de medida de fuerza(Centro Español de Metrología 2019).

### 2.3.1.5.    *Error relativo de fluencia (creep),* $c$

El error relativo de fluencia es el error calculado tiempo antes y después de ser aplicadas las cargas en el sensor.

Se puede calcular la diferencia de las indicaciones tomadas a los 30 s ($i_{30}$) y a los 300 s ($i_{300}$) después de la aplicación o retirada de la fuerza de calibración máxima y se expresa esta diferencia como porcentaje de la deflexión máxima.

$$c = \left| \frac{i_{300} - i_{30}}{X_N} \right| 100 \qquad (2.11.)$$

### 2.3.2. Cálculo de incertidumbre

Para la estimación y cálculo de las incertidumbres, se aplican los criterios establecidos en la norma UNE-EN ISO 376.

Para los instrumentos clasificados por interpolación, la incertidumbre de calibración es la incertidumbre en el valor de fuerza calculado de la ecuación de interpolación, solamente para fuerza crecientes en las pruebas experimentales.

Los instrumentos clasificados para fuerzas específicas la incertidumbre de clasificación es la incertidumbre de la fuerza correspondiente a cualquier deformación igual a una de las deformaciones medias obtenidas durante la calibración para fuerzas crecientes.

Para cada fuerza de calibración, se calcula de las lecturas obtenidas durante la calibración una incertidumbre típica relativa combinada $w_c$ (Centro Español de Metrología 2019).

$$w_c = \sqrt{\sum_{i=1}^{8} w_i^2} \qquad (2.12.)$$

donde:

$w_1$ = incertidumbre típica relativa asociada a la fuerza de calibración aplicada;

$w_2$ = incertidumbre típica relativa asociada a la reproducibilidad;

$w_3$ = incertidumbre típica relativa asociada a la repetibilidad;

$w_4$ = incertidumbre típica relativa asociada a la resolución del indicador;

$w_5$ = incertidumbre típica relativa asociada a la fluencia;

$w_6$ = incertidumbre típica relativa asociada a la deriva en la salida del cero;

$w_7$ = incertidumbre típica relativa asociada a la temperatura del instrumento y

$w_8$ = incertidumbre típica relativa asociada a la interpolación.

### 2.3.2.1. *Incertidumbre de fuerza de calibración,* $w_1$

$w_1$ es la incertidumbre típica relativa asociada a las fuerzas aplicadas por la máquina de calibración. Esta es igual al error de la máquina, expresada en términos relativos, dividido por el valor de $k$ especificado (probablemente igual a 2) (Centro Español de Metrología 2019).

### 2.3.2.2. *Incertidumbre de reproducibilidad,* $w_2$

$w_2$ es, en cada escalón de fuerza aplicada, la desviación típica de la deformación creciente media obtenida durante la calibración, expresada como un valor relativo.

$$w_2 = \frac{1}{\left| \overline{X}_r \right|} \sqrt{\frac{1}{6} \sum_{i=1,3,5} \left( X_i - \overline{X}_r \right)^2} \qquad (2.13.)$$

Donde:

$X_i$ = deformación obtenida en series crecientes 1, 3, y 5, y $\overline{X}_r$ es el valor medio de estos tres valores.

### 2.3.2.3. *Incertidumbre de repetibilidad,* $w_3$

$w_3$ es, en cada escalón de fuerza aplicada, la contribución debida a la repetibilidad de la deformación medida en una sola orientación, expresada como valor relativo. Se calcula de la Ecuación 2.12:

$$w_3 = \frac{b'}{100\sqrt{3}} \qquad (2.14.)$$

Donde $b'$ es el error relativo de repetibilidad del instrumento.

### 2.3.2.4. *Incertidumbre de resolución,* $w_4$

Cada valor de deformación es calculado como la diferencia entre dos lecturas (la lectura en la fuerza aplicada restada de la lectura en una fuerza cero). Consecuentemente, la resolución del indicador debe ser incluida dos veces como dos distribuciones rectangulares. Esto es equivalente a una distribución triangular y debe ser expresada en cada escalón de fuerza como un valor relativo:

$$w_4 = \frac{1}{\sqrt{6}} \frac{r}{F} \qquad (2.15.)$$

### 2.3.2.5.   Incertidumbre de fluencia (creep), $w_5$

Este componente de incertidumbre es debido a la posibilidad de que la deformación del instrumento puede ser influenciada por su historial de carga a un corto plazo previo,

$$w_5 = \frac{c}{100\sqrt{3}} \qquad (2.16.)$$

donde $c$ es el error relativo de fluencia (creep) del instrumento.

Si el ensayo de fluencia no se realiza durante la calibración, esta contribución de incertidumbre será reemplazada por la contribución debida a la reversibilidad dividida por un factor de tres.

### 2.3.2.6.   Incertidumbre del cero, $w_6$

Este componente de incertidumbre es debido a la posibilidad de que la salida para carga cero del instrumento podría variar entre series de medida, por lo tanto, es una función del tiempo transcurrido a fuerza cero.

Una medida de esta variación es el error de cero $f_0$ así este efecto se puede estimar como:

$$w_6 = \frac{f_0}{100} \qquad (2.17.)$$

Donde siempre se estima $f_0$ considerando el mayor efecto.

### 2.3.2.7.   Incertidumbre por temperatura, $w_7$

Esta contribución es debida a la variación de temperatura a lo largo de toda la calibración, junto con la incertidumbre en la medición de este intervalo de temperatura de calibración.

La sensibilidad del instrumento de medida de fuerza a temperatura debe ser determinada, ya sea por pruebas experimentales o por las especificaciones del fabricante.

Este componente toma el mismo valor en cada escalón de fuerza y, expresado como un valor negativo, es igual a,

$$w_7 = K \frac{\Delta T}{2} \frac{1}{\sqrt{3}}$$  (2.18.)

donde $K$ es el coeficiente de temperatura del instrumento, en $^oC^{-1}$, y $\Delta T$ es el intervalo de temperatura de calibración, teniendo en cuenta la incertidumbre en la medición de temperatura.

En general un valor típico para K es 0.00027 $^o\,C^{-1}$, pero lo ideal es obtenerlos de las especificaciones del fabricante.

### 2.3.2.8. *Incertidumbre por interpolación,* $w_8$

Este componente de incertidumbre solamente se considera para instrumentos clasificados para interpolación, ya que una ecuación por interpolación no es aplicable a instrumentos clasificados sólo para fuerzas específicas.

Se estima la componente en cada fuerza de calibración como la diferencia entre la deformación media medida, $\overline{X}_r$, y el valor calculado de la ecuación de interpolación, $\overline{X}_a$, expresado como valor relativo:

$$w_8 = \left| \frac{X_r - \overline{X}_a}{\overline{X}_r} \right|$$  (2.19.)

### 2.3.2.9. *Incertidumbre típica combinada e incertidumbre expandida*

De estas incertidumbres relativas típicas combinadas, obtenemos la incertidumbre expandida y de esta manera se determinan los coeficientes de un ajuste por mínimos cuadrados para estos valores, que permita dar un valor de incertidumbre expandida U para cualquier fuerza dentro del intervalo de calibración.

La forma de la línea de ajuste (es decir, lineal, polinomial, exponencial) dependerá de los resultados de la calibración. Si se obtienen valores que son significativamente más pequeños que los valores calculados, en cualquier parte del alcance de fuerza de calibración, se debe aplicar un ajuste más conservador o debe especificarse un valor mínimo para la incertidumbre en las partes relevantes del alcance de fuerza.

## 2.4. INTRODUCCIÓN A LOS SISTEMAS NEUMÁTICOS

La neumática ha jugado desde hace mucho tiempo un papel importante como una tecnología en la realización de trabajos mecánicos. También se utiliza en el desarrollo de soluciones de automatización generalmente. La neumática se puede definir, como el conjunto de tecnologías que usan un gas como medio para transmitir energía. Usualmente este gas suele ser aire o nitrógeno.

El aire es una mezcla de gases y tiene la siguiente composición:

- Aproximadamente 78 Vol. % de nitrógeno.
- Aproximadamente 21 Vol. % de oxígeno.

El aire contiene, además trazas de dióxido de carbono, argón, hidrogeno, neón, criptón, xenón y helio(H y D 1997).

### 2.4.1. Presión

La presión se produce en un fluido cuando se somete a una fuerza. La presión en un fluido, se puede definir como la fuerza que actúa por unidad de área, como se puede observar en la siguiente Ecuación 2.20:

$$P = \frac{F}{A} \tag{2.20}$$

Donde:

$P$ = Presión [$Pa$].

$F$ = Fuerza [$Kgf$].

$A$ = Área [$m^2$].

Existen tres distintas formas para realizar la lectura una lectura de presión. Casi todos los transductores lecturan la diferencia de presión entre dos puertos de entrada. Esta forma de medición es conocida como lectura de presión diferencial.

La segunda forma, es cuando uno de los puertos de entrada esta abierto a la atmosfera, conocida como presión manométrica, es la diferencia entre la presión absoluta y atmosférica.

La tercera forma, es cuando uno de lo puertos de entrada esta conectado al vacío, conocida como presión absoluta, nos permite realizar la lectura de la presión con respecto al vacío (Rapp 2016).

### 2.4.1.1. Presión atmosférica

La presión en la superficie terrestre es denominada presión atmosférica. Esta presión también es denominada presión de referencia. La presión atmosférica no es constante. Su valor cambia en función a la ubicación geográfica y a las condiciones meteorológicas.

### 2.4.2. Efectos de la compresibilidad

Las propiedades de los gases reales difieren de las del gas ideal cuyo comportamiento sigue exactamente la Ecuación del gas:

$$PV = nRT \qquad \qquad 2.21$$

Donde:

$P$ = Presión [$Pa$].

$V$ = Volumen [$m^3$].

$n$ = Moles [mol].

$R$ = Constante universal del gas.

$T$ = Temperatura [$K$].

Para la mayoría de los propósitos prácticos, esta ecuación es adecuada para cálculos de ingeniería. En donde a veces es necesario tener en cuenta el efecto de la compresibilidad del gas (Parr 2006). Esto se hace posible reemplazando la ecuación del gas por la ecuación 2.22:

$$PV = ZnRT \qquad \qquad 2.22$$

Donde:

$Z$ = factor de compresibilidad.

Las tablas de compresibilidad están disponibles para varios gases. Estos se derivan experimentalmente y deben usarse si están disponibles. Alternativamente, se puede utilizar un gráfico generalizado (Parr 2006).

.

### 2.4.3. Leyes de los gases

Las leyes de los gases se basan en el comportamiento de gases perfectos o mezclas de gases perfectos.

Un gas ideal son aquellos gases cuyas moléculas no interactúan entre sí y se mueven aleatoriamente. En condiciones normales y en condiciones estándar, la mayoría de los gases presentan comportamiento de gases ideales (Barber 1997).

Las leyes de los gases son:

### 2.4.3.1.  Ley de Boyle

Establece que el volumen ($V$) de un gas, a temperatura constante varía inversamente a la presión ($P$). Se representa con las siguientes ecuaciones (Parr 2006):

$$\frac{V_2}{V_1} = \frac{P_1}{P_2} \qquad\qquad 2.23$$

$$P_1 V_1 = P_2 V_2 \qquad\qquad 2.24$$

### 2.4.3.2.  Ley de Charles

Establece que el volumen de un gas, a presión constante, varía directamente con la temperatura absoluta ($T$)

$$\frac{V_1}{V_2} = \frac{T_1}{T_2} \qquad\qquad 2.25$$

### 2.4.3.3.  Ley de Amonton

Establece que la presión de un gas, a volumen constante, varía directamente con la temperatura absoluta.

$$\frac{P_1}{P_2} = \frac{T_1}{T_2} \qquad\qquad 2.25$$

### 2.4.3.4.  Ley de Avogadro

Establece que los volúmenes iguales de todos los gases bajo la misma condición de presión y temperatura contienen el mismo número de moléculas. Dado que un mol de una sustancia contiene el mismo número de moléculas, el volumen molar de todos los gases es el mismo. El número de moléculas en un mol es 6.02257 x 10 23 (Parr 2006).

## 2.5. CONCLUSIONES DEL CAPÍTULO II

En el presente capítulo se presentan los fundamentos teóricos relacionados con el sensor de fuerza neumático. Los temas que se abordaron son las especificaciones técnicas de los sensores que describen las especificaciones generales de los sensores, el estado del arte de los sensores de fuerza donde se abordaron los diferentes tipos de sensores de fuerza que actualmente son utilizados en diversas áreas, se presentaron los diferentes cálculos de errores e incertidumbres utilizados para la calibración y clasificación del sensor de fuerza según la norma UNE-EN ISO 376 y finalizando se introdujo a los sistemas neumáticos. En el siguiente capítulo, se abordará el modelo matemático del sensor de fuerza neumático y su simulación con el software *Matlab/SolidWorks.*

# CAPÍTULO III

# MODELO MATEMÁTICO Y SIMULACIÓN

Al realizar el análisis y diseño de cualquier sistema, es importante la creación previa de un modelo matemático dinámico del sistema. El desarrollo del modelo matemático probablemente sea una de las fases más cruciales al diseñar cualquier sistema. Un buen modelo matemático, provee al diseñador toda la información que necesita sobre la dinámica del sistema en cuestión de cómo se comportará en la aplicación deseada (Blagojević y Stojiljković 2007).

En el presente capítulo se desarrolla el modelo matemático de un sensor de fuerza neumático y su respectiva simulación, la cual permite obtener los parámetros necesarios para el dimensionamiento del sensor y la validación del sistema para así verificar el correcto funcionamiento.

Para la modelación matemática del sistema se utilizó el software matemático *Matlab* (*Matrix Laboratory*) en conjunto con *Simulink*.

## 3.1. MODELACIÓN MATEMÁTICA DEL SISTEMA

Existen muchas maneras para describir el funcionamiento y comportamiento de diferentes fenómenos y/o sistemas. Se puede usar desde palabras, dibujos o bocetos, modelos físicos, programas de computadora o ecuaciones matemáticas. En otras palabras, la actividad de la modelación se puede realizar en diversos lenguajes y en ocasiones simultáneamente (Kulakowski, Gardner y Shearer 2007).

Como se utiliza el lenguaje de las matemáticas para realizar los modelos, estos se llaman modelos matemáticos.

Al hablar de la definición de modelos matemáticos, se quiere saber cómo generar representaciones, hacer, usar, validar modelos matemáticos y saber cuándo su uso se encuentra limitado.

Como el modelado matemático de diferentes sistemas, es esencial tanto para la ingeniería y la ciencia, los ingenieros y los científicos tienen razones prácticas para realizar el modelado matemático (Kulakowski, Gardner y Shearer 2007).

### 3.1.1. Ecuaciones

#### 3.1.1.1. Ecuación de la Continuidad

La ecuación de la continuidad consiste en el desarrollo y la síntesis del principio de la conservación de la masa, es una ecuación útil para el análisis en sistemas neumáticos.

Para el control del volumen donde se encuentra un sólo caudal de entrada y un solo caudal de salida, el principio de conservación de la masa establece que el volumen no cambie en el tiempo, el caudal de entrada debe ser igual al caudal de salida. En esta situación podemos expresar la ecuación de la continuidad en la ecuación (3.1):

$$m_{entrada} = m_{salida} \qquad (3.1.)$$

donde:

$m$ = masa[kg];

Para el control de volumen con múltiples caudales de entrada y salida, el principio de conservación de la masa requiere que la suma de los caudales másicos de entrada en el control de volumen sea igual a la suma de caudales másicos de salida en el control de volumen (Rapp 2016).

La ecuación de continuidad para esta situación más general puede expresarse en la Ecuación 3.2.

$$\sum m_{entrada} = \sum m_{salida} \qquad (3.2.)$$

Una aplicación de la ecuación de la continuidad es la de determinar el cambio de la velocidad del fluido cuando el diámetro de una tubería de expande o contrae.

La aplicación de dicha ecuación en el sensor de fuerza neumático, es la de representar la dinámica de la presión del aire comprimido dentro del compartimiento interno del sensor. El sensor al tener una estructura mecánica similar a un cilindro neumático, se realizó una analogía entre ambos, al aplicar la ecuación de continuidad en el modelo matemático del sensor (Doll, Neumann y Sawodny 2015).

Para la modelación del sistema se consideró que el aire tiene las propiedades de un gas ideal, por consiguiente, se realiza la suposición de que la relación de presión y volumen, ocupan una temperatura constante. La ecuación dinámica de la presión dentro la cámara interna del sensor de fuerza neumático se basó en el modelo de un cilindro neumático de doble efecto como Ledezma (2012), como se puede observar en la Ecuación 3.3:

$$\dot{p_a} = \frac{RT_0 qm_a - p_a A_a \dot{x}}{A_a x + V_{di}}$$
(3.3)

donde:

$p_a$ = presión de la cámara interna del sensor de fuerza neumático $[Pa]$;

$x$ = posición del pistón del sensor de fuerza neumático $[m]$;

$\dot{x}$ = velocidad del pistón $\left[\dfrac{m}{s}\right]$;

$R$ = constante de gas del aire $\left[\dfrac{J}{kgK}\right]$;

$T_0$ = constante de temperatura del aire estándar $[K]$;

$A_a$ = constante del área útil de la cara del pistón $[m^2]$;

$qm_a$ = caudal másico que entra a la cámara interna $[kg/s]$;

$V_{di}$ = constante del volumen muerto de la cámara interna $[m^3]$.

Tomando en cuenta las siguientes diferencias entre el sensor de fuerza neumático y el cilindro de doble efecto del modelo anterior:

- El compartimiento interno del sensor no tiene caudales de entrada y salida a diferencia del cilindro neumático, por lo tanto $qm_a$ es igual a 0 en el sensor.

- El volumen muerto de la cámara interna ($V_{di}$) en el sensor de fuerza es el volumen que une el compartimiento interno con el sensor de presión.

Se realizo una adaptación del modelo matemático de la Ecuación 3.3 para su aplicación en el sensor de fuerza neumático. Se puede observar en la siguiente ecuación:

$$\dot{p_a} = \frac{p_a A_a \dot{x}}{A_a x + V_{di}} \tag{3.4.}$$

En la Ecuación 3.4., se puede observar que la de presión ( $p_a$ ) varía en función a la de posición ( $x$ ). Lo cual expone la dinámica de la variación de presión dentro de la cámara interna del sensor, ante la variación de posición del pistón como resultado de una fuerza externa o, de entrada.

Por otro lado, igual se puede observar la influencia de la constante de Volumen muerto del compartimiento interno del sensor ( $V_{di}$ ), el cual en función a su valor, sea mayor o menor, tiene influencia en la variación de la presión ( $p_a$ ). Lo que demuestra los efectos de la propiedad de compresibilidad del aire dentro del compartimiento interno del sensor.

### 3.1.1.2.    *Ecuación del Movimiento*

La ecuación de movimiento que representa la dinámica del sensor de fuerza neumático, esta derivada de la aplicación de las leyes de Newton, y se puede expresar como en la siguiente Ecuación 3.4:

$$m\ddot{x} = F_p - F_{fr}(dx) - F_E \tag{3.5.}$$

Siendo $m$ la carga [$kg$], y por el lado derecho consiste en cuatro fuerzas: la fuerza neumática $F_p$ [$N$], la fuerza de fricción $F_{fr}$ [$N$] y la fuerza de entrada o externa en el sensor $F_E$ [$N$].

La fuerza neumática está calculada por la fuerza generada por la presión de la cámara interna y de su correspondiente área en el pistón, lo cual puede expresarse en la siguiente Ecuación 3.5.:

$$F_p = A_a p_a \tag{3.6.}$$

La fuerza de gravedad es calculada en función a la masa $m$ [$kg$] y el ángulo $\alpha$ , como se expresa en la siguiente Ecuación 3.7.

$$F_g = mg \cdot \sin(\alpha) \tag{3.7.}$$

Finalizando, la fuerza de fricción es una combinación de la fricción de Coulomb $f_{cf}$ [N] y la fricción viscosa $f_{vf}$ [N], como se puede expresar en la siguiente Ecuación 3.7 (Doll, Neumann y Sawodny 2015):

$$F_{fr}(dx) = f_{cf} \operatorname{sgn}(dx) + f_{vf} dx \tag{3.8.}$$

En donde la fricción de coulomb es una fuerza de magnitud constante que actúa en dirección opuesta al movimiento y la fricción viscosa es la fuerza de magnitud proporcional a la velocidad en sentido contrario.

En la Figura 3.1 podemos observar un Diagrama de cuerpo libre en el sensor de fuerza neumático.

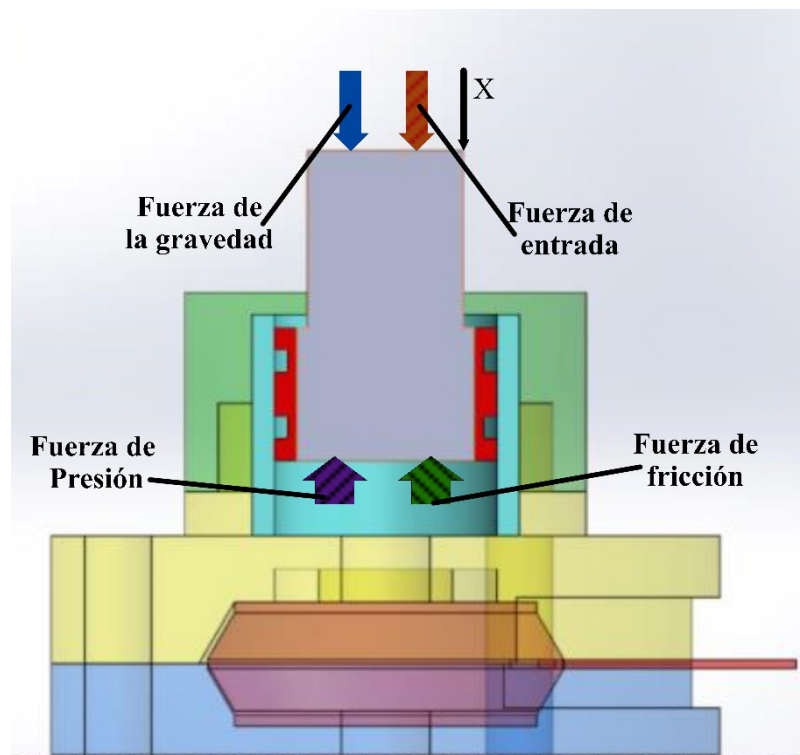

**Figura 3.1**. Diagrama de cuerpo libre del sensor de fuerza neumático.
**Fuente:** Elaboración propia.

### 3.1.2. *Sistema completo del sensor de fuerza neumático*

Para definir el modelo matemático completo del Sensor de Fuerza Neumático, se utilizó la ecuación de movimiento y la ecuación de continuidad, los cuales representan partes del sistema del sensor. El modelo matemático completo puede ser definido de la siguiente forma, con las subsiguientes variables (Doll, Neumann y Sawodny 2015):

$$x_1 = x \tag{3.9.}$$

$$x_2 = \dot{x} \tag{3.10.}$$

$$x_3 = p_a \tag{3.11.}$$

el sistema se establece como:

$$\dot{x}_1 = x_2 \tag{3.12.}$$

$$\dot{x}_2 = \frac{1}{m}(A_a(x_3) - mg\sin(\alpha) - F_{fr}(x_2)) \tag{3.13.}$$

$$\dot{x}_3 = \frac{RT_0 q m_a - x_3 A_a x_2}{A_a x_1 + V_{db}} \tag{3.14.}$$

Usando las variables de estado $x = [x_1, x_2, x_3]$ y la entrada $u$, representamos las ecuaciones del modelo matemático del sensor de fuerza neumático en el siguiente sistema de ecuaciones de estado:

$$\dot{x}(t) = f(x(t), u(t)) \tag{3.15}$$

## 3.2. SIMULACIÓN DEL MODELO MATEMÁTICO DEL SENSOR DE FUERZA NEUMÁTICO

### 3.2.1. Software de diseño y simulación

#### 3.2.1.1. Matlab y Simulink

Matlab es un lenguaje de programación similar a otros lenguajes conocidos como Java, C #, C++, etc., que viene con su propio IDE (el cual es el entorno de desarrollo integrado) y una serie de bibliotecas. Matlab es una abreviatura del término "Laboratorio de matrices", ya que inicialmente se denominó lenguaje de programación matricial (D. Hahn y T. Valentine 2019).

Matlab como una herramienta matemática contiene diferentes bibliotecas de funciones matemáticas que le permite realizar diferentes tipos de cálculos u operaciones. Se pueden crear modelos de datos, prototipos y simulación de datos. También se puede diseñar interfaces para los usuarios y otras aplicaciones de programación para facilitar el trabajo con Matlab (D. Hahn y T. Valentine 2019).

Usando Matlab se puede implementar, diseñar diferentes algoritmos y sistemas. Se puede cargar datos de diferentes fuentes, como archivos, bases de datos para analizar sus datos y visualizarlos en una gran variedad de gráficos para elegir. También facilita el trabajo con conjuntos de datos más grandes.

Para realizar el diseño del modelo matemático y la simulación del mismo se utilizó la herramienta de Matlab llamada *Simulink Toolbox*. Simulink se utiliza para modelar, simular y analizar sistemas dinámicos. Con lo cual es posible responder preguntas sobre un sistema, modelar el sistema y observar lo que sucede.

Con *Simulink* se puede crear modelos desde cero o modificar modelos existentes según los requerimientos. *Simulink* admite sistemas lineales y no lineales, modelados en tiempo continuo, tiempo muestreado o un híbrido de los dos. Los sistemas también pueden ser multitarea y; tener diferentes partes que se muestrean o actualizan en graficas. Es un entorno de programación gráfica, como se puede observar en el ejemplo de la figura 3.2. (D. Hahn y T. Valentine 2019).

### 3.2.2. *Parametrización inicial*

Para poder realizar el diseño del modelo matemático y la simulación del sistema en el software, es necesario definir los parámetros iniciales del sensor de fuerza neumático, como por ejemplo la presión de carga del compartimiento interno y sus dimensiones.

El objetivo principal del sensor de fuerza neumático es que pueda ser utilizado para aplicaciones de control de fuerza en manipuladores robóticos, por lo tanto, el rango de lectura de fuerza del sensor debe ser la adecuada para la manipulación de objetos.

En función al parámetro de fuerza máxima ejercida por una persona sobre un agarre de potencia cilíndrico con una mano, el cual es de 30 N, se considerará que la fuerza máxima de lectura en un sensor de fuerza neumático es 40 N, un tercio más de fuerza a la fuerza de agarre de una persona por seguridad (Pérez-González, Jurado-Tovar y Sancho-Bru 2011).

El segundo parámetro, que se tomó en cuenta en la parametrización del sistema, es la presión máxima del compartimiento interno del sensor de fuerza neumático. Cuando el sensor de fuerza neumático se encuentre contraído por el efecto de una fuerza entrante, la presión de la cámara interna subirá a un valor de presión máxima que es medida por el sensor de presión interna. Consecuentemente, la presión máxima del compartimiento interno del sensor de fuerza neumático es de 5 bar, por el motivo de que es la presión máxima de lectura del sensor de presión MPX5500D, el cual es utilizado para la lectura de la presión mencionada.

Posteriormente, se calcula el parámetro del diámetro del pistón del sensor de fuerza neumático en función a los dos parámetros anteriores, la presión máxima del compartimiento interno y la fuerza máxima de lectura del sensor de fuerza. Para realizar este cálculo se maneja la ecuación de relación Presión-Fuerza para consecuentemente despejar el diámetro del pistón.(Lab-Volt 1999) máxima de lectura del sensor de presión utilizado (MPX5500D, Anexo 2) es de 5 bar,

$$F = P\pi\left(\frac{d}{2}\right)^2 \tag{3.16.}$$

Donde:

$F$ = Fuerza máxima medida por el sensor de fuerza neumático [N];

$P$ = Presión máxima del compartimiento interno del sensor de fuerza neumático [Pa];

$d$ = Diámetro del pistón del sensor de fuerza neumático[m].

Con los parámetros de una fuerza máxima de 40 $N$ y una presión máxima de 500 kPa, se obtiene que el diámetro del pistón del sensor de fuerza neumático es de 10.09 mm, aproximándolo a unos 10 mm.

El siguiente parámetro que se toma en cuenta fue la distancia de recorrido del pistón en el sensor de fuerza neumático, el mismo es obtenido en función al largo del vástago utilizado en el sensor. Como se utiliza una válvula *Schrader* modificada como vástago, el mismo que cuenta con una largo de 8 mm libres de recorrido, fue designado un largo de 4 mm de recorrido del pistón y de esta forma obtener otros 4 mm del vástago que sirven como punto de presión para las fuerzas de entrada al sensor de fuerza neumático.

### 3.2.3. Diseño del modelo matemático mediante el software Matlab/Simulink

En cuanto a la simulación del modelo matemático del Sensor de Fuerza Neumático, se realiza en la herramienta de Matlab, llamada Simulink, la misma cuenta con las herramientas necesarias para dicha tarea.

El modelo matemático del sistema, en el software comprende la interacción en el tiempo de sus dos ecuaciones esenciales, la ecuación de continuidad que presenta la dinámica de la presión de la cámara interna del sensor y la ecuación de movimiento que presenta la dinámica del movimiento del pistón del sensor.

```matlab
clear all
clc
% ___________________________________________
%
%        Datos del Modelo del Sensor de Fuerza Neumatico
% ___________________________________________

% --------------------------
% Datos del Aire comprimido
% --------------------------

gamma = 1.4;                  % Relacion calor especifico
R     = 287;                  % Constante universal del gas [J/kgK]
patm  = 1.013e5;              % Presion Atmosferica [Pa]
p0    = patm;                 % Normal pressure [Pa]
T0    = 293.15;               % Temperatura normal [K]
T     = 293.15;               % Temperatura Ambiente (20 °C) [K]
psP   = 2.37e5;               % Presion de alimentacion del sensor[Pa]
rho   = p0/(R*T0);

% --------------------------
% Sensor de Fuerza Neumatico
% --------------------------

dh = 4e-3;                    % Diametro del volumen muerto [m]
Lh = 3e-3;                    % Largo del volumen muerto [m]
VAoP = (dh/2)^2*pi*Lh;        % Volumen muerto acoplado a la camara interna
MP  = 8e-3;                   % Masa del piston [kg]
Dp = 10e-3;                   % Diametro del Piston [m]
Au  = pi*((Dp^2)/4);         % Area efectiva [m^2]

% --------------------------
% Condiciones Iniciales
% --------------------------

pAoP = psP;                   %Presion inicial de la camara interna del sensor [Pa]
xoP = 4e-3;                   %Posicion inicial del piston neumatico [m]
B= 190;                       %Constante de Friccion Viscosa [N]
Frc= 10;                      %Constante de Friccion Coulomb [N]
```

**Figura 3.2.** Datos del Modelo del Sensor de Fuerza Neumático.
**Fuente:** Elaboración propia.

En segundo lugar, se realiza el modelo matemático del sensor de fuerza neumático en el software *Simulink*, la siguiente Figura 3.3. muestra el modelo general del sistema en bloques.

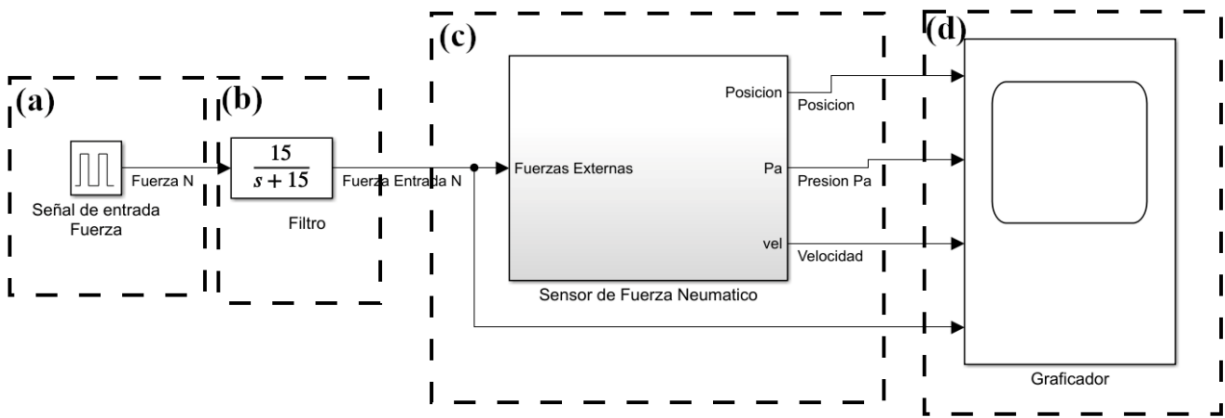

**Figura 3.3.** Vista general del modelo matemático del sensor de fuerza neumático en *Simulink*.
**Fuente:** Elaboración propia.

En la anterior Figura 3.3, se observa:

a) el bloque generador de pulsos que trabaja como señal de entrada de fuerza en el sistema, el parámetro de esta señal es una señal cuadrada,

b) un bloque de un sistema de primer grado que cumple la función de filtro paso-bajo para poder suavizar la señal, y de esta forma aproximarlo a la señal real,

c) el bloque del modelo matemático del sensor de fuerza neumático

d) y finalizando está el bloque *Scope* (Graficador), el cual recolecta los datos y los grafica en tablas.

En el bloque del modelo matemático del sensor de fuerza neumático se encuentra el siguiente sistema de bloques:

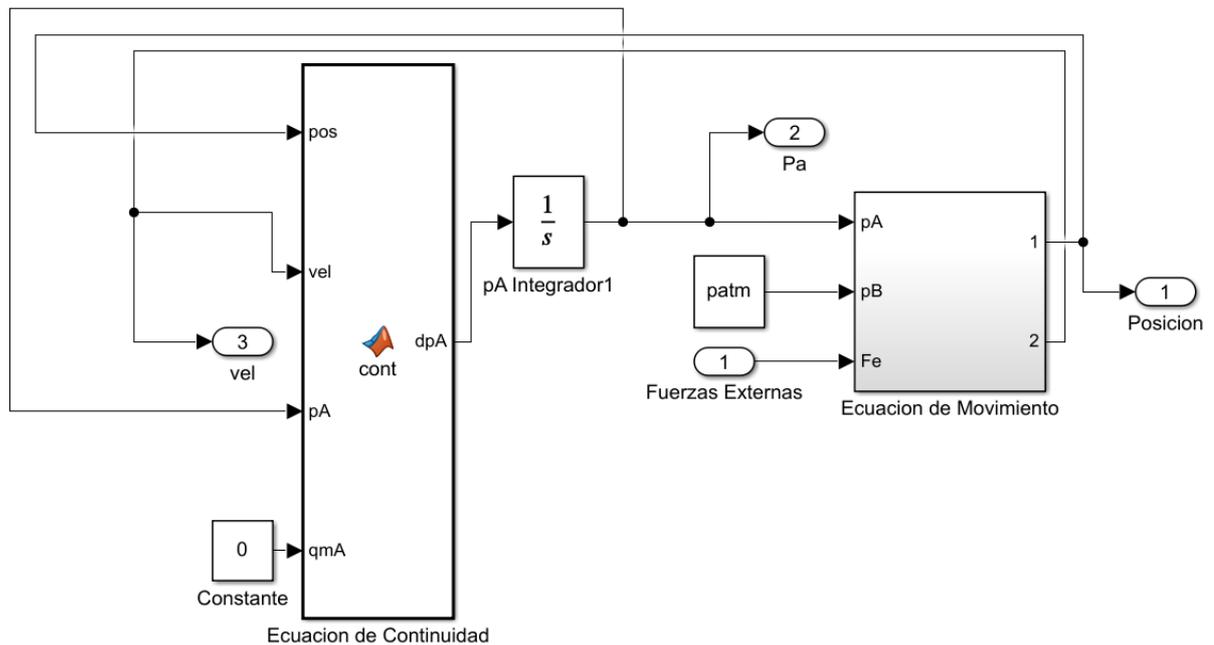

**Figura 3.4.** Bloque del modelo matemático del sensor de fuerza neumático.

**Fuente:** Elaboración propia.

En el diagrama de bloques de la Figura 3.4., se encuentran los siguientes bloques:

- el bloque de la ecuación de continuidad,
- el bloque de la ecuación de movimiento,
- la entrada de la variable de las fuerzas externas,
- la salida de las variables de posición, presión interna del sensor y velocidad,
- y finalizando encontramos los bloques de integración y derivación.

Comenzando con el bloque de la Ecuación de la continuidad, se encuentra la ecuación de la misma en Matlab, como se puede observar en la figura 3.5.:

```
function dpA = cont(pos,vel,Au,pA,qmA,R,T,gamma,VAoP)

dpA = (R*T*gamma* qmA - Au*gamma*vel* pA)/(Au*pos + VAoP);
```

**Figura 3.5.** Ecuación de la continuidad o dinámica de la presión del compartimiento interno del sensor de fuerza neumático.

**Fuente:** Elaboración propia.

En la anterior figura se encuentra la función de la continuidad, la cual representa una parte del modelo matemático del Sensor de Fuerza Neumático. Como entrada de la función se tuvo las siguientes variables: posición, velocidad, Área efectiva, presión interna, caudal a la cámara interna, coeficiente del gas, temperatura ambiente y volumen muerto continuo a la cámara interna. Como salida de la función se tiene la variable de la derivada de la presión de la cámara interna del sensor de fuerza neumático.

Continuando con el bloque de la Ecuación de movimiento, ésta representa al siguiente diagrama de bloques en la Figura 3.6.

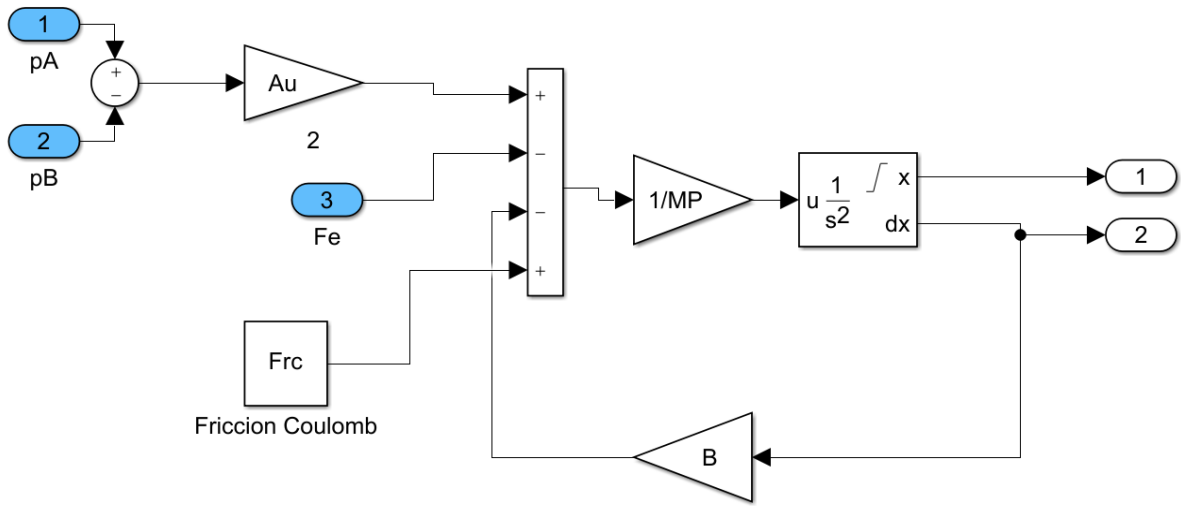

**Figura 3.6.** Ecuación de movimiento o dinámica del pistón del sensor de fuerza neumático.
**Fuente:** Elaboración propia.

La figura 3.6. es una representación de la ecuación de movimiento en la cual se encuentra como entrada las siguientes variables: presión de la cámara interna del sensor de fuerza neumático [Pa], presión $P_b$ [Pa] que representa presión atmosférica, fuerza externa [N], $MP$ [kg] que representa la masa del pistón y el sensor, $B$ que representa el coeficiente de fricción viscosa y $Frc$ representa la fricción de coulomb[N]. Finalizando, como salida del bloque se encuentra la variable de posición.

### 3.2.4. Resultados de la simulación

Una vez completa la implementación del modelo matemático del Sensor de Fuerza Neumático en Simulink, se procedió a la simulación del sistema por un tiempo de 50 segundos. Con la utilización del bloque *Scope* (Graficador) se obtuvo 4 gráficos, las cuales corresponden a las variables de:

- posición[m];
- presión de la cámara interna del sensor [Pa];

- velocidad[m/s] y;
- la fuerza de entrada en el sistema [N].

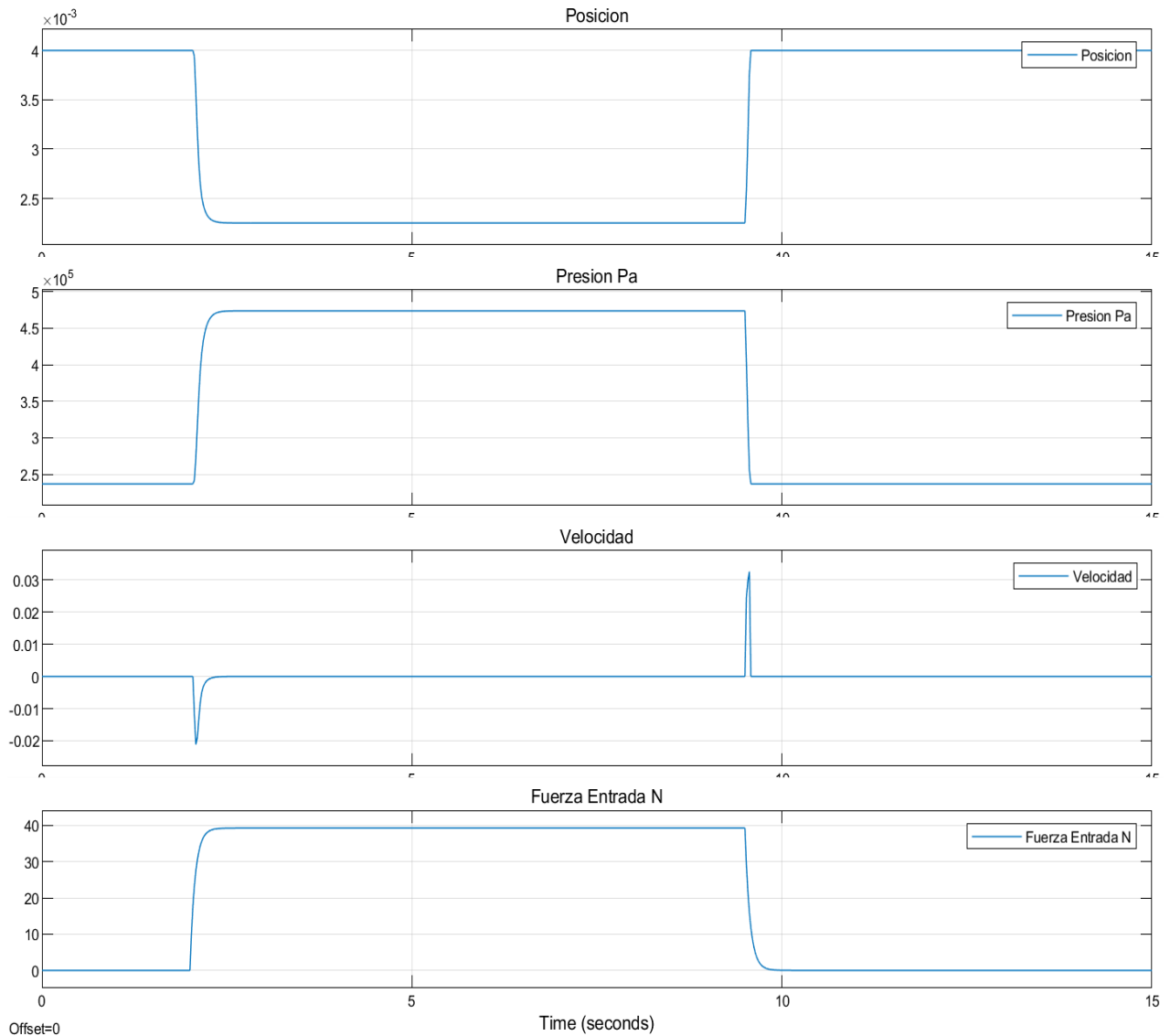

**Figura 3.7.** Resultados de la simulación del modelo matemático del sensor de fuerza neumático.
**Fuente:** Elaboración propia

Para terminar, se realiza un análisis de los resultados de la simulación del modelo matemático del Sensor de Fuerza Neumático el cual se observa en la Figura 3.7, y se encuentra una coherente respuesta para la elección de las dimensiones y componentes para el sensor de fuerza neumático. Se obtiene una respuesta de la variación de la presión interna del sensor en función a la fuerza de entrada en la simulación.

## 3.3. CONCLUSIONES DEL CAPÍTULO III

En el presente capítulo se desarrollaron las ecuaciones de continuidad y de movimiento, que forman parte del modelo matemático del sensor de fuerza neumático. Una vez el modelo matemático desarrollado, se procedió a utilizar el software Matlab/ Simulink, para poder simular el sistema. Con los resultados obtenidos de la simulación, en capítulos posteriores se realizará la validación del sistema.

Al realizar la simulación del sistema también se obtuvieron los parámetros necesarios para realizar el diseño mecánico del sensor de fuerza neumático, lo cual será el tema principal del siguiente capítulo.

# CAPÍTULO IV

# DISEÑO DEL SENSOR DE FUERZA NEUMATICO

## 4.1. ESQUEMA GENERAL DEL SENSOR DE FUERZA NEUMÁTICO

Para el desarrollo del sensor de fuerza neumático se consideran varios factores o parámetros, que son vitales para el dimensionamiento de sus diversos componentes. Se dimensiona el sensor de fuerza neumático en función de:

- o El volumen del compartimiento de aire comprimido calculado en el capítulo 3.
- o Las dimensiones de los componentes comerciales utilizados, como, por ejemplo: la válvula de aire *Schrader* y los *o-ring* utilizados.
- o El objetivo de utilizar el menor volumen dimensionalmente posible.

Como se puede observar en el Capítulo II, en la actualidad se utilizan una amplia variedad de métodos para la medición de la fuerza, de los cuales cada uno tiene sus ventajas y desventajas en función a la aplicación que se le proporcione.

La diferencia del sensor de fuerza neumático con el resto de los sensores se encuentra en el funcionamiento del mismo, se basa en el principio de la lectura de la variación de presión neumática del compartimiento de aire comprimido interno del sensor de fuerza neumático, de esta forma calcula indirectamente la fuerza aplicada en el sensor propuesto.

El sensor de fuerza neumático se compone de 6 componentes, en la Figura 4.1. se los puede observar y localizar su posición y funcionamiento en el sensor.

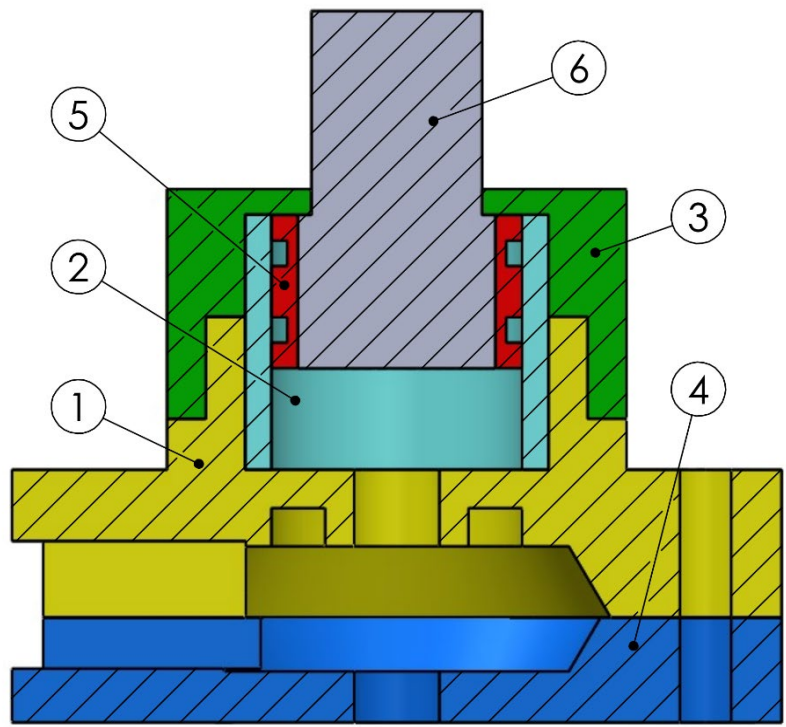

**Figura 4.2** Esquema estructural del sensor de fuerza neumático.

**Fuente:** Elaboración propia.

**Tabla 4.1** Descripción de componentes.

| Numero de Ítem | Nombre del Componente |
|---|---|
| 1 | Tapa superior del sensor. |
| 2 | Cilindro interno |
| 3 | Tapa superior del cilindro. |
| 4 | Tapa inferior del sensor. |
| 5 | Cabeza del pistón |
| 6 | Vástago del pistón (Válvula *Schrader* modificada) |

**Fuente**: Elaboración propia.

## 4.2. DIMENSIONAMIENTO DEL SENSOR DE FUERZA NEUMÁTICO

Cuando se realiza el diseño del sensor de fuerza neumático (Figura 4.2.), se utiliza dos criterios al diseñar las piezas del sensor. El primer criterio para el desarrollo del sensor de fuerza neumático es cumplir con las dimensiones del espacio de trabajo de los puntos de contacto de diferentes sistemas mecatrónicos con aplicaciones de control de fuerza, como, por ejemplo: Prótesis Biomecatrónicas, Brazos Robóticos Didácticos o Industriales, Robots Cuadrúpedos, etc. El segundo criterio es que todos los componentes a que se utilizan en el sensor de fuerza neumático tienen que poder ser adquiridos comercialmente.

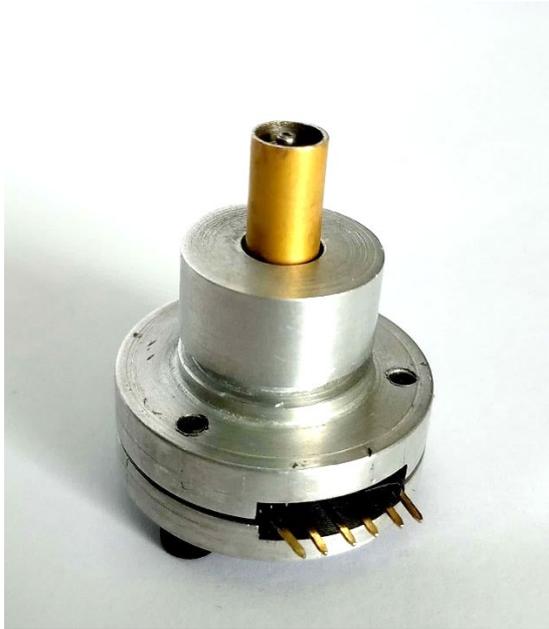

**Figura 4.2** Fotografía del prototipo terminado.
**Fuente:** Elaboración propia.

El sensor de fuerza neumático trabaja de la siguiente forma: la fuerza que será medida por el sensor de fuerza neumático es aplicada en la punta del componente 6 (Vástago del pistón) que está conectado al pistón interno dentro del sensor. Al aplicarse la fuerza en el vástago del pistón dentro el sensor, este comprime un compartimiento interno de aire comprimido en el sensor, como se observa en la Figura 4.3. b). La presión del compartimiento interno de aire comprimido dentro del sensor varía en función al rango de medida de fuerza que el sensor requiere medir. Al ser comprimido el compartimiento interno de aire, este aumenta su presión interna.

La mencionada variación de presión del compartimiento interno de aire comprimido del sensor de fuerza es recopilada por un sensor de presión incorporado en el sensor de fuerza neumático. En función a los datos recopilados de la variación de presión interna dentro del sensor de fuerza neumático, se obtuvo indirectamente las medidas de fuerza aplicada sobre el sensor de fuerza neumático.

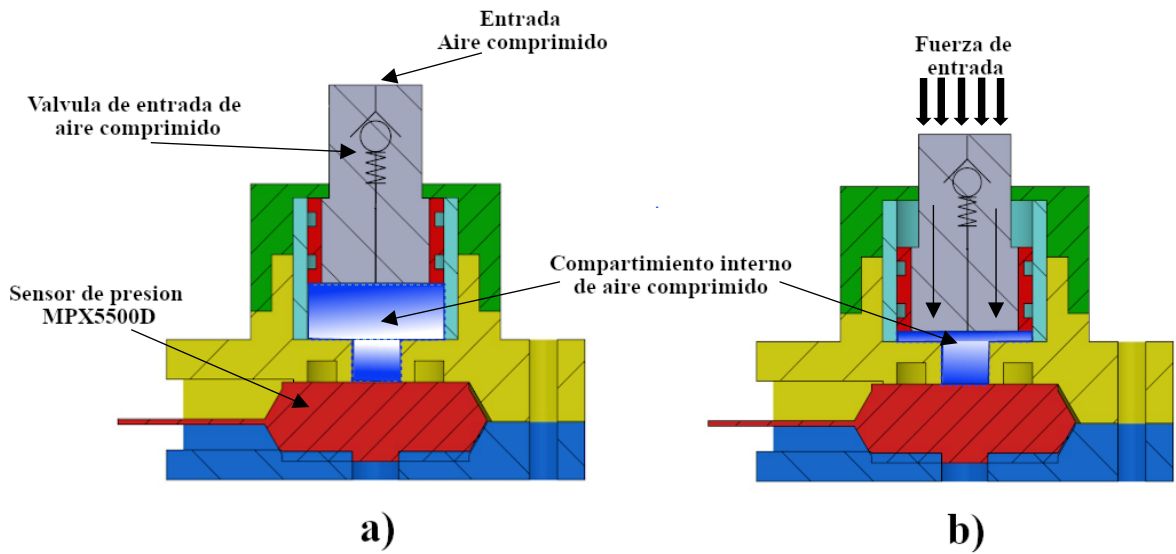

**Figura 4.3** a) Sensor de fuerza neumático expandido, en reposo. b) sensor de fuerza neumático comprimido, ante una entrada de fuerza.
**Fuente:** Elaboración propia.

El diseño del sensor está pensado para que trabaje de la forma planteada, conformado el mismo en 6 partes.

El componente numero 5 es la cabeza del pistón en el sensor de fuerza neumático (ver la figura 4.1.), su función es la de albergar los *o-ring* que sellaran el compartimento de aire comprimido cuando este en movimiento el pistón y la de acoplarse mediante rosca con el componente número 6, el vástago del pistón del sensor de fuerza neumático, el mismo que puede observarse en la figura 4.4. Se dimensiona las ranuras para los *o-rings*, en función de las tablas de diseño de *O-ring* de su fabricante, esto debido a que en el mercado existen medidas preestablecidas para estos componentes.

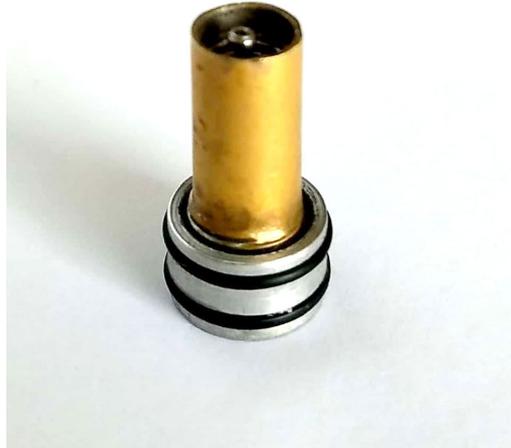

**Figura 4.4** Cabeza del pistón y el pistón del sensor de fuerza neumático.
**Fuente:** Elaboración propia.

El componente numero 6 es el vástago del pistón del sensor de fuerza neumático (ver figura 4.4), su función es de recibir la fuerza que será calculada por el sensor de fuerza neumático, al estar conectado con el componente número 5 que es la cabeza del pistón del sensor de fuerza neumático. Otra función de este componente es la de trabajar a su vez como la válvula de entrada de aire comprimido para el compartimiento interno del sensor de fuerza neumático. Para cumplir dicha tarea, el cuerpo del componente numero 6 es la modificación de una válvula de neumático de bicicleta tipo Schrader (Figura 4.5). La modificación realizada fue la realización de un corte en la sección en la cual se aloja la válvula y fue conservada de esta forma parte de la rosca de la válvula para luego ser acoplada con, el componente número 5, la cabeza del pistón.

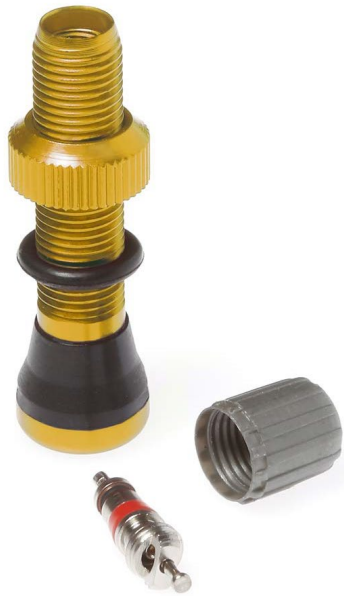

**Figura 4.5** Válvula de neumático de bicicleta tipo Schrader.
**Fuente:** (*Robot Check* 2020)

El componente numero 2 es el cilindro interno del sensor de fuerza neumático (ver figura 4.1). Su función del mismo es la de albergar el pistón del sensor de fuerza neumático que es el acoplamiento de los componentes número 5 y 6; y el compartimiento de aire comprimido dentro del sensor.

Los componentes número 1 y 3, son tapa superior del cilindro y la tapa superior del sensor de fuerza neumático. La función de estos dos componentes es la de encapsular y sellar al cilindro interno, el pistón y el compartimiento de aire comprimido. Estos dos componentes se unen mediante una rosca sellada con Teflón. A su vez, el componente numero 1 el cual es la tapa inferior, contiene el *O-ring* y la forma necesaria para acoplar el sensor de presión y sellarlo con el compartimiento interno de aire del Sensor de Fuerza Neumático.

El componente numero 4 es la tapa inferior del sensor de presión (ver figura 4.1). Su función es la de ajustar en su lugar al sensor de presión utilizado en el sensor de fuerza neumático. El componente numero 1 contiene 3 orificios dispuestos en triangulo para ajustarse mediante 3 pernos con el componente número 4 que es la tapa inferior del sensor de fuerza neumático.

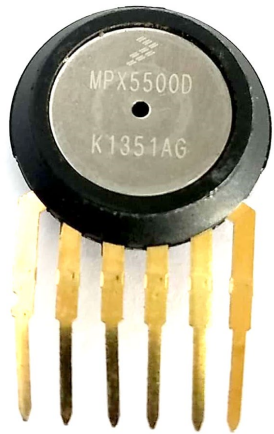

**Figura 4.6.** Sensor de presión MPX5500D
**Fuente:** Elaboración propia.

El sensor de presión MPX5500D (ANEXO 1) fue seleccionado para su uso en el sensor de fuerza neumático por que cumple con el rango de medición, las dimensiones y la sensibilidad requerida, en función al modelo matemático y simulación del sistema. Tiene la función de recopilar la variación de presión en el compartimiento de aire comprimido dentro del sensor de fuerza neumático. Para cumplir mencionada tarea, el sensor de presión se acopla a presión entre los componentes número 5 y 6, en donde tienen un canal para que puedan ajustarse correctamente los pines del sensor de presión, como se puede observar en la Figura 4.3.

### 4.2.1. Requerimientos

En el proceso de desarrollo del Sensor de Fuerza Neumático, se toma en cuenta una serie de requerimientos, para que el sensor cumpla con su trabajo óptimamente.

Los requerimientos del sensor de fuerza neumático son:

1. Utilizar un material ligero, sin comprometer la estructura del sensor.

2. El sensor de fuerza neumático tiene que tener las dimensiones óptimas para que pueda medir fuerza dentro de un rango de 0 a 30 N, el mismo rango de fuerza de agarre cilíndrico promedio de una mano humana según (Pérez-González, Jurado-Tovar y Sancho-Bru 2011).

3. Dimensionar el sensor de fuerza neumático en base a los siguientes puntos:

    • En función de los parámetros utilizados en el modelo matemático de la simulación del sistema del sensor de fuerza neumático.

- En base a las dimensiones de los componentes comerciales utilizados, como, por ejemplo: el sensor de presión MPX5500D, los pernos de la tapa del sensor y la válvula de neumático de bicicleta Tipo *Schrader*.
- La tecnología que se utiliza para la fabricación de los componentes del sensor de fuerza neumático tiene que poder ser utilizada localmente.
- Obtener las dimensiones más reducidas posibles en los componentes, en función al resto de los puntos.

### 4.2.2. Dimensionamiento

Cuando se habla de sensores de fuerza, se encuentran de diferentes formas y tamaños, por el motivo de que cada sensor de fuerza está especializado para una aplicación específica. Dependiendo de las especificaciones de: rango de medición, histéresis, tiempo de respuesta, entrada y salida del sensor, entre otras especificaciones, el sensor de fuerza varia su tipo , dimensiones y forma (Fraden 2010).

En el caso del sensor de fuerza neumático, se realiza el diseño del mismo en función a los requerimientos de la sección anterior. Se toma un especial cuidado al igualar el rango de fuerza de medición del sensor de fuerza neumático con el promedio de la fuerza máxima de agarre de una mano en forma de agarre de potencia cilíndrica (Pérez-González, Jurado-Tovar y Sancho-Bru 2011).

Al mismo tiempo, se busca poder facilitar su implementación en superficies de contacto de robots manipuladores. Por consiguiente, las dimensiones del sensor de fuerza neumático son reducidas en lo posible en la etapa de diseño.

Una de las restricciones que se toma en cuenta en el dimensionamiento del sensor de fuerza neumático, es que todas las paredes en los componentes del sensor mínimamente tienen 1 milímetro de grosor. Por el motivo, que los componentes al tener dimensiones reducidas, la fabricación manual en torno de los componentes del sensor se fuerza neumático aumentaría su complejidad.

Tomando en cuenta los puntos mencionados anteriormente, se realizó la modelación matemática del sistema para determinar las dimensiones del sensor. El modelo matemático del sensor de fuerza neumático puede ser relacionado con la ecuación de movimiento de un pistón neumático sin entradas de caudal.

### 4.3. SENSOR DE PRESION MPX5500D

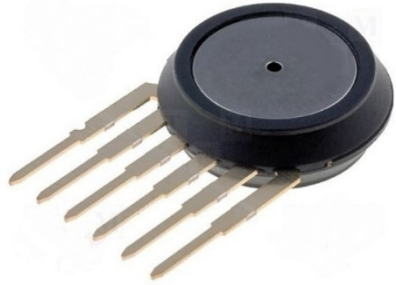

**Figura 4.7** Sensor de Presión MPX5500D.
**Fuente:** (Freescale 2009)

El Sensor de Fuerza Neumático, para poder realizar el cálculo de la variación de presión de su compartimiento interno de aire comprimido, utiliza al sensor de presión (MPX5500D) (Figura 3.5). El sensor de presión cuenta con su propio compartimiento dentro del sensor de fuerza neumático.

El sensor de presión (MPX5500D), es un sensor de presión de silicio monolítico diseñado para una amplia gama de aplicaciones, contiene un rango de medición de presión diferencial de 0 a 500 kPa, y proporciona una salida analógica precisa de 0.2 a 4.8 V. Todas sus especificaciones técnicas se encuentran en el ANEXO 2.

### 4.4. DIAGRAMA DE FLUJO DE FUNCIONAMIENTO

Un diagrama de flujo es una representación gráfica o simbólica de un proceso. Cada paso del proceso está representado por un símbolo diferente y contiene una breve descripción del paso del proceso. Los símbolos del diagrama de flujo están vinculados entre sí con flechas que muestran la dirección del flujo del proceso (Delgado y Amador 2014).

El siguiente diagrama de flujo representa el funcionamiento del sensor de fuerza neumático.

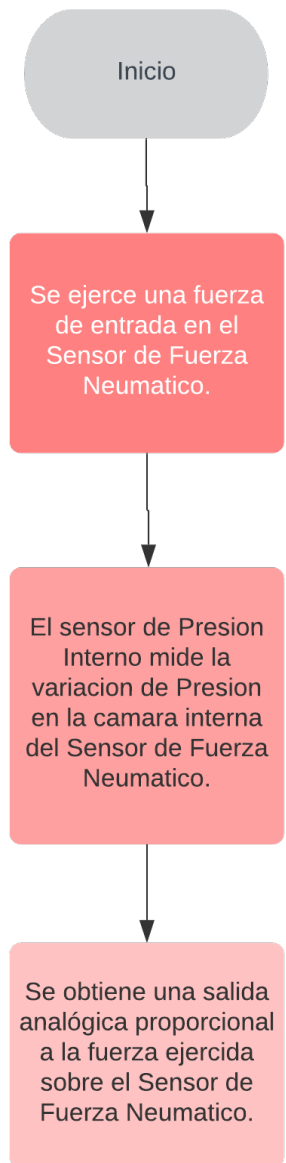

**Figura 4.8** Diagrama de flujo de funcionamiento del sensor de fuerza neumático.
**Fuente:** Elaboración Propia

## 4.5. CONCLUSIONES DEL CAPÍTULO IV

En el presente capítulo se propone el diseño mecánico del sensor de fuerza neumático, en función a los parámetros obtenidos de la simulación del modelo matemático del sistema desarrollado en el capítulo III. Después se describió el funcionamiento de cada uno de los 6 componentes del sensor de fuerza neumático.

Con el diseño mecánico del sensor finalizado, en el siguiente capítulo se desarrollarán los prototipos de los diseños avanzados.

# CAPÍTULO V

# PROTOTIPOS DE SENSORES DE FUERZA NEUMATICOS Y BANCOS DE PRUEBAS

En este capítulo se muestra el proceso de construcción de los elementos mecánicos e instalación de los componentes electrónicos, del sensor de fuerza neumático y el banco de pruebas para sensores de fuerza, incluyendo los pasos que se siguieron para llegar al prototipo final. A diferencia del Capítulo III Diseño del sensor de fuerza neumático en donde se estudia el diseño de funcionamiento del sensor de fuerza neumático, en el presente capítulo se avanzan los pasos que se tomaron para la construcción del prototipo final y la elección de materiales para su construcción.

## 5.1. CONSTRUCCIÓN DEL SENSOR DE FUERZA NEUMÁTICO

En el capítulo anterior, se avanzó el diseñó del sensor de fuerza neumático, lo que comprende el cómo es el proceso de funcionamiento mecatrónico y las funciones de los componentes en prototipo final del sensor de fuerza neumático. De este modo, aunque todo el diseño del sensor está en base a la información recopilada en el modelo matemático del Capítulo III. Todos los prototipos construidos en esta etapa fueron un camino que llevó al prototipo final del sensor de fuerza neumático descrito en el Capítulo IV.

En el presente capítulo se aclara cuáles fueron las características y los cambios que se realizaron en los distintos prototipos desarrollados para el sensor de fuerza neumático, así como también las experiencias, dificultades y observaciones en la construcción de los mismos.

### 5.1.1. Prototipos

#### 5.1.1.1. Prototipo 1

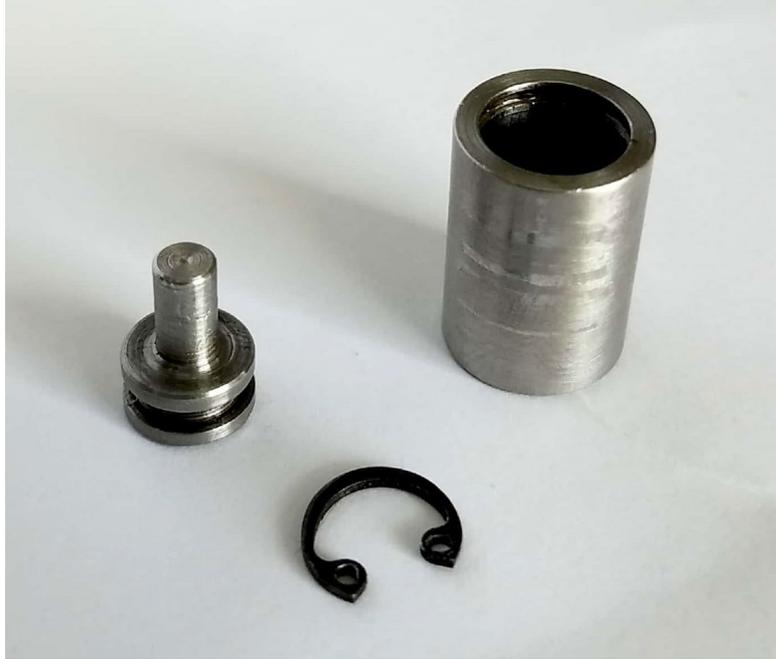

**Figura 5.1.** Fotografía de los componentes del primer prototipo de sensor de fuerza neumático.
**Fuente:** Elaboración propia.

El primer prototipo de sensor de fuerza se puede observar en la Figura 5.1, que es un prototipo sencillo de fabricar, contando con la menor cantidad de componentes en comparación al resto de prototipos, los cuales solo son dos. El pistón del sensor de fuerza neumático y la base del sensor de fuerza neumático, como se puede observar en la Figura 5.2.

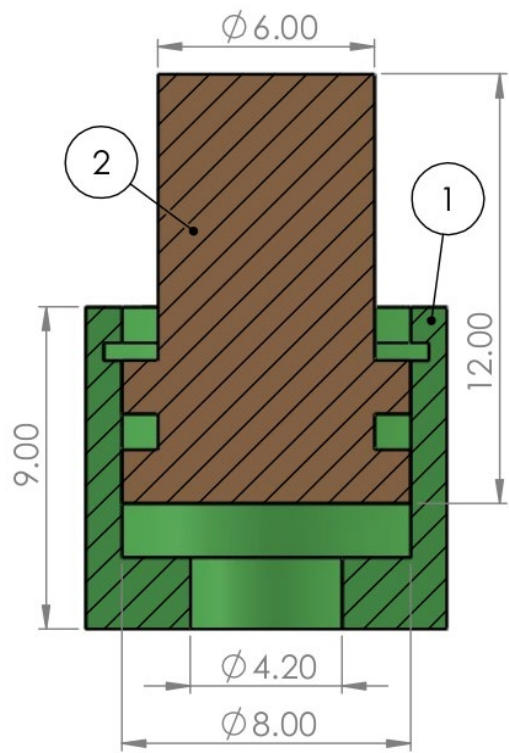

**Figura 5.2.** Vista de componentes del primer prototipo de sensor de fuerza neumático.
**Fuente:** Elaboración propia.

Una de las características del primer prototipo es el material de fabricación utilizado, este prototipo se fabricó con acero inoxidable ANSI 420, es el único de los prototipos que fue construido con acero inoxidable porque en los prototipos posteriores se utilizó aluminio 5052. En un principio se fabricó el prototipo con acero inoxidable debido a sus características mecánicas y por su propiedad de ser inoxidable.

Los motivos por el que se cambió de material a aluminio en prototipos posteriores son los siguientes:

- Debido a los bajos esfuerzos que se genera en la estructura, es posible utilizar el aluminio al no requerir una alta resistencia del material.
- Al utilizar aluminio se redujo el peso del sensor de fuerza neumático, a diferencia de haber utilizado acero inoxidable.
- El aluminio al tener la propiedad de ser más maquinable que el acero inoxidable, facilito la tarea de construcción de los prototipos posteriores.
- El aluminio es más económico que el acero inoxidable.

Varios motivos definieron el cambio de material, en especial, la maquinabilidad del material. El sensor de fuerza neumático al contar con un tamaño reducido en su diseño y debido al método de fabricación convencional utilizado, el cual se realizó en un torno, al utilizar aluminio se facilitó su fabricación, enfatizando en los componentes más complejos en los prototipos posteriores.

Uno de los cambios que se realizó en los siguientes prototipos fue la implementación de 2 *o-ring* en lugar de sólo un solo *o-ring* como en el primer prototipo. La razón de este cambio es debido a que el pistón al tener solo un punto de contacto mediante el *o-ring* con la cámara del pistón, éste al realizar el recorrido, logro desviarse de la posición perpendicular a la base. Una solución a este problema en los prototipos posteriores fue la implementación de un segundo *o-ring* en el pistón del sensor de fuerza neumático, lo cual trajo como beneficio un correcto desplazamiento sin desviaciones del pistón en su recorrido y un doble sello de la cámara interna de aire comprimido.

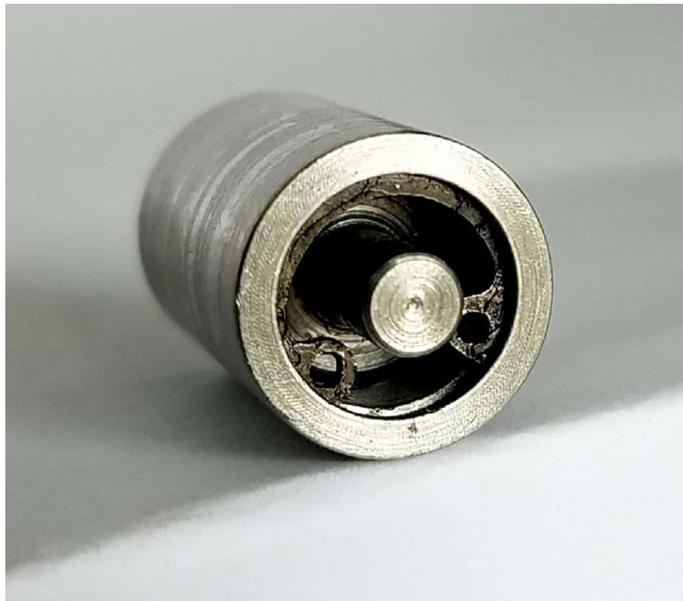

**Figura 5.3** Vista superior del primer prototipo de sensor de fuerza neumático.
**Fuente:** Elaboración propia

El primer prototipo, implementó una arandela de seguridad como tope límite para el pistón del sensor de fuerza neumático como se puede observar en la Figura 5.3, en cambio en los prototipos posteriores se cambió este sistema por una componente, tipo tapa, que cumpla la función de tope límite al pistón.

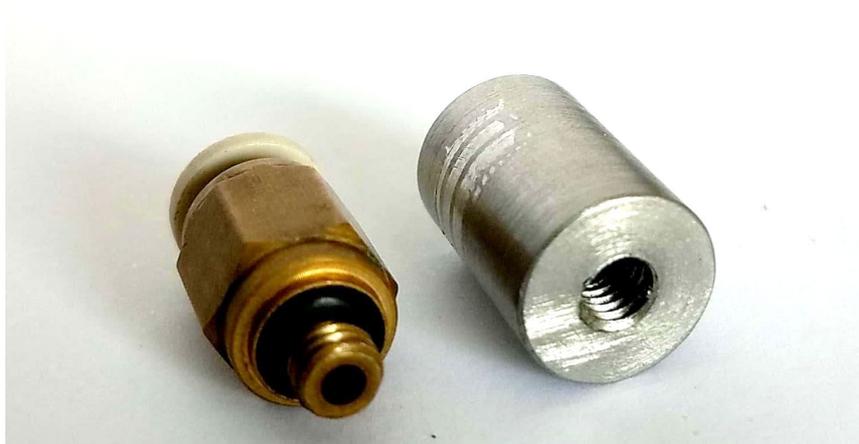

**Figura 5.4.** Vista inferior del primer prototipo del sensor de fuerza neumático.
**Fuente:** Elaboración propia

La entrada de aire comprimido para la cámara interna del primer prototipo se realizó mediante la utilización de una rosca M5 que se encuentra en la base del sensor y en la misma se acopló a un conector rápido M5 que se podría conectar a una manguera con aire comprimido como se observa en la figura 5.4.

### 5.1.1.2. Prototipo 2

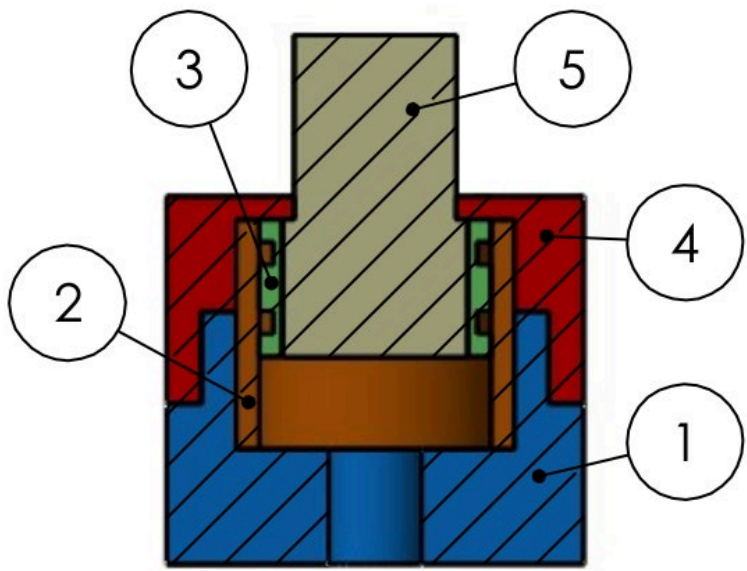

**Figura 5.5.** Vista de componentes del segundo prototipo del sensor de fuerza neumático.
**Fuente:** Elaboración propia.

A diferencia del primeo, el segundo prototipo del sensor de fuerza neumático cuenta con una mayor cantidad de componentes lo que lo convirtió en un prototipo más complejo y mejorado, gracias a lo aprendido en el primer prototipo.

El segundo prototipo de sensor de fuerza neumático fue fabricado en un torno mecánico y se usó como material el aluminio 5052. Un detalle por resaltar es que, al ser los componentes de tamaño reducido, con paredes de los componentes de hasta 1mm de grosor, lo ideal es fabricarlo con una maquina CNC, pero debido a la falta de disponibilidad local no se pudo fabricar mediante esta herramienta y se optó por su fabricación convencional.

El prototipo consta de 5 componentes, utilizando la Figura 5.5 como referencia en la numeración, los componentes del segundo prototipo del sensor de fuerza son los siguientes:

1. Tapa inferior.
2. Cilindro Interno.
3. Cabeza del pistón.
4. Tapa Superior.
5. El vástago (Válvula *Schrader*).

Una de las mejoras que se realizó a diferencia del primer prototipo fue el separar como único componente el cilindro interno sobre el cual se desliza el pistón del sensor de fuerza neumático, el cual comprende al componente 6 de la Figura 5.5. El motivo de este cambio fue facilitar la fabricación y el proceso de pulido del cilindro, ya que su parte interna debe contar con la menor fricción con el pistón del sensor posible.

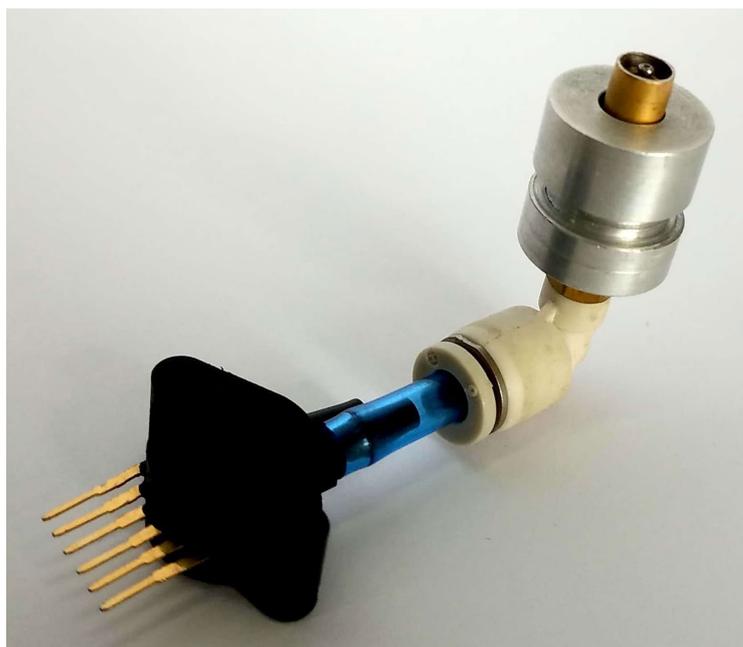

**Figura 5.6** Fotografía del segundo prototipo del sensor de fuerza neumático.
**Fuente:** Elaboración propia.

La segunda mejora destacable del segundo prototipo es la inclusión de una válvula Schrader modificada, para que cumpla la función de válvula de entrada de aire comprimido para la cámara interna del sensor de fuerza neumático y a la vez al ser acoplado mediante rosca con la cabeza del pistón, trabaja como el vástago del sensor. Una válvula *Schrader*, es un tipo de válvula utilizada para neumáticos de bicicletas. Este componente es el número 5 de la Figura 5.5.

El segundo prototipo, cuenta con dos componentes (la tapa inferior y la tapa superior), que tienen la función de sellar mediante una rosca los componentes internos del sensor, los cuales son la cabeza del pistón, el vástago y el cilindro interno. El motivo de esta decisión fue la de disminuir el desvió de los componentes en movimiento, y lograr un sellado de la cámara interna del sensor.

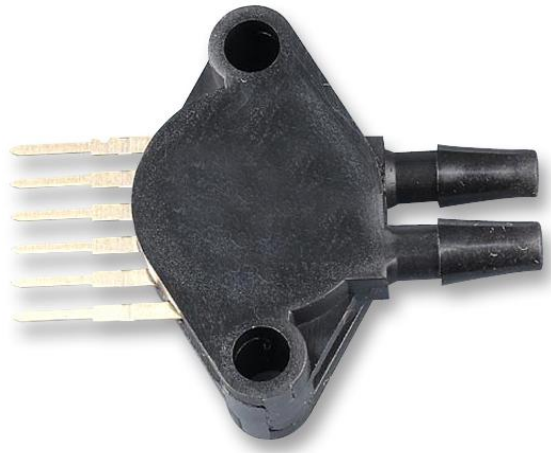

**Figura 5.7** Sensor de presión MPX5500DP.
**Fuente:** Elaboración propia.

El segundo prototipo al igual que el primero, cuenta con una rosca M5, que se utilizó para acoplar un racor que tenía la función de unir la cámara interna del sensor con el sensor de presión mediante un tubo neumático. El sensor de presión utilizado en los dos primeros prototipos es el MPX5500DP, Figura 5.7.

Una de las desventajas que se tuvo con esta configuración, fue que al tener al sensor de presión externamente del cuerpo del sensor de fuerza neumático, esto representaba una mayor cantidad de volumen muerto en la cámara interna del sensor lo cual afectaría la dinámica del sensor y consecuentemente las lecturas de fuerza. El aumento de este volumen muerto afecto directamente al rango de variación de presión de la cámara interna debido al módulo de compresibilidad del aire. Por este motivo se replanteó el prototipo y se llegó al prototipo final.

### 5.1.1.3.   Prototipo Final

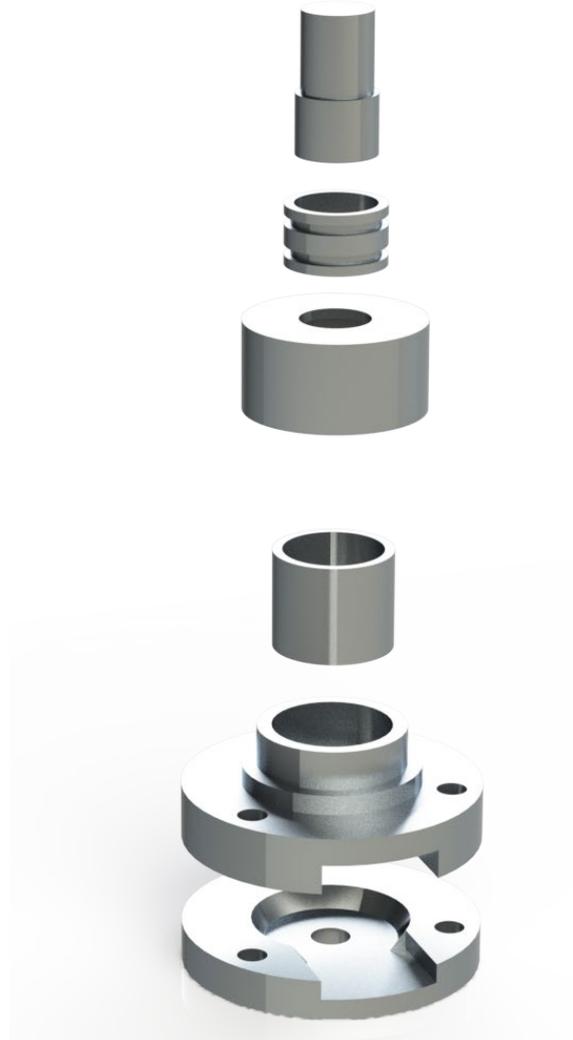

**Figura 5.8** Renderizado del modelo CAD del sensor de fuerza neumático.
**Fuente:** Elaboración propia.

Con el aprendizaje y experiencia obtenido de los prototipos anteriores, a través de una larga etapa de desarrollo y corrección. Se logró llegar hasta el prototipo final del sensor de fuerza neumático, el cual es un diseño que cumple con los límites y alcances planteados en el Capítulo I del presente documento.

El prototipo final del sensor de fuerza neumático cuenta en su estructura con 6 componentes individuales (Figura 5.10), como se expuso en el Capítulo IV.

A diferencia de los dos primeros prototipos, el prototipo final fue diseñado para que pueda albergar internamente el sensor de presión. Lo que permitió crear una conexión directa entre la cámara interna y el sensor de presión, disminuyendo el volumen muerto en la misma, esto fue beneficioso, al aumentar la variación de presión de la cámara interna al ser comprimida cuando el sensor de fuerza neumático este detectando una fuerza. Los componentes tapa inferior y la tapa superior del sensor, los cuales son los dos componentes inferiores en la Figura 5.8, son los que fueron diseñados específicamente para poder albergar al sensor de presión MPX5500D, el mismo que se observa en la Figura 3.5.

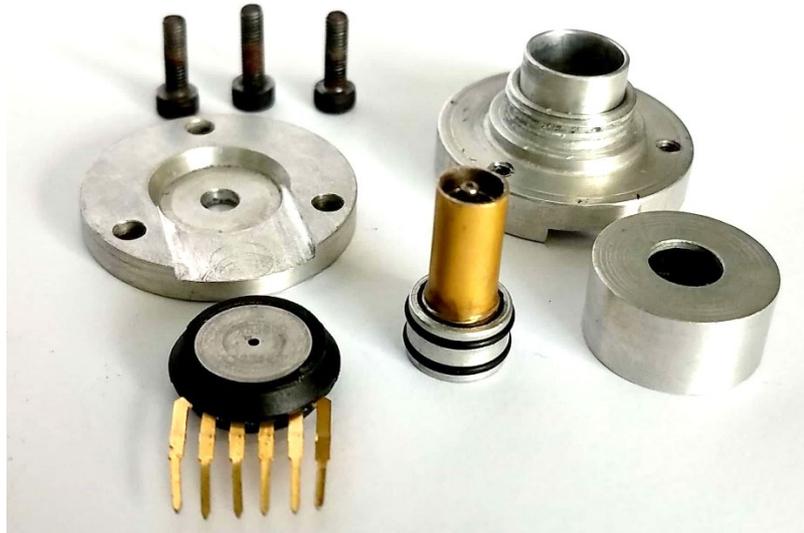

**Figura 5.9** Desglose de componentes del prototipo final del sensor de fuerza neumático.
**Fuente:** Elaboración propia.

Finalmente, una observación a destacar que se realizó sobre el prototipo final del sensor de fuerza neumático es que su tamaño reducido permite usar el sensor diferentes aplicaciones de control de fuerza, por ejemplo: en brazo robótico industrial, en prótesis de mano biomecatrónicas, en drones, entre varios.

Los planos del modelo CAD del prototipo final del sensor de fuerza neumático se encuentran en el ANEXO 1.

### 5.1.1.3.1.    Costos del prototipo final

El prototipo final del sensor de fuerza neumático tuvo un costo total de:

**Tabla 4.1** Costos del prototipo de sensor de fuerza neumático

| Número | Ítem | Cantidad | Costo Unitario [Bs] | Costo Total [Bs] |
|--------|------|----------|---------------------|------------------|
| 1 | Cilindro de aluminio 5052, 50x120mm | 1 | 80 | 80 |
| 2 | Sensor de presión MPX5500D | 1 | 243 | 243 |
| 3 | O-rings 10*0.5mm | 4 | 0.5 | 2 |
| 4 | Tornillos M3*10mm | 3 | 0.5 | 1.5 |
| 5 | Válvula *Schrader* | 1 | 15 | 15 |
| 6 | Costo de diseño del sensor de fuerza neumático (construcción de prototipos, horas de trabajo) | 1 | 1201 | 1021 |
| | | | **Costo total [Bs]** | 1362.5 |

**Fuente**: Elaboración propia.

## 5.2. CONSTRUCCIÓN Y MONTAJE DEL BANCO DE PRUEBAS DE FUERZA

Una vez desarrollado y construido, el sensor de fuerza neumático, esta apto para realizar las pruebas experimentales. Para llevar a cabo esta tarea, fue necesario utilizar un banco de pruebas de fuerza, ya que tiene la función de ser el instrumento para diagnosticar y estudiar el comportamiento del sensor desarrollado.

Al no contar con un banco de pruebas de fuerza localmente, se procedió con el trabajo de construir uno. En este caso, el desarrollo y construcción de un banco de pruebas tuvo varios prototipos, al igual que en el caso de la construcción del sensor de fuerza neumático.

Para la construcción del banco de pruebas se decidió que el procedimiento de las pruebas experimentales estaría basado en la normativa "Procedimiento ME-002 para la calibración de los instrumentos de medida de fuerza" del Centro Español de Metrología, el mismo que está basado en la normativa ISO 376:2011 "Estándar para la Calibración de Transductores de Fuerza". Se tuvo que desarrollar los prototipos de banco de prueba de fuerza, según los requerimientos establecidos en la normativa (Centro Español de Metrología 2019).

### 5.2.1.   Objetivos de un banco de pruebas de fuerza

Los objetivos de un banco de pruebas de fuerza son:

- Debe tener la capacidad de soportar cargas externas con la confiabilidad suficiente, para no poner en riesgo la vida del usuario.
- Tiene que tener la capacidad de ejercer una salida dentro del rango de fuerza del sensor a estudiar.
- Tiene que poder ser construido con las herramientas y materiales adquiribles localmente.

### 5.2.2. Prototipos

#### 5.2.2.1. Prototipo 1

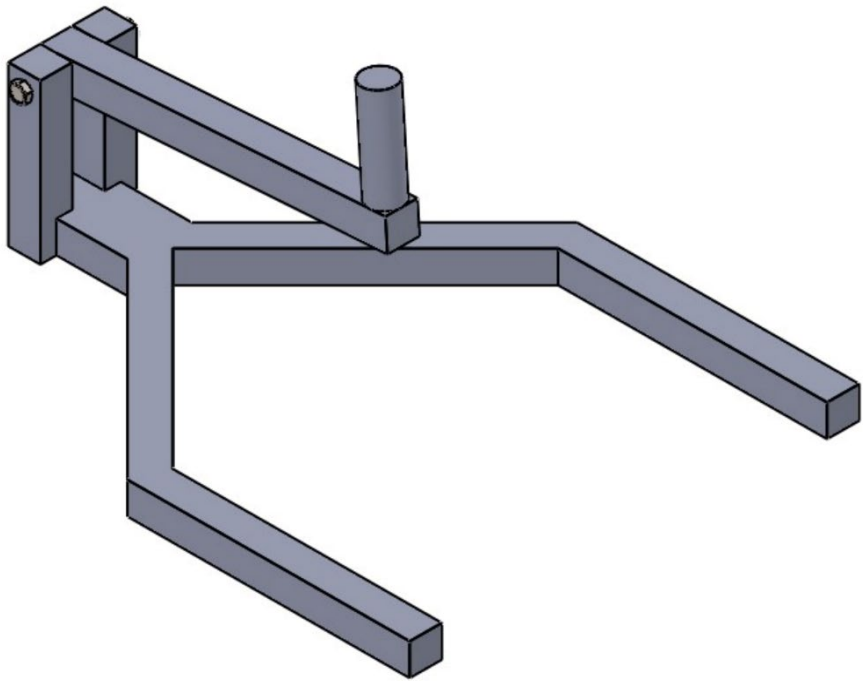

**Figura 5.10.** Modelo CAD del prototipo 1 del banco de pruebas de fuerza.
**Fuente:** Elaboración propia.

En primer lugar, al comenzar la etapa de fabricación del banco de pruebas de fuerza para realizar las pruebas experimentales se tuvo en mente trabajar con un sistema manual, el cual podría variar la fuerza que se ejerce contra el sensor variando la combinación de pesos en las masas conocidas que se acoplarían en el banco de pruebas.

De esta manera se procedió a diseñar y construir el primer prototipo de banco de pruebas de fuerza, el mismo que se puede observar en la Figura 5.10.

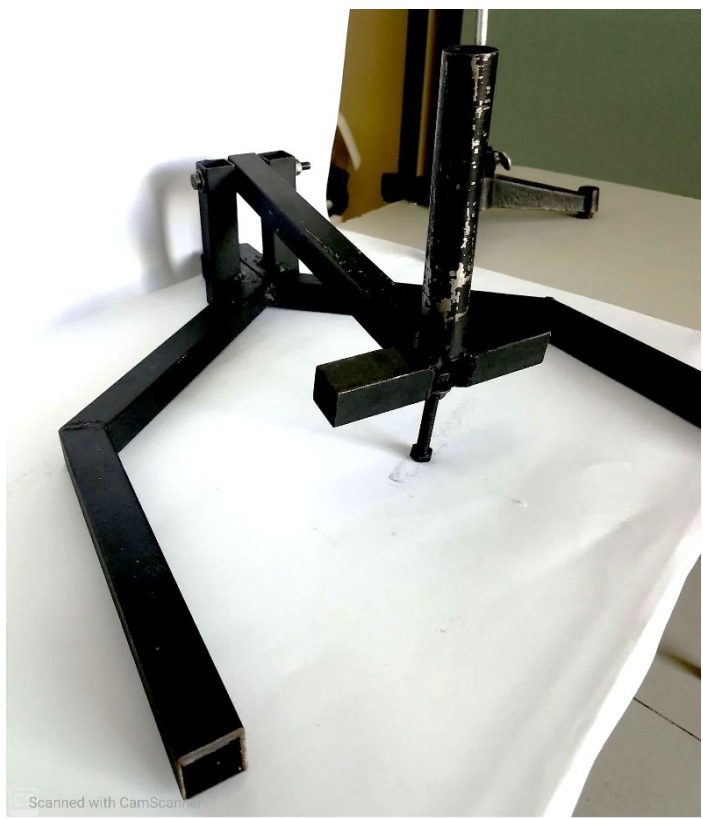

**Figura 5.11** Primer prototipo de banco de pruebas de fuerza.

**Fuente:** Elaboración propia.

El primer prototipo de Banco de Pruebas de Fuerza fue construido en su totalidad con perfiles y tubos de acero, por su buena maquinabilidad, lo que facilito su construcción, como se puede observar en la Figura 5.11.

El funcionamiento de este primer prototipo es el siguiente, el prototipo al contar con un brazo central que gira sobre un eje elevado, se lo utiliza para albergar una serie de diferentes combinaciones de pesas de mancuernas con diferentes tamaños. Este brazo del prototipo de banco de pruebas de fuerza ejerce una fuerza directa sobre el sensor de fuerza neumático que se ubicaba bajo el soporte de las pesas del brazo del prototipo.

Una de las observaciones que se realizó, es que el brazo del prototipo al ser el que ejerce la fuerza sobre el sensor, y el mismo tiene un movimiento rotatorio sobre un eje. Este generaba fuerzas axiales sobre el Sensor. Lo que perjudico la toma y precisión se los datos capturados del sensor de fuerza neumático.

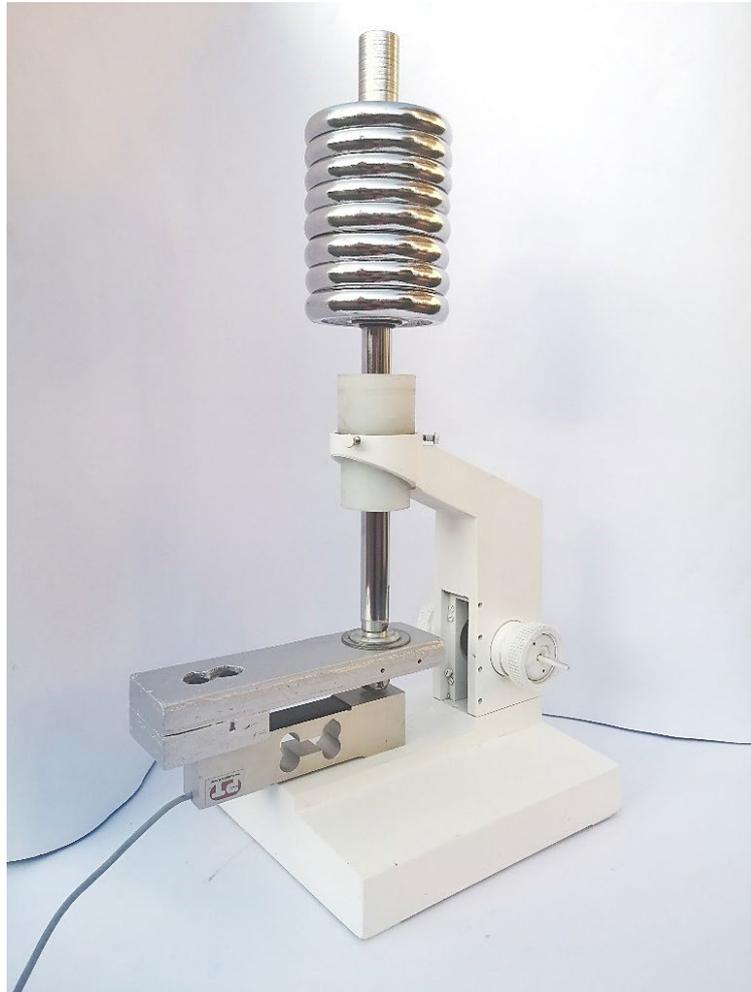

**Figura 5.12** Segundo prototipo de banco de pruebas de fuerza.
**Fuente:** Elaboración propia.

Cuando se realizó el montaje del prototipo final de banco de pruebas de fuerza, se tuvo como objetivo principal corregir el mayor defecto del anterior prototipo de Banco de Pruebas, el cual es la aparición de fuerzas axiales que afectan la exactitud de la precisión y repetibilidad al tomar las muestras del sensor.

Para lograr este objetivo, se realizó el diseño de un banco de pruebas de fuerza manual, con un cilindro metálico tipo vástago con un buje de teflón, para que el cilindro metálico tipo vástago realizara un recorrido vertical, como se puede observar en la Figura 5.12.

Una de las ventajas del prototipo final es su facilidad de manejo al momento de aplicar la carga sobre el sensor de fuerza neumático y la celda de carga, como se puede observar en la Figura 5.12. Siendo capaz además de aplicar rango de fuerzas debido a la posibilidad de utilizar varios pesos conocidos a la vez.

## 5.3. SISTEMA DE ADQUISICIÓN DE DATOS

Cuando se habla de un sistema de adquisición de datos, generalmente abreviado como *DAQ (Data acquisition)*, nos referimos al proceso de hacer mediciones de fenómenos físicos, registrarlo de alguna manera y analizarlos posteriormente.

En un sistema de adquisición de datos, las señales analógicas se convierten a señales digitales y luego son grabadas en una memoria. Un sistema de adquisición de datos cuenta con múltiples canales de circuitos de acondicionamiento de señal que trabajan como la interfaz entre los sensores y el sistema de conversión analógico/ digital. (Dewesoft 2020)

Un sistema de adquisición de datos generalmente puede realizar mediciones de los siguientes fenómenos físicos:

- Voltaje.
- Corriente.
- Choque y vibración.
- Distancia y desplazamiento.
- Galgas extensiométricas y presión.
- Peso.
- Temperatura.
- R.P.M. y ángulo.

El propósito principal de un DAQ es el de adquirir y almacenar los datos, pero también están destinados a proporcionar una visualización y análisis de los datos en tiempo real y para su posterior grabación. Además, la mayoría de los sistemas de adquisición de datos tienen incorporadas algunas capacidades analíticas y de generación de informes (Dewesoft 2020).

Para poder realizar la comparativa entre los resultados de la celda de carga y el sensor de fuerza neumático se utilizaron dos sistemas de adquisición de datos, cada sistema estaba enfocado a visualizar y grabar los datos de un sensor en específico, el motivo de utilizar dos sistemas separados es que cada sistema solo puede realizar la adquisición de datos de un sensor a la vez.

### 5.3.1. NI cDAQ-9174

El primer sistema de adquisición de datos utilizado fue el NI cDAQ 9174 (figura 5.13) con el módulo de Serie universal para entradas analógicas NI 9403, especificaciones en el ANEXO 2. Una de las características por el que fue utilizado, es debido a su capacidad de realizar lecturas en tiempo real y de guardar las lecturas de la celda de carga utilizada en el banco de pruebas.

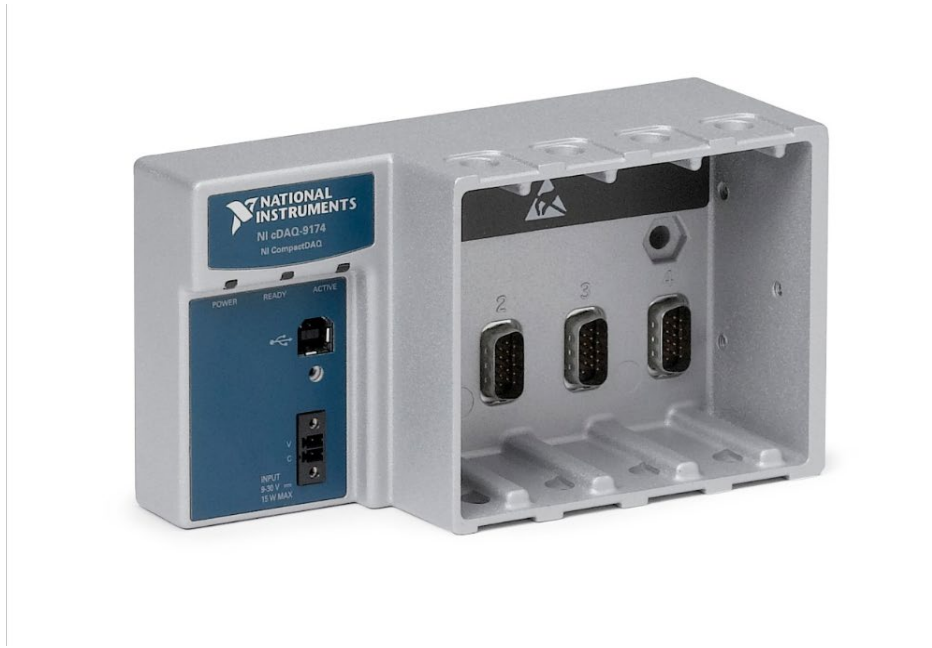

**Figura 5.13.** National Instruments cDAQ-9174.
**Fuente:** («cDAQ-9174 Chasis CompactDAQ» 2020)

El software DAQExpress es un software fabricado y distribuido por la compañía *National Instruments,* y está diseñado para realizar la lectura y almacenamiento en tiempo real, mediante conexión USB, del sistema de adquisición de datos cDAQ-9174.

En la Figura 5.14. se observa la interfaz del software DAQExpress, utilizado para la adquisición de datos de la Celda de Carga.

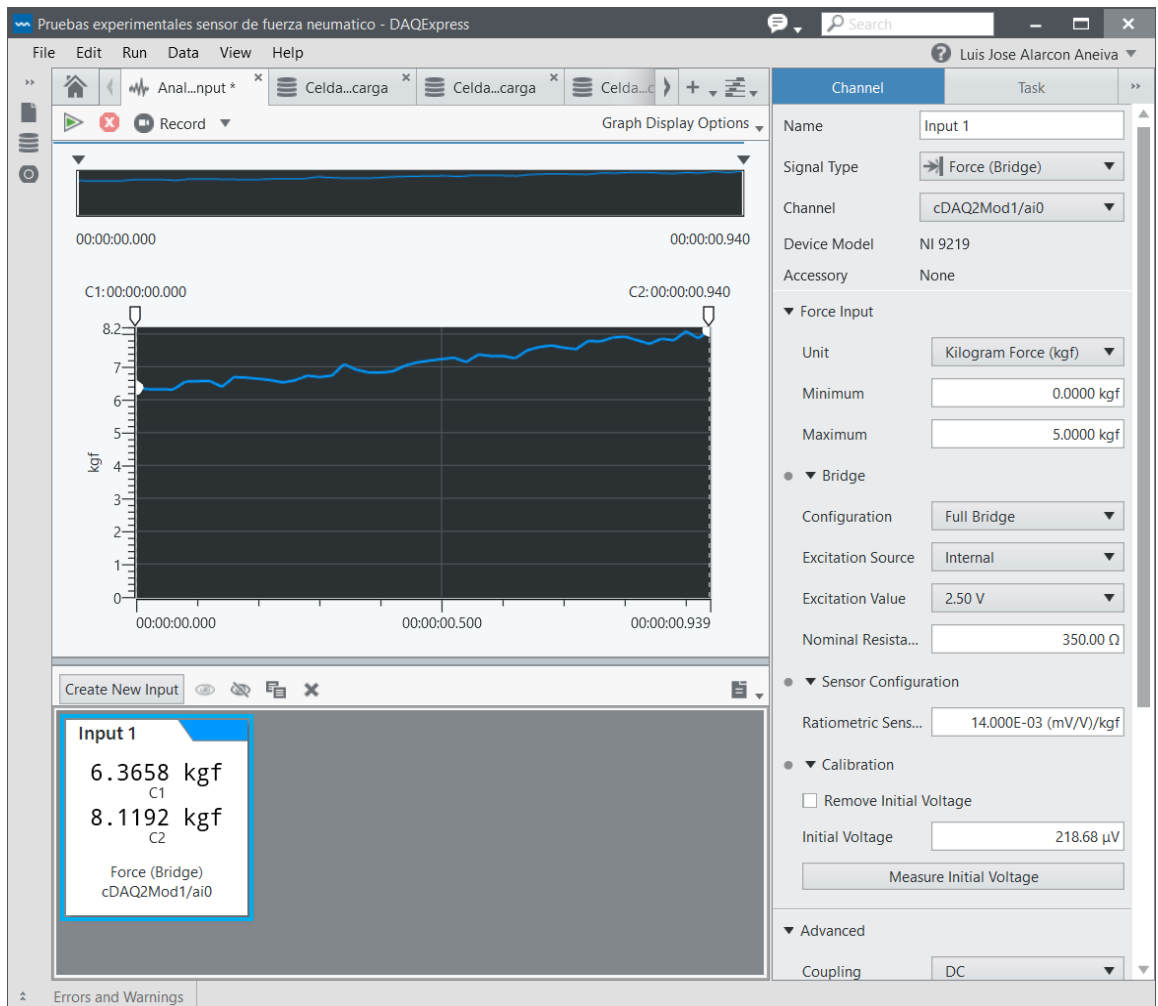

**Figura 5.14** Interfaz del software DAQExpress.
**Fuente:** Elaboración propia.

Para realizar la conexión eléctrica de la celda de carga con el sistema de adquisición de datos cDAQ-9174, se tiene que siguió el diagrama de la Figura 5.15, donde se realiza una conexión en modo *Full Bridge*, donde el puerto EX+ corresponde la terminal 3, EX – corresponde a la terminal 5, HI a la terminal 4 y, por último, LO corresponde a la terminal 6.

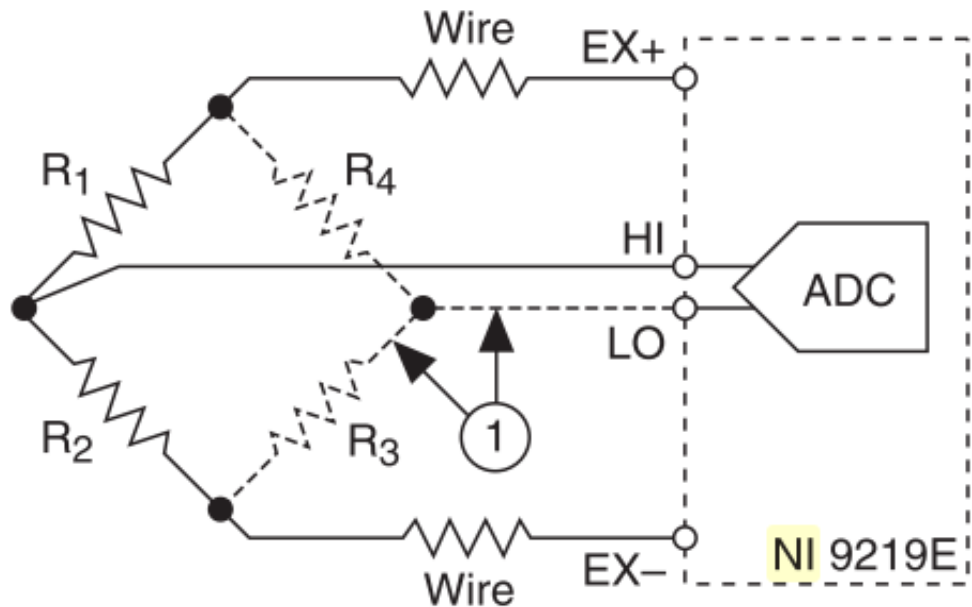

**Figura 5.15** Esquema electrónico de la conexión de la celda de carga con el cDAQ-9174.
**Fuente:** (NATIONAL INSTRUMENTS CORP. 2010)

### 5.3.2. *NI myRIO 1900*

El segundo sistema de adquisición de datos utilizado fue el NI myRIO 1900 (figura 5.14), es un dispositivo embebido de tiempo real que fue introducido por *National Instruments*. Se lo puede usar para desarrollar sistemas que requieren *FPGA* y/o un microprocesador a bordo. El lenguaje de programación que utiliza es el *LabVIEW*. El *myRIO* fue escogido como el segundo sistema de adquisición de datos, por su alta resolución de lectura de entradas analógicas de 0-5 V, la salida que utiliza el sensor de presión interna del sensor de fuerza neumático, por su capacidad de realizar lecturas en tiempo real y su interfaz gráfica que fue desarrollada en *LabVIEW*, en la cual se controla todo el sistema.

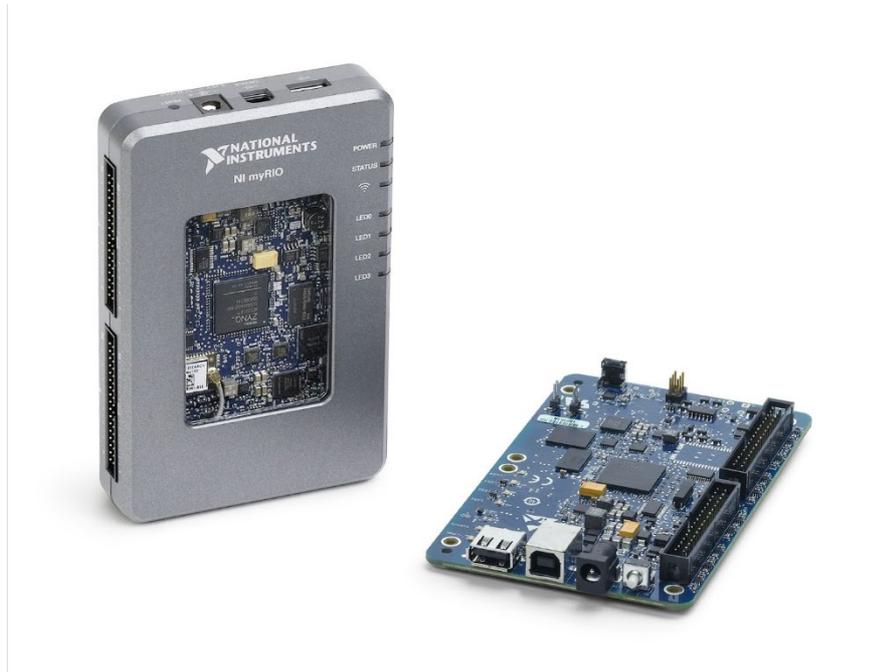

**Figura 5.14.** NI myRIO-1900.

**Fuente:** («myRIO Embedded Device - National Instruments» 2020)

Para realizar la lectura y adquisición de datos del sensor de presión con el NI myRIO-1900, se programo en el lenguaje *LabVIEW*, una interfaz que permite la visualización en tiempo real de los datos del sensor de presión, y la lectura y almacenamiento de dichos datos.

El código de programado en *LabVIEW* desarrollado y la interfaz del mismo se pueden observar en la Figura 5.15 y la Figura 5.16.

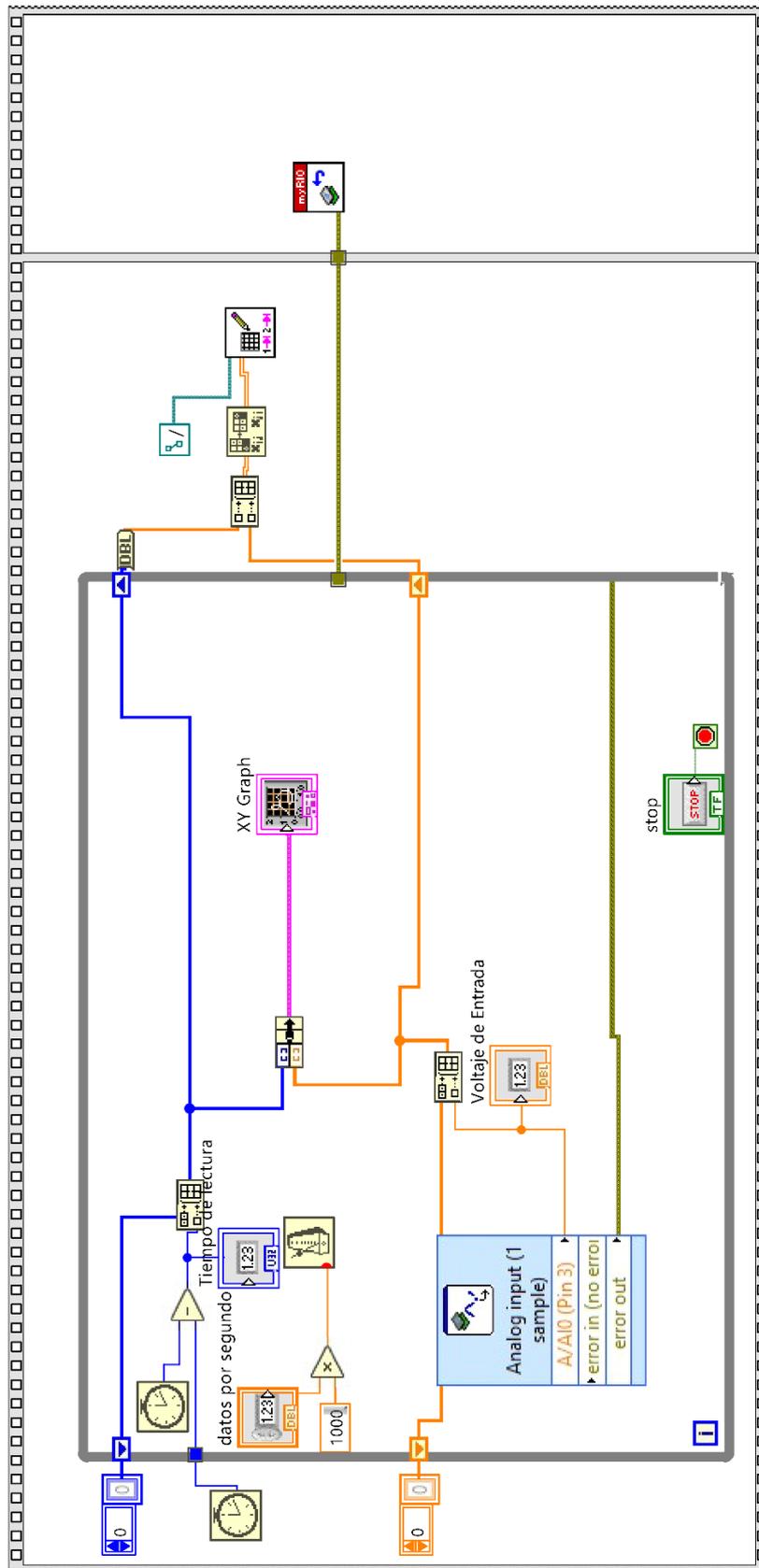

**Figura 5.15** Programación en bloques en *LabVIEW,* del sistema de adquisición de datos con el NI myRIO-1900.
**Fuente:** Elaboración propia.

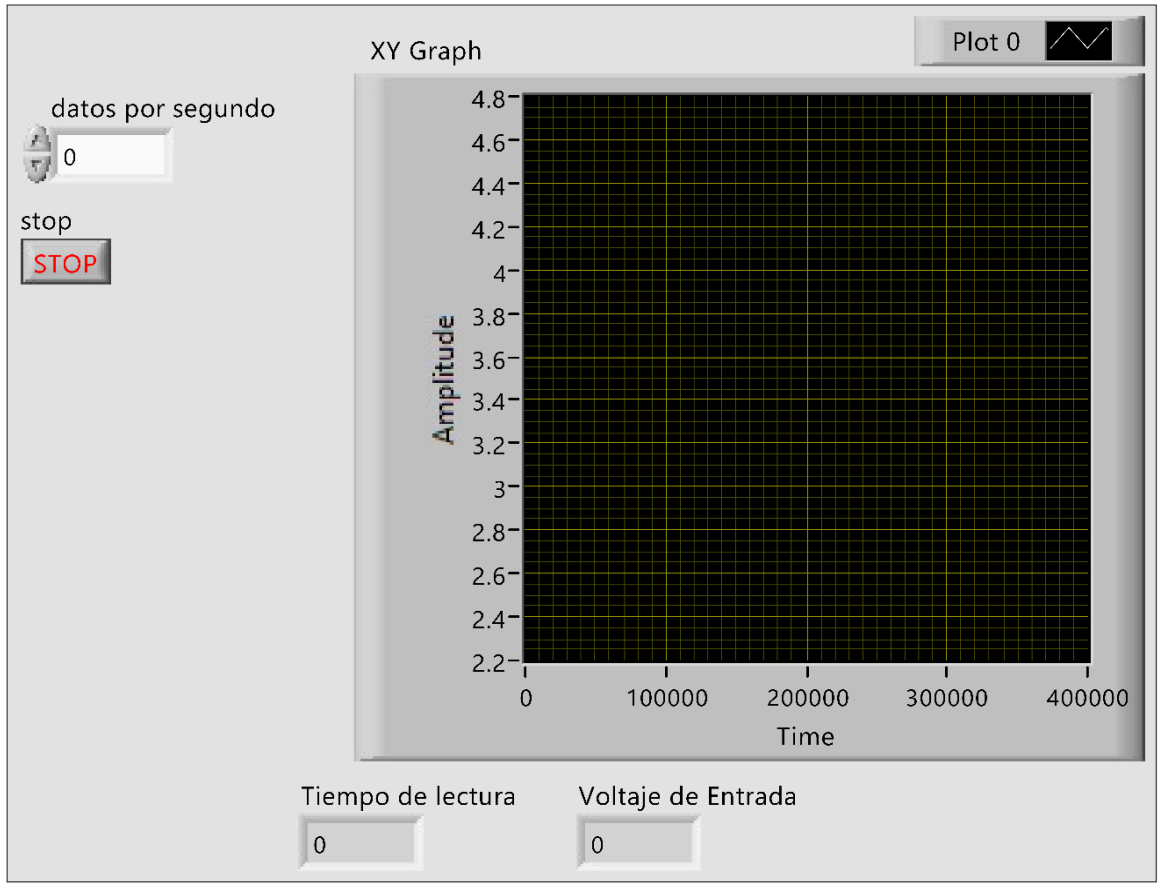

**Figura 5.16** Interfaz del programa desarrollado para la lectura y almacenamiento de datos del sensor de presión, usando el myRIO-1900.
**Fuente:** Elaboración propia.

Para realizar la conexión electrónica entre el sensor de presión MPX5500D y el myRIO-1900, se utilizó el diagrama eléctrico de la siguiente figura.

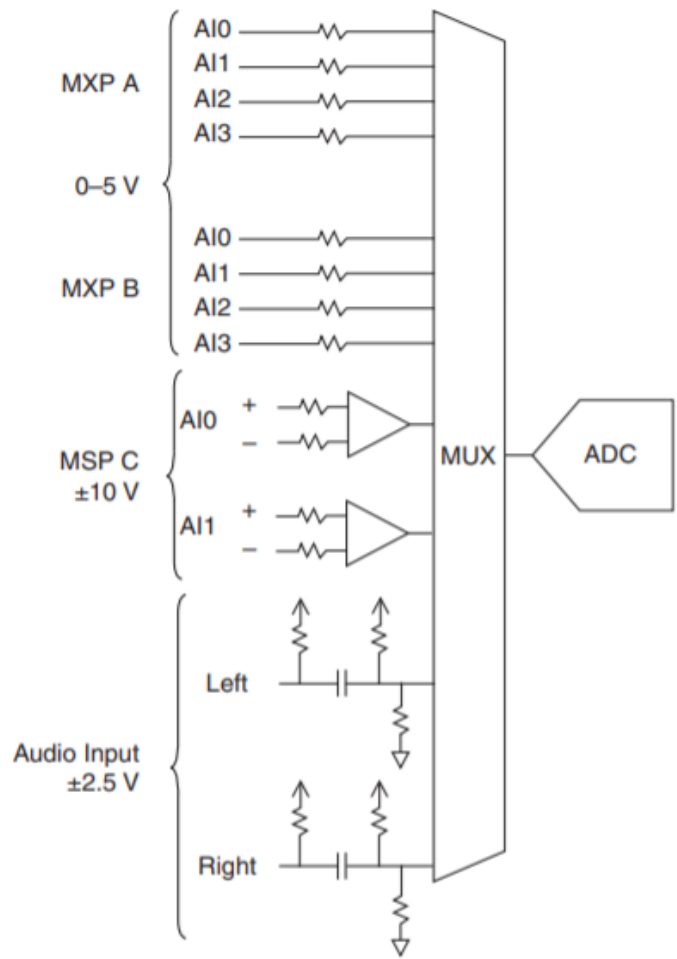

**Figura 5.17** Diagrama de puertos de entradas analógicas del NI myRIO-1900.
**Fuente:** (National Instruments 2013)

En la Figura 5.17, se puede observar los puertos de entrada analógica del NI myRIO-1900. Para realizar la conexión del sensor de presión con el NI myRIO-1900 se utilizó la entrada AI1, por el motivo de que en ese pin se pueden realizar lecturas de 0 a 5 [$V$]. La misma salida del sensor de presión MPX5500D utilizado.

## 5.4. CONCLUSIONES DEL CAPÍTULO V

En el presente capítulo se avanzó el desarrollo de los prototipos de sensor de fuerza neumático hasta llegar al prototipo final funcional. De igual forma se desarrollaron los prototipos de banco de prueba de fuerza, para que en el siguiente capítulo se lo pueda utilizar como la herramienta para realizar las pruebas experimentales del prototipo de sensor de fuerza neumático.

# CAPÍTULO VI

# PRUEBAS EXPERIMENTALES Y ANALISIS DE RESULTADOS

En el presente capítulo, se desarrolló el procedimiento para la calibración, clasificación del sensor de fuerza neumático y la validación del modelo matemático del sistema.

Cuando se realiza las pruebas experimentales del sensor, tiene dos partes. La primera parte fue para realizar la calibración y clasificación en conformidad con el "Procedimiento ME002, Calibración de instrumentos de medida de fuerza" del Centro Español de Metrología, el mismo que en consecuencia está basado en la norma UNE-EN ISO 376 "Norma de calibración de los instrumentos de medida de fuerza "(Centro Español de Metrología 2019).

La segunda parte se realizó para la validación de la simulación del modelo matemático del sistema con las pruebas experimentales realizadas, para así poder verificar la concordancia de los resultados.

Para concluir el capítulo, se realizó el análisis de los resultados obtenidos en la calibración y clasificación, como en la validación del sistema, y por consiguiente así determinar todas las propiedades del sensor de fuerza neumático.

## 6.1. NORMATIVA PARA LA CALIBRACION Y CLASIFICACION DE INSTRUMENTOS DE MEDIDA DE FUERZA

El procedimiento para la calibración de medida de fuerza ME-002 (Centro Español de Metrología 2019) basado en la norma UNE- EN ISO 376, describe el procedimiento para la clasificación de instrumentos de medición de fuerza, el cual aplica para los instrumentos de medida de fuerza que se basan en métodos indirectos, como, por ejemplo: el sensor de fuerza neumático.

### 6.1.1. Equipos y materiales

Para la calibración de los instrumentos basados en la presente normativa requiere utilizar el siguiente equipo y material auxiliar:

#### 6.1.1.1. Sistema de generación de fuerzas de referencia

La máquina de calibración de fuerza tiene que contar con valores de incertidumbre en función al instrumento a calibrar.

La normativa recomienda que la máquina de calibración de fuerza este diseñada y construida teniendo en cuenta los siguientes principios:

- La estructura de la maquina debe ser rígida, no se tienen que producir deformaciones que pudieran desvirtuar la fuerza de calibración a aplicar sobre el sensor a calibrar.
- La máquina de calibración de fuerza tiene que generar fuerzas con la estabilidad necesaria para la toma de medidas y así permitir la repetición de dicha fuerza de calibración tantas veces como fuera necesario.
- El diseño de la máquina de calibración de fuerza tiene que permitir la aplicación de una fuerza de calibración de forma axial, siendo mínimas las componentes parasitas, transversales, momentos flexores y momentos torsores.

#### 6.1.1.2. Dispositivo indicador de medida

Se tiene que incorporar un dispositivo indicador de medida compatible con la exactitud y clase esperada del instrumento de medida de fuerza a calibrar, tomando en cuenta la condición de que la incertidumbre del dispositivo indicador, no de influir de forma significativa en la incertidumbre del conjunto.

En la prueba experimental, como indicador de medida se utilizaron las interfaces graficas de los sistemas de adquisición de datos, el cDAQ 9174 y el myRIO 1900.

#### 6.1.1.3. Transductores patrones de referencia

La normativa recomienda utilizar transductores de referencia con el mismo alcance máximo que el alcance máximo del instrumento de medida de fuerza a calibrar. En caso de no ser posible, se debe tener calibrado el transductor de referencia para la fuerza nominal del instrumento de medida de fuerza a calibrar.

En la prueba experimental se utilizó como transductor de patrón de referencia, la celda de carga PT-2000, sus especificaciones técnicas se encuentran en el ANEXO 1.

### 6.1.1.4. *Dispositivo medidor de temperatura*

Para la determinación de las condiciones ambientales, la normativa recomienda utilizar un termómetro con resolución de indicación de al menos 0.01°C y una incertidumbre de 1 °C. Para la prueba experimental, se utilizó un módulo DHT11 para la determinación de la temperatura ambiente.

### 6.1.2. *Operaciones previas*

Antes de realizar el procedimiento de las pruebas del sensor de fuerza neumático, la normativa establece una serie de pasos a realizar previamente.

1) Se procedió a la identificación del alcance máximo del sensor de fuerza neumático, se aplica una carga creciente en el tiempo, hasta identificar la presión máxima admitida.

2) Se realizo el acondicionamiento metrológicamente de la sala donde se realizarán las pruebas. La sala tiene que tener una temperatura estable y comprendida entre 18°C y 28° C permitiéndose una variación máxima de ± 1 C, durante la realización de la misma, siendo la temperatura de la sala de 23° C al momento de la prueba.

3) Se procedió a limpiar adecuadamente las caras del sensor de fuerza neumático y de los componentes del banco de pruebas de fuerza, utilizando paños no abrasivos.

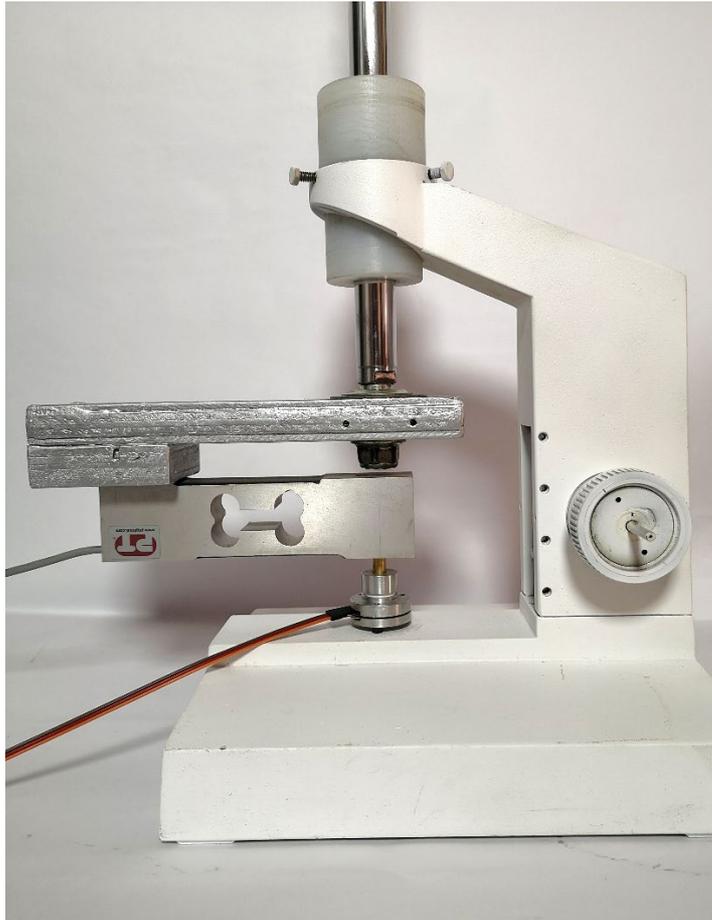

**Figura 6.1** Instalación del sensor de fuerza neumático en el banco de pruebas de fuerza.
**Fuente:** Elaboración propia**.**

4) Se instalo el sensor de fuerza neumático en el banco de pruebas de fuerza y antes de realizar las pruebas, como se observa en la figura 6.1, se llevó a cabo:

a. Un estudio visual de la idoneidad y buen estado de los diferentes componentes a emplear, así como del mismo sensor de fuerza neumático.

b. Se comprobó el cumplimiento de los siguientes requisitos:

i. <u>Resolución del dispositivo indicador</u>: Se verifico la resolución del dispositivo indicador y su correcta visualización.

ii. <u>Fuerza mínima</u>: Para poder verificar la correcta interacción entre el sensor de fuerza neumático y el banco de pruebas, se aplicó una fuerza mínima para comprobar su correcto funcionamiento.

c. Se verifica que el sensor de fuerza neumático sea apto para ser calibrado en el banco de pruebas, realizando los siguientes ensayos preliminares:

i.   <u>Ensayo de sobrecarga:</u> El sensor de fuerza neumático se somete cuatro veces seguidas a una sobrecarga que sobrepaso la fuerza máxima entre un 8 y 12%. Cada ensayo de sobrecarga duro entre 1 y 1.5 minutos.

ii.  <u>Ensayo bajo tensión eléctrica variable:</u> el sensor de fuerza neumática requiere de una fuente de alimentación eléctrica, según la normativa respectiva, se verifico que una variación de $\pm$ 10% de la tensión nominal de la red no tiene un efecto negativo significativo. Se utilizo una fuente de voltaje variable para el ensayo.

### 6.1.3. Proceso de Calibración

En base a la normativa establecida, los siguientes procesos para la calibración del sensor de fuerza neumática son:

Primero, se tomó el valor de indicación de cero antes de someter el sensor de fuerza neumática a fuerza alguna.

El valor de indicación de cero es: 2.361530725 V.

En el proceso de calibración del sensor de fuerza neumático, se realizan varias series de aplicación de fuerza con el banco de pruebas de fuerza. Cada serie es importante porque aporta los datos necesarios para realizar una correcta calibración del sensor. En la Figura 6.2. se muestra gráficamente todas las series de aplicación de fuerza en la calibración y clasificación.

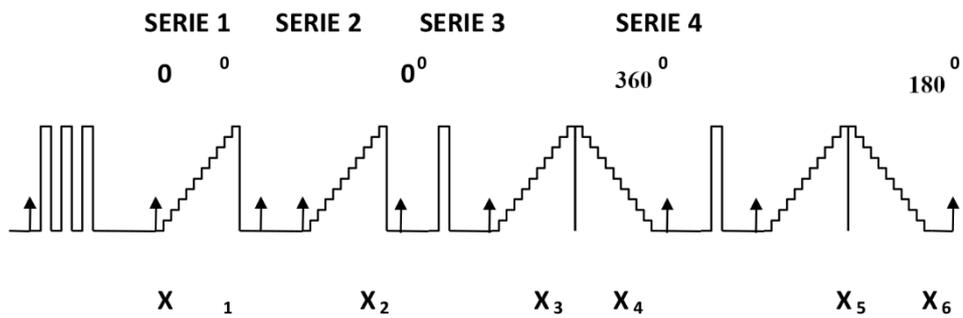

**Figura 6.2** Grafico de las series de fuerza realizadas para la calibración de sensores del sensor de fuerza neumático en base a la norma ME-002.

**Fuente:** (Centro Español de Metrología 2019)

Después, se realizaron 3 cargas con la fuerza del alcance máximo del sensor de fuerza neumático, que es de 4kg. La duración de cada carga de fuerza fue de 1 minuto y los intervalos de espera entre las cargas fueron de 3 minutos. No es necesario la recolección de datos de estas cargas para la calibración.

Posteriormente, se realizaron dos series de carga para fuerzas monótonamente crecientes con al menos 8 valores de fuerzas de calibración distribuidas uniformemente en el campo de medida calibrado, de 0 a 4 Kg. El intervalo entre las aplicaciones de fuerzas consecutivas tuvo que ser lo más uniforme posible. El sensor de fuerza neumático, durante estas dos series, fue situado en la misma posición de referencia angular (0 grados) en el banco de pruebas. Se puede observar la primera serie en la Figura 6.3 y la segunda serie en la Figura 6.4.

Entre las ejecuciones de cada serie de carga, existe un intervalo de tiempo de tres minutos. Se registraron los valores de fuerza de referencia, los valores ante carga nula al inicio y al final de cada serie, así como los valores para los diferentes escalones de fuerza de las series.

Se tomaron los valores de temperatura, al inicio y al final de cada una de las series, porque son relevantes para obtener la incertidumbre de temperatura.

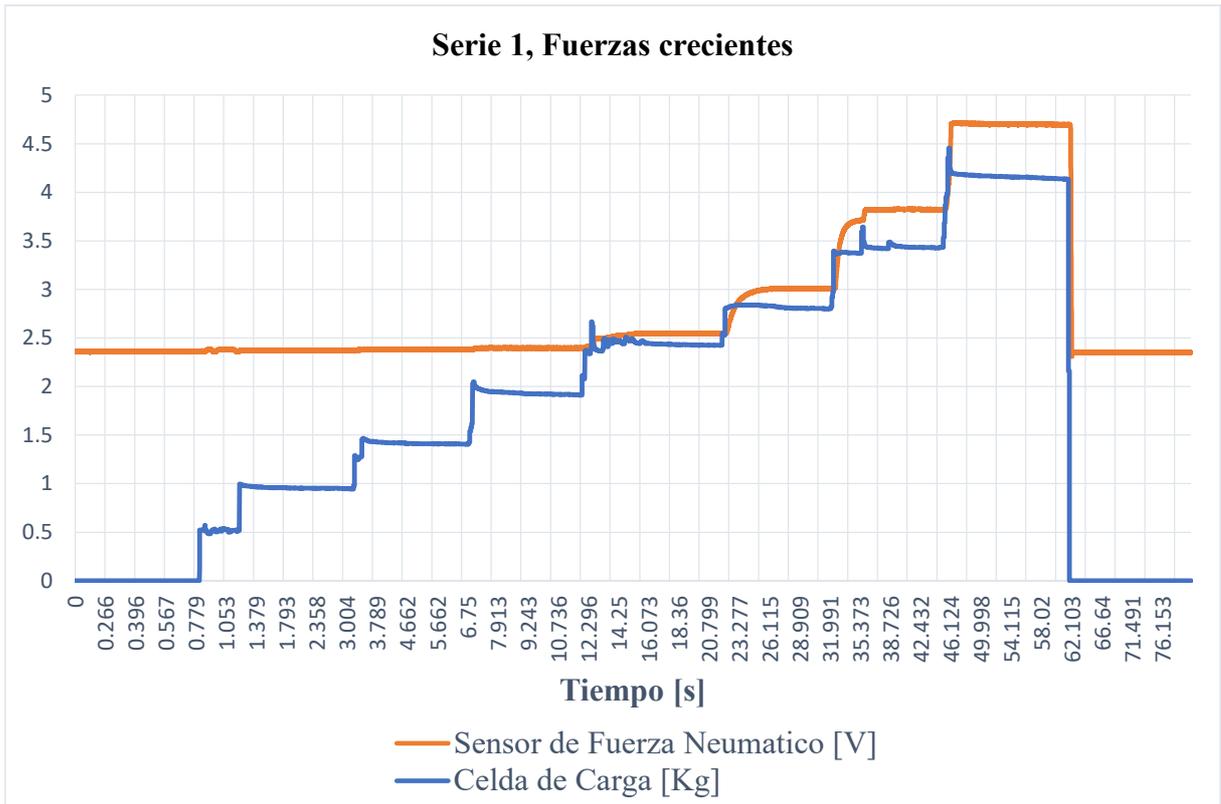

**Figura 6.3.** Resultado para la primera serie para fuerzas crecientes a 0 grados.

**Fuente:** Elaboración propia

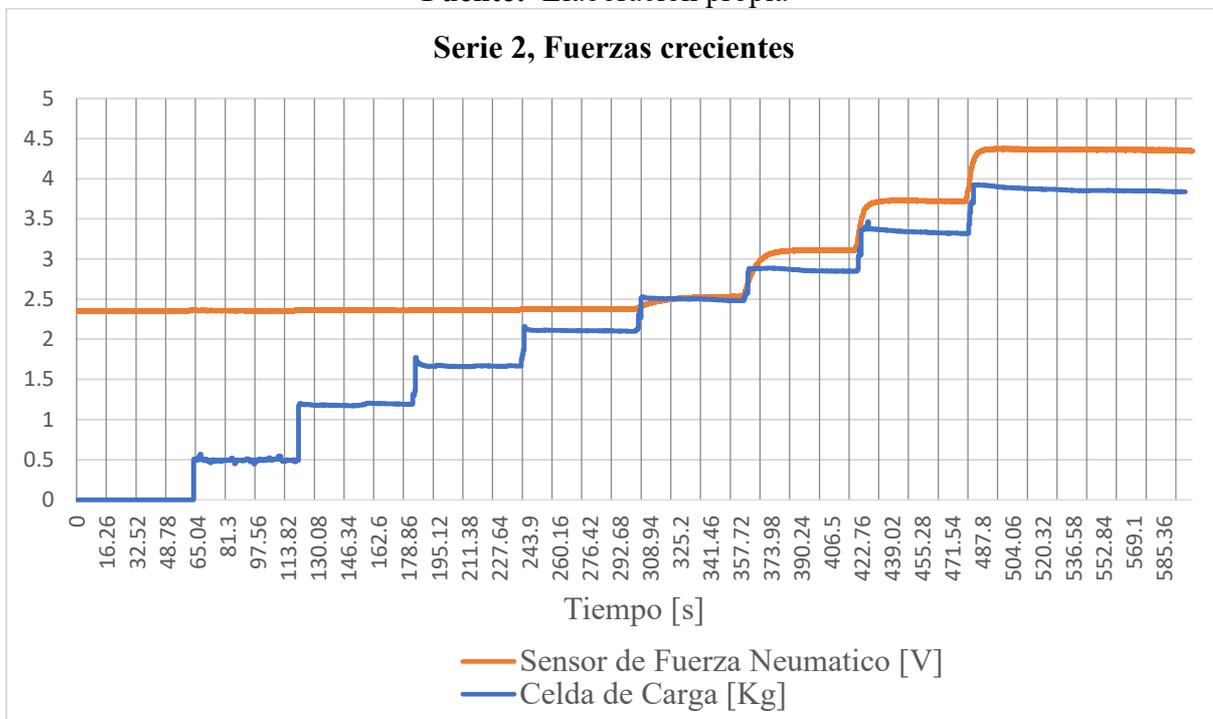

**Figura 6.4.** Resultados para la segunda serie para fuerzas crecientes a 0 grados.

**Fuente:** Elaboración propia.

A continuación, se realizaron otras dos series de cargas para fuerzas monótonamente creciente y decreciente variando la posición del sensor de fuerza neumático respecto a su eje en posiciones uniformemente sobre 180° y 360° de la posición inicial, se realizó la medición con el sensor invertido de posición.

En cada una de estas series se utilizaron los mismos valores de fuerza de referencia que en las dos series anteriores, en sentido monótonamente creciente y en sentido monótonamente decreciente hasta carga nula.

Antes del inicio de cada una de las series, se realizó una carga del peso máximo del alcance del sensor de fuerza neumático, con una duración de 1 minuto, y un intervalo de espera entre cargas de 3 minutos. Se tomaron los valores de temperatura al inicio y al final de cada una de las series. En la Figura 6.5 y la Figura 6.6, se observa las gráficas de los resultados de la prueba de las dos series.

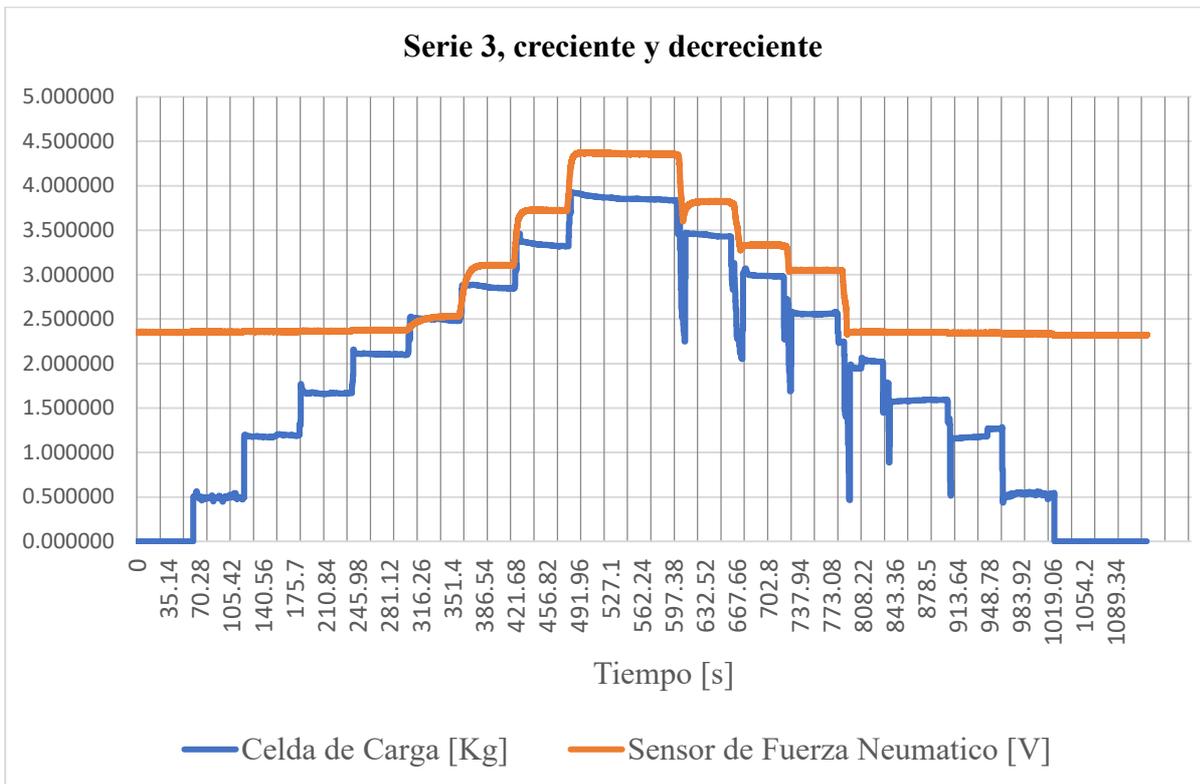

**Figura 6.5.** Resultado para la serie de fuerzas crecientes y decrecientes en un ángulo de 360°.
**Fuente:** Elaboración propia.

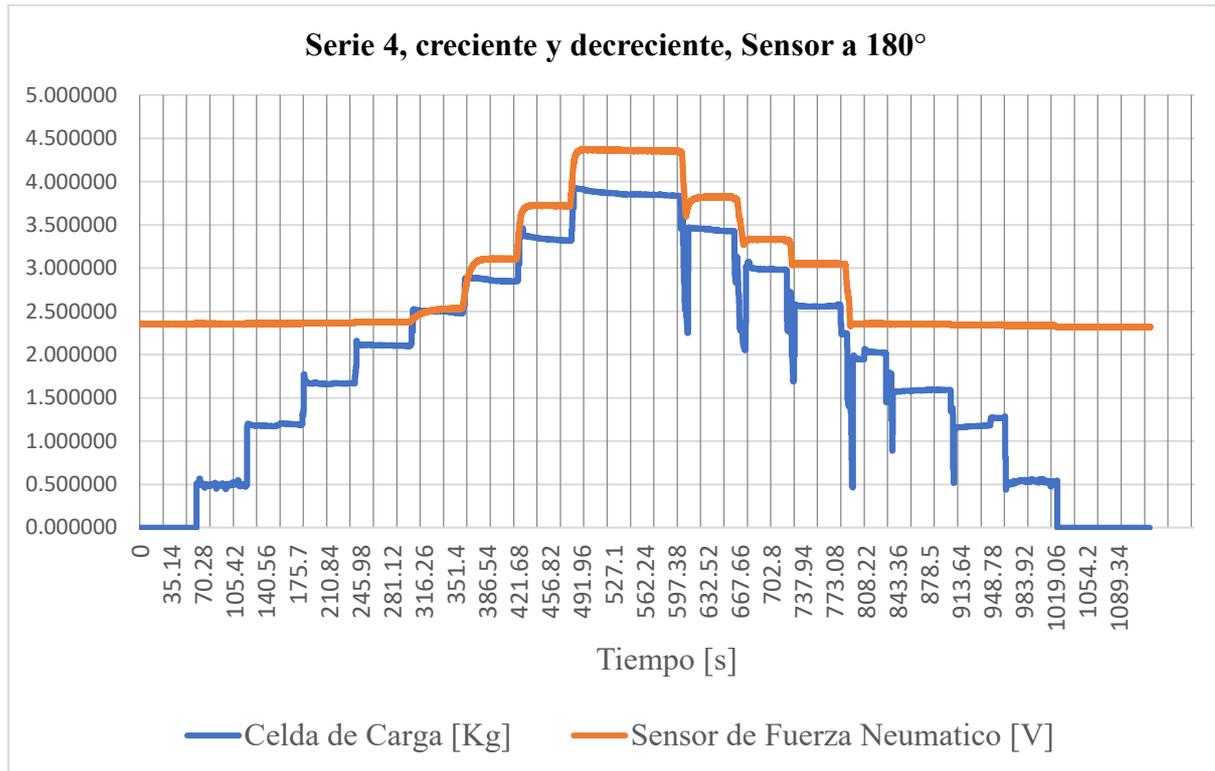

**Figura 6.6.** Resultado para la serie de fuerzas crecientes y decrecientes en un ángulo de 180°.
**Fuente:** Elaboración propia.

### 6.1.4. Toma y tratamiento de datos

La toma de datos, en base a la normativa establecida, que se realiza de forma manual o mediante un ordenador y un bus de comunicación que controle el sistema de adquisición de datos. En este caso la toma de datos fue mediante un ordenador y un DAQ como bus de comunicación, mediante los programas informáticos *DAQExpress* y *LabVIEW* se conservaron los ficheros de los datos de las pruebas para poder reconstruir la calibración realizada.

A partir de los datos registrados en los diferentes pasos establecidos en el apartado anterior, se calcularon los valores de los siguientes parámetros que caracterizaron la calibración y sirven de base para la clasificación del sensor de fuerza neumático según UNE-EN ISO 376. Los datos son reflejados en la Tabla 3.

**Tabla 6.1.** Medidas obtenidas en el transcurso de la calibración

| Indicación a fuerza nula | 2.341453525 |
|---|---|

| Indicaciones en V | | | | | |
|---|---|---|---|---|---|
| Serie de medidas – Posiciones angulares | | | | | |
| Fuerza de referen cia [Kg] | $X_1$ - 0° | $X_2$ - 0° | $X_3$ - 360° | $X_4$ - 360° | $X_5$ - 180° | $X_6$ - 180° |
| 0 | 2.3608374 26 | 2.3533884 03 | 2.34131 | 2.3223028 02 | 2.3527685 62 | 2.3181139 6 |
| 0.5 | 2.3720051 33 | 2.3578189 91 | 2.3424429 41 | 2.3277088 88 | 2.3566230 44 | 2.3356551 2 |
| 1 | 2.38038 | 2.3678480 64 | 2.3502566 17 | 2.3420160 81 | 2.3602329 61 | 2.3446770 62 |
| 1.5 | 2.3807885 55 | 2.3764122 67 | 2.3602815 19 | 2.3552470 51 | 2.3660881 91 | 2.3519907 17 |
| 2 | 2.3968393 67 | 2.3882875 88 | 2.3779530 42 | 2.3741968 64 | 2.3771084 09 | 2.3580074 08 |
| 2.5 | 2.5399582 18 | 2.6009539 77 | 2.4079278 43 | 2.3766366 55 | 2.5112106 02 | 3.0508268 97 |
| 3 | 2.9792101 04 | 3.2224803 6 | 2.8933519 42 | 3.1616030 74 | 3.0994227 58 | 3.3321250 42 |
| 3.5 | 3.8229031 02 | 3.7986288 74 | 3.6142274 71 | 3.5126558 37 | 3.7240425 56 | 3.8225828 1 |
| 4 | 4.7017696 21 | 4.4970694 11 | 4.4361818 7 | 4.4327372 07 | 4.3673214 89 | 4.3598355 7 |
| 0 | 2.3495379 62 | 2.3541150 39 | ----- | ----- | ----- | ----- |

Fuente: Elaboración propia.

Tomando los valores de las deformaciones de la Tabla 3, donde cada columna es una serie de las pruebas experimentales y cada línea es una carga conocida aplicada en la serie respectiva, se calcularon las medias con rotación ($\overline{X}_r$) y los diferentes errores en el procedimiento, descritos en mayor detalle en el capítulo 2.

**Tabla 6.2** Tabla de medida de errores

| Fuerza de referencia<br><br>F<br>en [Kg] | Media con rotación<br><br>$\bar{X}_r$<br>[V] | Error relativo de reproducibilidad con rotación<br><br>$\mathbf{b}$ (%) | Error relativo de repetibilidad sin rotación<br><br>$\mathbf{b}'$ (%) | Error relativo de reversibilidad<br><br>v (%) |
|---|---|---|---|---|
| 0.5 | 0.01116771 | 1.254217006 | 0.5998592 | 0.75937412 |
| 1 | 0.01954257 | 1.274457942 | 0.52785738 | 0.50485301 |
| 1.5 | 0.01995113 | 0.865621763 | 0.18398586 | 0.40455653 |
| 2 | 0.03600194 | 0.827652304 | 0.35743165 | 0.48074891 |
| 2.5 | 0.17912079 | 5.310175552 | 2.3729547 | 11.3939 |
| 3 | 0.61837268 | 6.890475856 | 7.84528856 | 8.38960809 |
| 3.5 | 1.46206568 | 5.60897036 | 0.63699079 | 2.72819186 |
| 4 | 2.34093219 | 7.429278904 | 4.4505662 | 0.12452842 |

Fuente: Elaboración propia.

**Tabla 6.3.** Tabla de error relativo a cero

| Serie | $X_1$ | $X_2$ | | $X_3 - X_4$ | $X_5 - X_6$ |
|---|---|---|---|---|---|
| $f_0$ (%) | 0.24032365 | 0.016158 | | 0.4284585 | 0.7948603 |

Fuente: Elaboración propia.

### 6.1.5. Incertidumbre

Posteriormente, se procedió al cálculo de la incertidumbre típica relativa combinada $w_c$ de calibración para cada fuerza ensayada, utilizando como magnitudes de entrada los errores anteriormente calculados, por consiguiente, utilizando los cálculos de incertidumbre descritos en el capítulo 2.

La incertidumbre típica relativa combinada de la calibración calculada es de:

$$w_c = 0.06384195$$

### 6.1.6. Interpretación de los resultados

**Principio de clasificación**

El campo de medida para que el instrumento de medida de fuerza sea clasificado se determina en base de cada fuerza de calibración sucesivamente, empezando por la fuerza máxima y disminuyendo hasta el valor mínimo de calibración. Este campo de clasificación se interrumpe en el último valor de fuerza para el cual las exigencias de calibración se satisfacen.

El rango de medida de clasificación de un instrumento de medida de fuerza debe cubrir al menos el rango del 50% al 100% del rango nominal.(Centro Español de Metrología 2019)

En base del procedimiento de calibración y clasificación de instrumentos de fuerza, el sensor de fuerza neumático se clasifica como un Caso B.

Caso B, son los instrumentos clasificados para fuerzas específicas y para cargas crecientes y decrecientes, los criterios que deben considerarse son:

- los errores de reproducibilidad y de repetibilidad;
- el error relativo de cero y
- el error relativo de reversibilidad.

De los resultados de las medidas realizadas, los valores de los errores e incertidumbres calculados en base del procedimiento. Caracterizan al instrumento de medida de fuerza y posibilita su clasificación según la norma UNE EN ISO 376, la misma que se describe en la siguiente tabla.

**Tabla 2.4** Tabla de clasificaciones de instrumentos de medida de fuerza.

| Clase | Error relativo del instrumento de medida de fuerza, % | | | | Resultado de medida |
|---|---|---|---|---|---|
| | de reproducibilidad repetibilidad | | de cero | de reversibilidad | Incertidumbre de calibración Min - Max |
| | b | b′ | $f_0$ | v | % |
| 0 | 0.05 | 0.03 | ±0.0025 | 0.07 | $W_{mcf}$ - 0.06 |
| 0.5 | 0.1 | 0.05 | ±0.05 | 0.15 | 0.06-0.12 |
| 1 | 0.2 | 0.1 | ±0.1 | 0.3 | 0.12-0.24 |
| 2 | 0.4 | 0.2 | ±0.2 | 0.5 | 0.24-0.45 |

Fuente:(Centro Español de Metrología 2019).

A continuación, se procedió a clasificar el sensor de fuerza neumático, comparando los datos de los errores e incertidumbres obtenidos con los valores máximos estipulados para cada clase en la Tabla 6.4.

Concluyendo, al coincidir en su mayoría con los rangos de la normativa, el sensor de fuerza neumático se clasifica como un instrumento de medida de fuerza Clase 2 según la norma UNE EN ISO 376.(Centro Español de Metrología 2019)

## 6.2. VALIDACIÓN DEL SISTEMA

A continuación, se comparan los valores de las pruebas experimentales realizadas al sensor de fuerza neumático, utilizando de entrada un escalón con una fuerza de 4 [Kgf], con los resultados de la simulación del modelo matemático del sistema, con una entrada de fuerza igual a la experimental. Comparación realizada en la Figura 6.7.

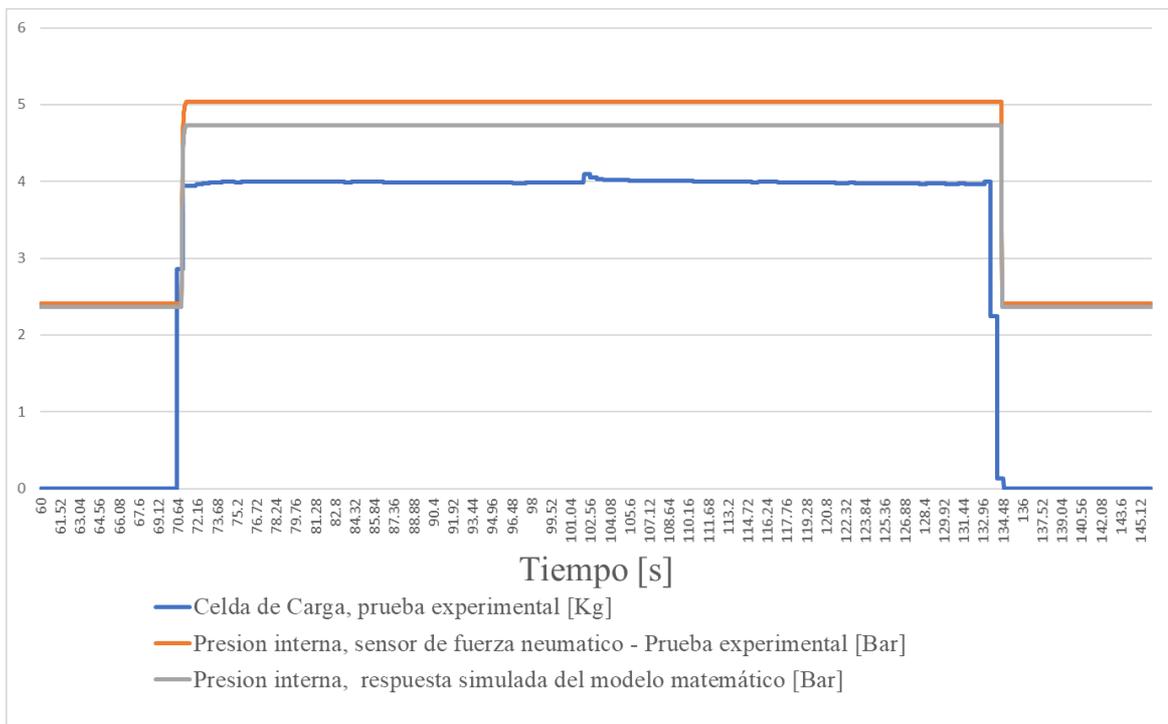

**Figura 6.7** Grafica de la respuesta experimental y simulada del sensor de fuerza neumático ante una señal escalonada.
**Fuente:** Elaboración propia.

Al analizar la Figura 6.7, en base a la respuesta ante una fuerza externa, en este caso de 4 *Kgf* en la simulación y en la prueba experimental, se puede observar los siguientes:

- Existe un retraso, en la curva del sensor de fuerza en la simulación y la prueba experimental, ante la respuesta del escalón de entrada de fuerza. Este retraso se debe a las características del sistema que, comparados con sus homólogos eléctricos, los sistemas neumáticos son sistemas más lentos.

- Se observa que, a la mitad de la prueba experimental, existe una entrada de ruido en los sistemas, al ser los sistemas neumáticos mas lentos, este es menos sensible a los ruidos a comparación de una celda de carga eléctrica.

- Se observa un desfase entre la respuesta de la prueba experimental y los resultados de la simulación. La diferencia entre ambas es de aproximadamente 0.4 Bar. Una vez analizada la observación, se determino de que uno de las posibles causas del desfase de la prueba experimental y simulada, se deba a la variación del comportamiento y/o de los parámetros utilizados, en la ecuación de movimiento, especificando en la fricción que se genera entre la cabeza del pistón y el cilindro interno. Los o-ring utilizados en el sensor ocasionan que el parámetro de fricción de coulomb sea difícil de determinar.

Tomando en cuenta las observaciones anteriores, se deduce que el comportamiento de la simulación y del sensor de fuerza neumático en la prueba experimental concuerdan, al existir una respuesta similar y coherente entre ambas.

## 6.3. CONCLUSIONES DEL CAPÍTULO VI

El propósito del presente capitulo fue el de realizar el procedimiento de calibración y clasificación del sensor de fuerza neumático en base de la norma UNE-EN ISO 376 [1]. De esta manera se lo clasifico al sensor de fuerza neumático como un sensor de clase 2.

Igualmente, se realizaron las pruebas experimentales para verificar el modelo matemático del sistema del sensor. Comparando los resultados de la simulación del modelo matemático, la prueba experimental y con la celda de carga, se observaron y analizaron los resultados, obteniendo una serie de ventajas y desventajas del sensor de fuerza neumático.

Las ventajas son las siguientes:

- El sensor de fuerza neumático es capaz trabajar como un elemento elástico y un sensor de fuerza simultáneamente. Lo cual permite el desarrollo de un transductor de fuerza de baja complejidad para robótica aplicada.

- El sensor de fuerza neumático al contar reducidas dimensiones, permite implementar sistemas con actuadores elásticos en series en espacios reducidos.

- El sensor de fuerza neumático soporta sobrecargas de fuerza externa, debido a que realiza una lectura de la fuerza en función a la variación de presión de la cámara interna del sensor. Por tal motivo, no existe un contacto directo entre el sensor de presión y la fuerza de entrada.

Las desventajas encontradas son:

- Observando y analizando los resultados obtenidos en las pruebas experimentales de series en el proceso de calibración del sensor de fuerza neumático. Se determino de que, debido a la propiedad de compresibilidad del aire encapsulado en el sensor, el rango de medición recomendado de lectura de fuerza del sensor es de 2.5 a 4kg. Siendo capaz de realizar lecturas de rangos menores, pero con menor sensibilidad.

- En las pruebas experimentales realizadas, se evidenció de que el sensor de fuerza neumático requiere de una recarga frecuente de aire comprimido en su compartimiento interno, debido a las microfugas existentes en sus o-rings, afectando a las lecturas de fuerza a largo plazo y no a corto plazo.

# CAPÍTULO VII

# MARCO CONCLUSIVO

## 7.1. CONCLUSIONES

El sensor de fuerza neumático fue desarrollado con el objetivo de facilitar la implementación de sistemas de control de fuerza de actuadores elásticos en serie. Al tener la ventaja, que el sensor de fuerza neumático a diferencia del resto de sensores de fuerza cuenta con dos funcionalidades en un sistema de control de fuerza. Primero trabajar como el sensor que realiza las mediciones de fuerza y a su vez incorpora la funcionalidad de trabajar como el elemento elástico en el sistema, De esta forma, abre la posibilidad de facilitar la implementación de control de fuerza para sistemas autómatas, manipuladores, tomando ventaja del reducido tamaño y de la capacidad del sensor.

El procedimiento desarrollado para el diseño del sensor comenzó con el modelado matemático del sensor de fuerza neumático, para luego simular el mismo, el software que se utiliza para su simulación fue Simulink, un complemento del Software matemático Matlab.

Una vez concluido el modelo matemático del sistema y su respectiva simulación con los parámetros iniciales, se procedió al desarrollo del prototipo del sensor de fuerza neumático. En total se desarrollaron 3 prototipos del sensor de fuerza neumático hasta llegar a la versión final. Cada prototipo presenta mejoras con respecto a sus predecesores, progresando hasta el prototipo funcional.

Para poder validar el sensor de fuerza neumático, se construyó un banco de pruebas de fuerza, para realizar las pruebas experimentales. Al realizar las pruebas, se siguió el procedimiento para la calibración de instrumentos de medida de fuerza del Centro español de metrología, el mismo procedimiento está basado en la norma UNE EN ISO 376. Una vez concluida la calibración y clasificación, se prosiguió a realizar la validación del modelo matemático del sensor propuesto en el documento.

Se realizo un análisis de los resultados obtenidos en las diferentes pruebas experimentales realizadas, obteniendo las ventajas y desventajas del sensor de fuerza neumático.

Concluyendo con la validación del modelo matemático que fue realizado, al obtener resultados que reflejan un comportamiento coherente entre la simulación y la prueba experimental.

En la calibración y clasificación del sensor de fuerza neumático, se logró la clase 2 en base a la norma de UNE EN ISO 376. Consecuentemente, se logró el objetivo de desarrollar el sensor de fuerza neumático, con la capacidad para su implementación en aplicaciones de control de fuerza, haciendo énfasis en manipuladores autómatas con SEA.

Es un aporte al campo de la metrología y a los sistemas de control de fuerza, el desarrollo de un nuevo tipo de sensor de fuerza, que se diferencia del resto por su sistema de cálculo indirecto de fuerza, que utiliza la variación de presión del aire dentro de su compartimiento interno. Así como la ventaja de su posible facilidad de implementación en aplicaciones de control de fuerza, al trabajar como el instrumento de medida de fuerza como del elemento elástico del actuador elástico en serie.

## 7.2. RECOMENDACIONES

Para el desarrollo de un sensor de fuerza neumático, es recomendable mejorar los compartimientos para los *o-ring* de la cabeza del pistón del sensor de fuerza neumático, por el motivo de que mejoraría en el funcionamiento del sistema el contar con menos fricción en el movimiento entre los componentes de la cabeza del pistón y el cilindro interno del sensor de fuerza neumático.

La segunda recomendación, es una mejora en el sistema de sellado del compartimiento interno del sensor de fuerza neumático, puesto que el que exista una fuga del compartimiento interno resultaría en una falla importante del sensor al perder su presión interna. Dejando la observación de que, en el prototipo final, cuando se realizaron las últimas pruebas experimentales, se observó una fuga de unas centésimas de bar por minuto, lo cual provocaba una variación de lectura ante entradas de fuerzas cercanas a 0.

### 7.3. TRABAJOS FUTUROS

Uno de los trabajos futuros con el sensor de fuerza neumático, es registrar la patente del sensor en el Servicio Nacional de Propiedad Intelectual de Bolivia en conjunto con la Universidad Católica Boliviana. Al ser un proyecto innovador en su aplicación y sistema de trabajo.

Un trabajo futuro por realizar con el sensor de fuerza neumático es la implementación del mismo dentro un sistema de control de fuerza, acoplándolo en un prototipo de mano biomecatrónica o una pinza de un brazo robótico y de esta manera realizar prácticas de manipulación de diferentes objetos, variando la fuerza aplicada sobre los mismos.

# BIBLIOGRAFÍA

# ANEXO 1
## Planos del sensor de fuerza neumatico

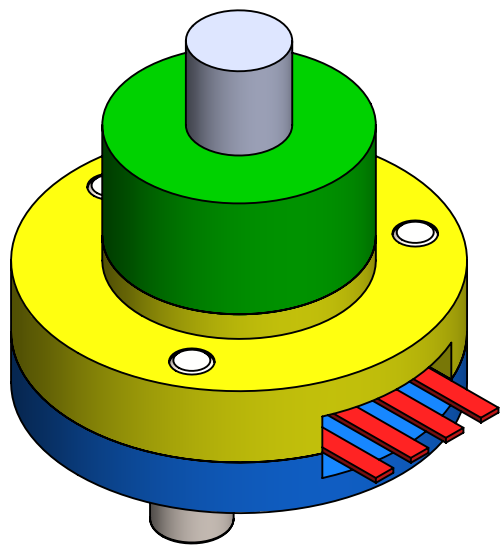

| Parte No. | Nombre | Cantidad |
|-----------|--------|----------|
| 1 | Tapa superior del sensor | 1 |
| 2 | Cilindro interno | 1 |
| 3 | Tapa superior del cilindro | 1 |
| 4 | Tapa inferior del sensor | 1 |
| 5 | Cabeza del piston | 1 |
| 6 | Valvula Schrader | 1 |
| 7 | Sensor MPX5500D | 1 |
| 8 | ISO 4762 M3 | 3 |

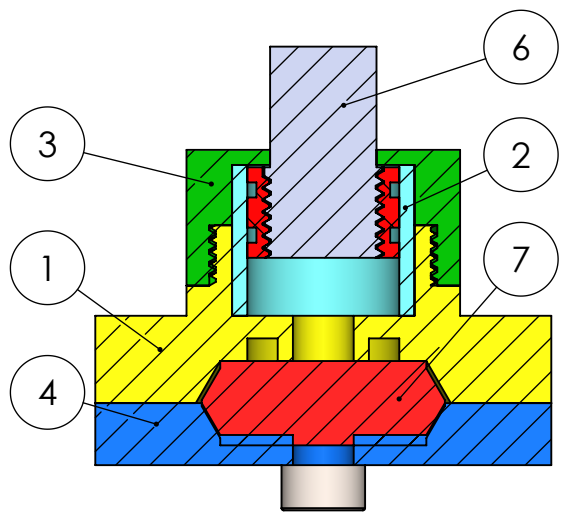

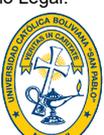

| Material: **Aluminio 5052** | | | Título **Ensamblaje del Sensor de Fuerza Neumático** | |
|---|---|---|---|---|

Acabado: **Pulido**

A menos que indique lo contrario:
Tol. Lineal.: **±0.01**, Tol. Angular.: **0°05'**
Acabado Superficial: **0.8μm**

Escala de dibujo: **2:1**

Peso Aprox.: **0.029**Kg

Dibujo producido en de acuerdo con:: **BS8888**

Metodo de proyeccion: **Tercer Angulo**

Tamaño de hoja:: **A4**

**Universidad Católica Boliviana San Pablo**

Dueño Legal:

UNIVERSIDAD **CATÓLICA** BOLIVIANA SANTA CRUZ

Dibujado por: **L.J.A.A.**

Fecha de dibujo **09/09/2020**

Verificado / Aprobado por: **D.R.A.**

Fecha de verificación / aprobación: **09/09/2020**

Número de parte -

Número de Dibujo: **1**

Hoja: **1 de 8**

Revisión:

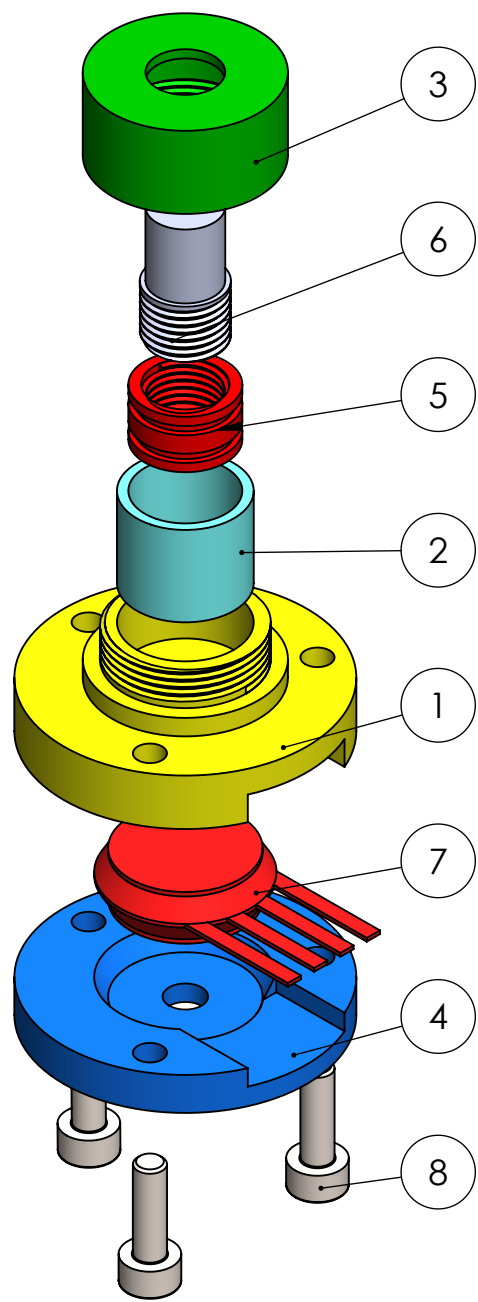

| Parte No. | Nombre | Cantidad |
|-----------|--------|----------|
| 1 | Tapa superior del sensor | 1 |
| 2 | Cilindro interno | 1 |
| 3 | Tapa superior del cilindro | 1 |
| 4 | Tapa inferior del sensor | 1 |
| 5 | Cabeza del piston | 1 |
| 6 | Valvula Schrader | 1 |
| 7 | Sensor MPX5500D | 1 |
| 8 | ISO 4762 M3 | 3 |

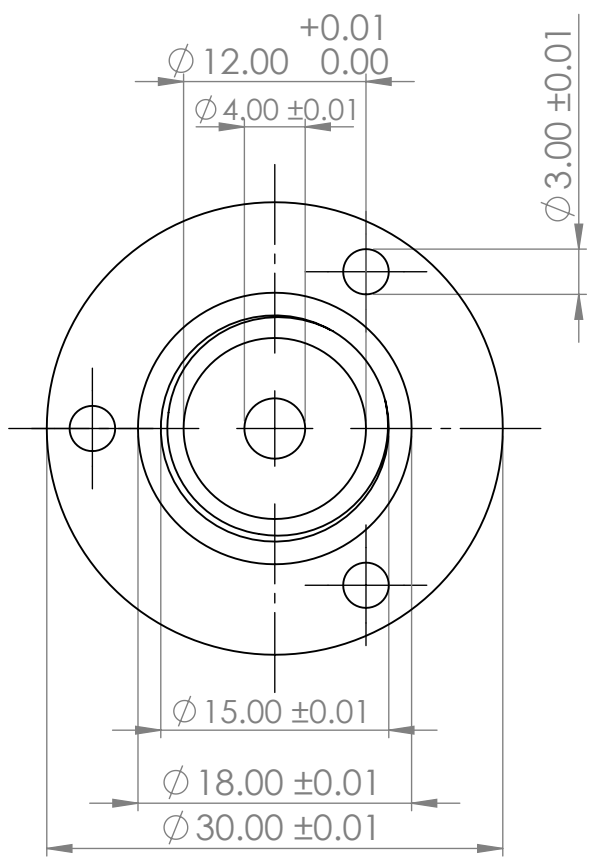

Ø 12.00 +0.01 0.00
Ø 4.00 ±0.01
Ø 3.00 ±0.01

Ø 15.00 ±0.01
Ø 18.00 ±0.01
Ø 30.00 ±0.01

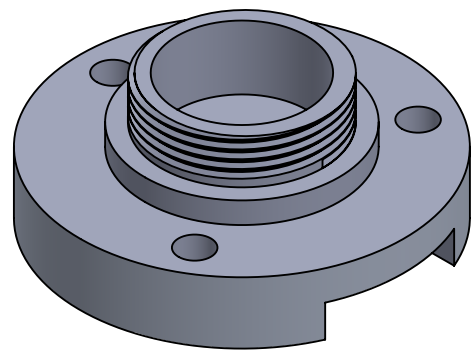

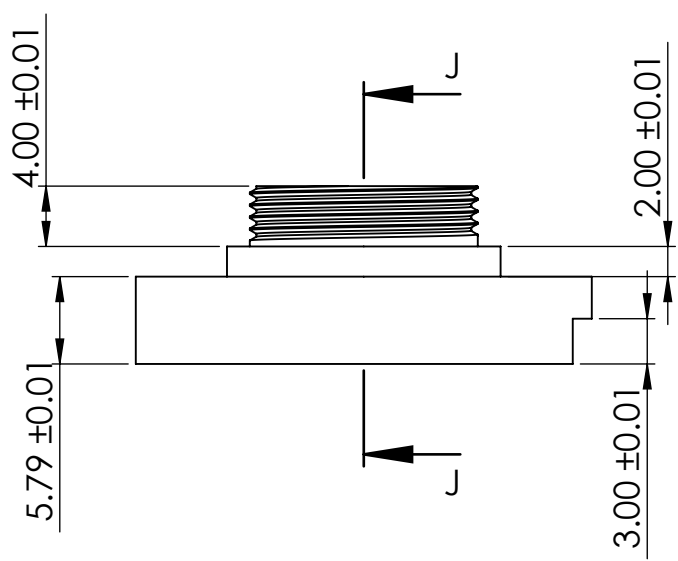

4.00 ±0.01
5.79 ±0.01
2.00 ±0.01
1.50 ±0.01
3.00 ±0.01

J

J

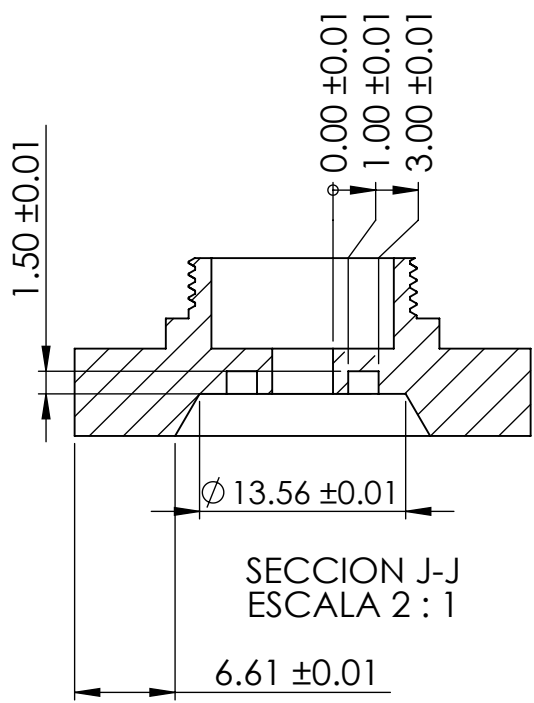

0.00 ±0.01
1.00 ±0.01
3.00 ±0.01

Ø 13.56 ±0.01

SECCION J-J
ESCALA 2 : 1

6.61 ±0.01

| Material: **Aluminio 5052** | Este dibujo y cualquier información o material descriptivo expuesto en él son propiedad y con derechos de la Universidad Catolica Boliviana San Pablo y NO DEBE SER COPIADO, PRESTADO total o parcialmente o utilizado para cualquier propósito sin el permiso por escrito. | Título **Tapa superior del sensor** | |
|---|---|---|---|
| Acabado: **Pulido** | | | |
| A menos que indique lo contrario: Tol. Lineal.: ±0.01, Tol. Angular.: 0°05' Acabado Superficial: 0.8µm | | Dibujado por: **L.J.A.A.** | Fecha de dibujo **09/09/2020** |
| Escala de dibujo: **2:1** | **Universidad Catolica Boliviana San Pablo** | Verificado / Aprobado por: **D.R.A.** | Fecha de verificación / aprobación: **09/09/2020** |
| Peso Aprox.: **0.002** Kg | Dibujo producido en de acuerdo con:: **BS8888** | Dueño Legal: | Número de parte **1** |
| Metodo de proyeccion: **Tercer Angulo** | Tamaño de hoja:: **A4** | UNIVERSIDAD CATÓLICA BOLIVIANA SANTA CRUZ | Número de Dibujo: **3** · Hoja: **3 de 8** · Revisión: |

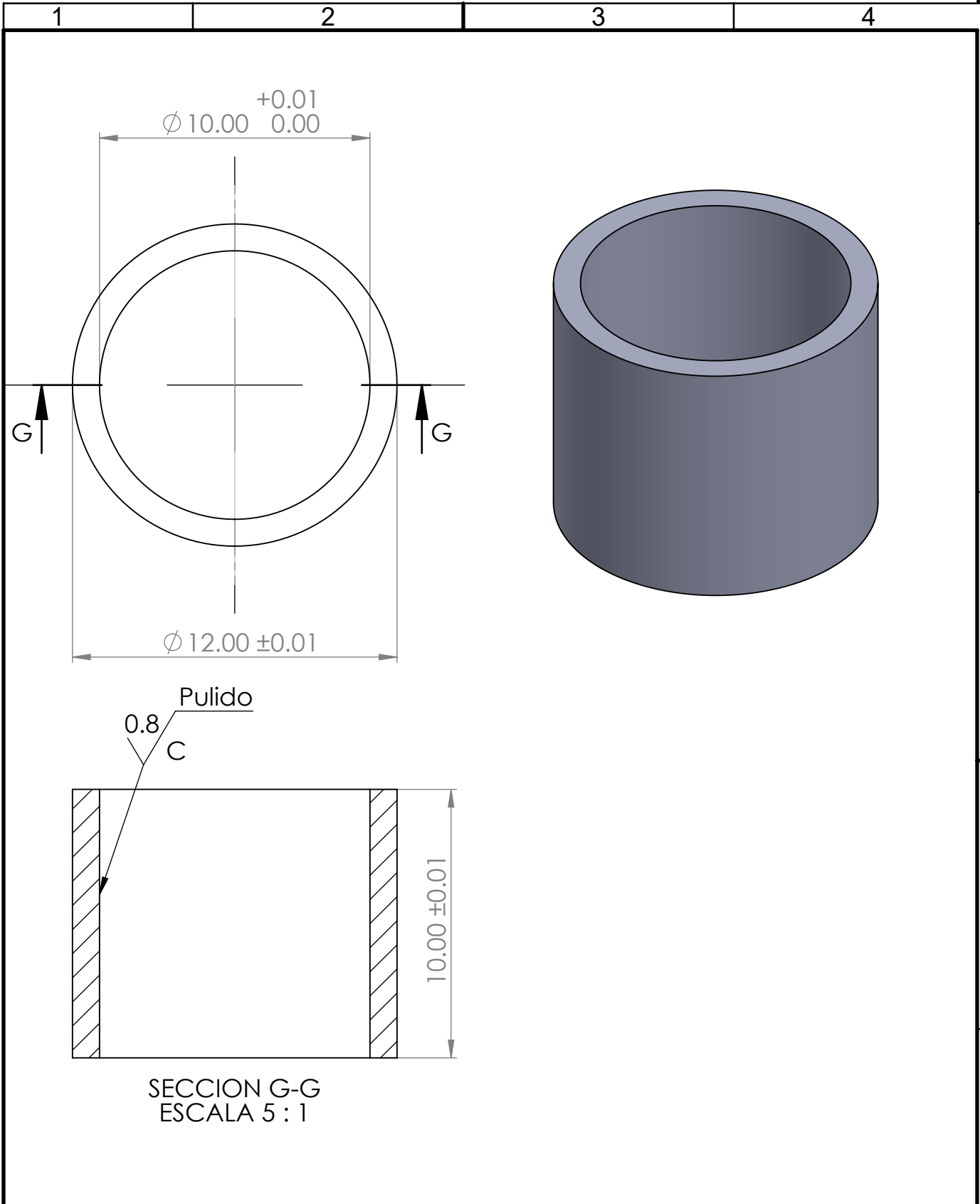

$\varnothing 10.00$ +0.01 / 0.00

$\varnothing 12.00 \pm 0.01$

G — G

Pulido

0.8

C

10.00 ±0.01

SECCION G-G
ESCALA 5 : 1

| | |
|---|---|
| Material: | **Aluminio 5052** |
| Acabado: | **Pulido** |
| A menos que indique lo contrario: Tol. Lineal.: **±0.01**, Tol. Angular.: **0˚05'** Acabado Superficial: **0.8μm** | |
| Escala de dibujo: | **5:1** |

Universidad Católica Boliviana San Pablo

| Peso Aprox.: **0.001** Kg | Dibujo producido en de acuerdo con:: **BS8888** |
|---|---|

Dueño Legal:

UNIVERSIDAD CATÓLICA BOLIVIANA SANTA CRUZ

| Metodo de proyeccion: **Tercer Angulo** | Tamaño de hoja:: **A4** |
|---|---|

Título
# Cilindro Interno

| Dibujado por: **L.J.A.A.** | Fecha de dibujo **09/09/2020** |
|---|---|
| Verificado / Aprobado por: **D.R.A.** | Fecha de verificación / aprobación: **09/09/2020** |

Número de parte
**2**

| Número de Dibujo: **4** | Hoja: **4 de 8** | Revisión: |
|---|---|---|

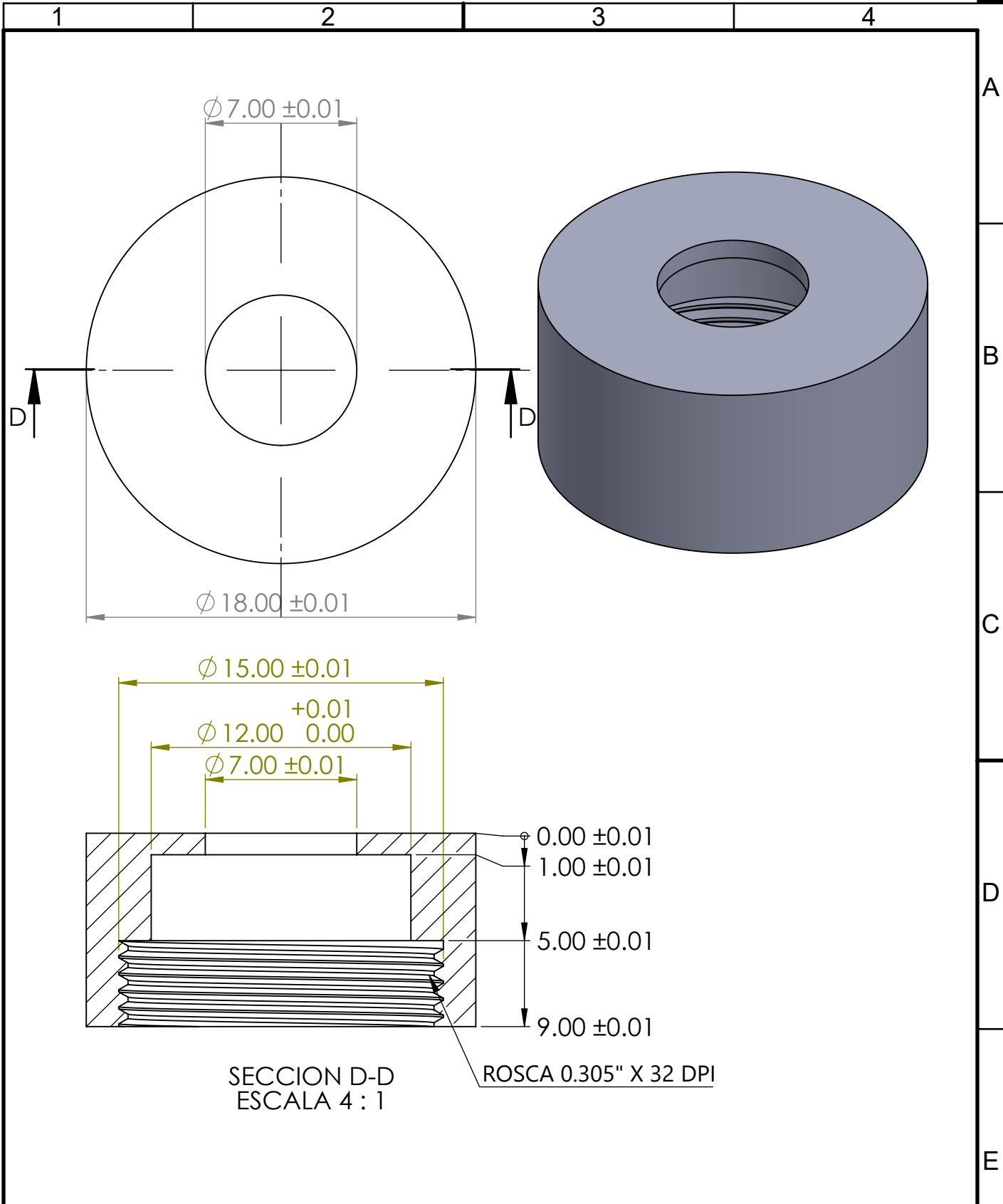

$\varnothing 7.00 \pm 0.01$

$\varnothing 18.00 \pm 0.01$

$\varnothing 15.00 \pm 0.01$

$\varnothing 12.00 \quad {}^{+0.01}_{\phantom{+}0.00}$

$\varnothing 7.00 \pm 0.01$

$0.00 \pm 0.01$

$1.00 \pm 0.01$

$5.00 \pm 0.01$

$9.00 \pm 0.01$

ROSCA 0.305" X 32 DPI

SECCION D-D
ESCALA 4 : 1

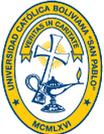

| Material: **Aluminio 5052** | Este dibujo y cualquier información o material descriptivo expuesto en él son propiedad y con derechos de la Universidad Catolica Boliviana San Pablo y NO DEBE SER COPIADO, PRESTADO total o parcialmente o utilizado para cualquier propósito sin el permiso por escrito. | Título **Tapa superior del cilindro** | |
|---|---|---|---|
| Acabado: **Pulido** | | Dibujado por: **L.J.A.A.** | Fecha de dibujo **09/09/2020** |
| A menos que indique lo contrario: Tol. Lineal.: **±0.01**, Tol. Angular.: **0˚05'** Acabado Superficial: **0.8µm** | **Universidad Catolica Boliviana San Pablo** | Verificado / Aprobado por: **D.R.A.** | Fecha de verificación / aprobación: **09/09/2020** |
| Escala de dibujo: **4:1** | | | |
| Peso Aprox.: **0.010** Kg | Dibujo producido en de acuerdo con:: **BS8888** | Dueño Legal: UNIVERSIDAD CATÓLICA BOLIVIANA SANTA CRUZ | Número de parte **3** |
| Metodo de proyeccion: **Tercer Angulo** | Tamaño de hoja:: **A4** | | Número de Dibujo: **5** | Hoja: **5 de 8** | Revisión: |

| 1 | 2 | 3 | 4 | |
|---|---|---|---|---|

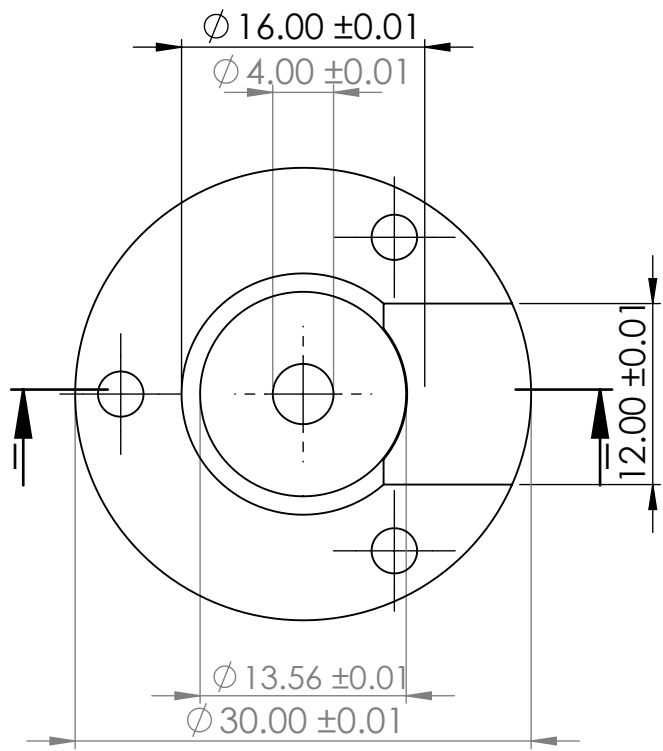

Ø 16.00 ±0.01
Ø 4.00 ±0.01
12.00 ±0.01
Ø 13.56 ±0.01
Ø 30.00 ±0.01

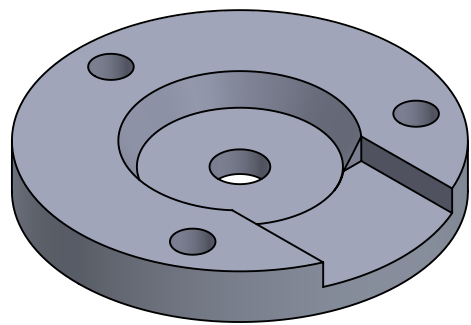

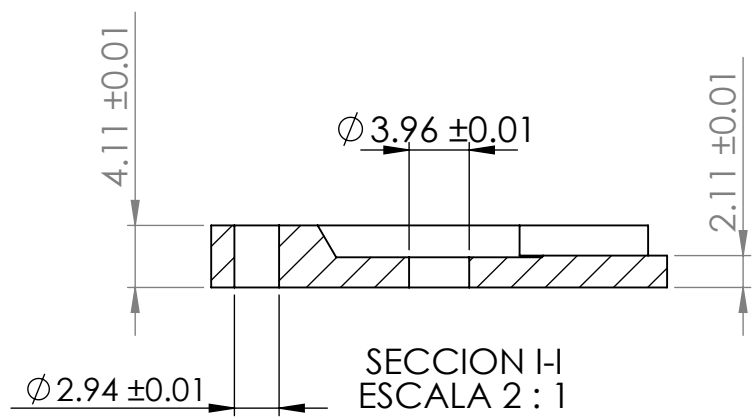

4.11 ±0.01
Ø 3.96 ±0.01
2.11 ±0.01
Ø 2.94 ±0.01

SECCION I-I
ESCALA 2 : 1

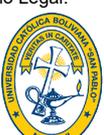

| Material: **Aluminio 5052** | Este dibujo y cualquier información o material descriptivo expuesto en él son propiedad y con derechos de la Universidad Catolica Boliviana San Pablo y NO DEBE SER COPIADO, PRESTADO total o parcialmente o utilizado para cualquier propósito sin el permiso por escrito. | Título **Tapa inferior del sensor** | |
|---|---|---|---|
| Acabado: **Pulido** | | | |
| A menos que indique lo contrario: Tol. Lineal.: **±0.01**, Tol. Angular.: **0°05'** Acabado Superficial: **0.8µm** | | Dibujado por: **L.J.A.A.** | Fecha de dibujo **09/09/2020** |
| Escala de dibujo: **2:1** | **Universidad Catolica Boliviana San Pablo** | Verificado / Aprobado por: **D.R.A.** | Fecha de verificación / aprobación: **09/09/2020** |
| Peso Aprox.: **0.006** Kg | Dibujo producido en de acuerdo con:: **BS8888** | Dueño Legal: | Número de parte **4** |
| Metodo de proyeccion: **Tercer Angulo** | Tamaño de hoja:: **A4** | UNIVERSIDAD **CATÓLICA** BOLIVIANA SANTA CRUZ | Número de Dibujo: **6** / Hoja: **6 de 8** / Revisión: |

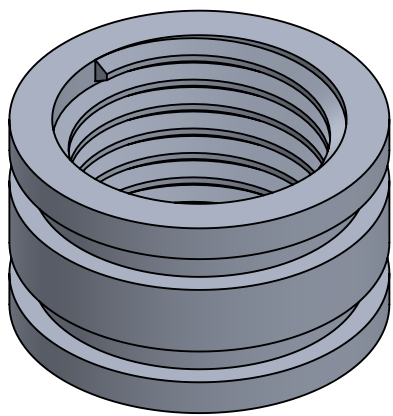

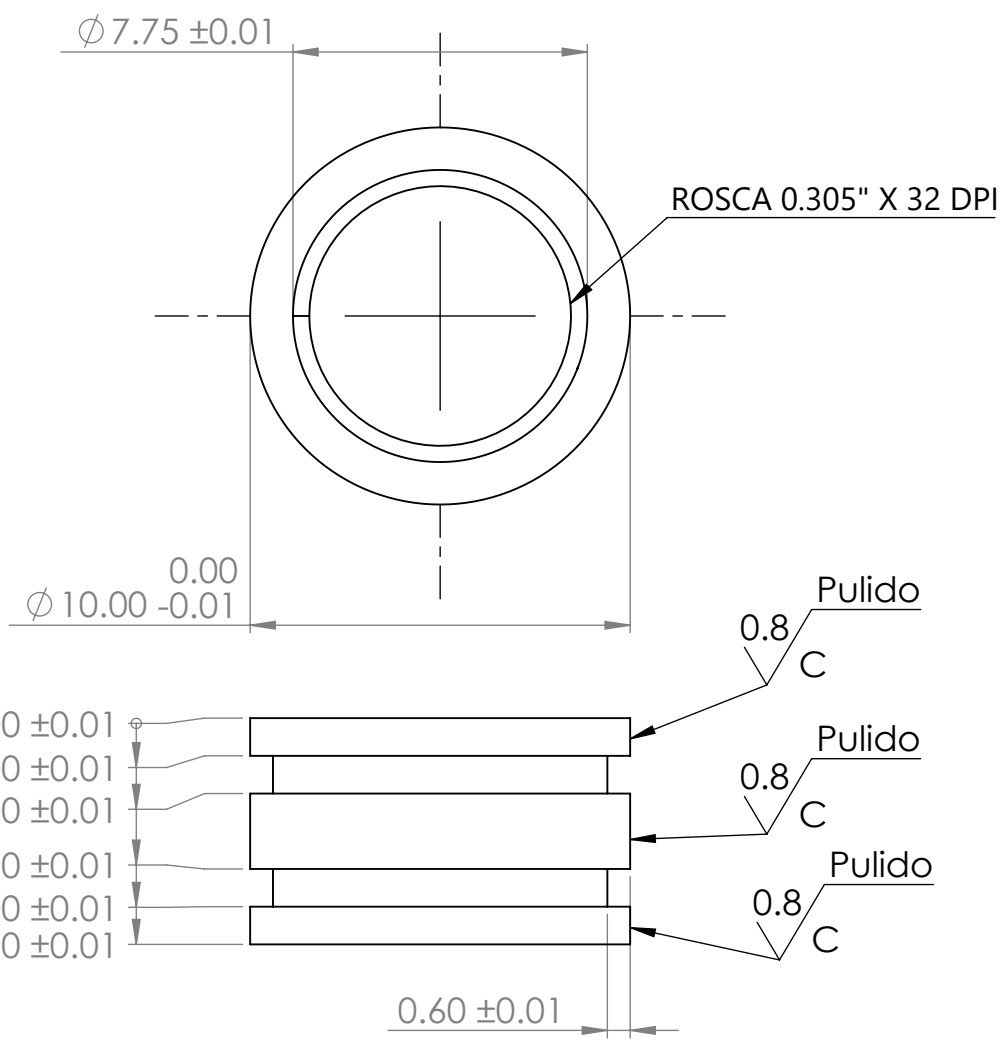

$\emptyset$ 7.75 ±0.01

ROSCA 0.305" X 32 DPI

0.00
$\emptyset$ 10.00 -0.01

Pulido

0.8 C

Pulido

0.8 C

Pulido

0.8 C

0.00 ±0.01
1.00 ±0.01
2.00 ±0.01
4.00 ±0.01
5.00 ±0.01
6.00 ±0.01

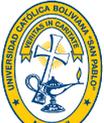

0.60 ±0.01

| | |
|---|---|
| Material: **Aluminio 5052** | Este dibujo y cualquier información o material descriptivo expuesto en él son propiedad y con derechos de la Universidad Católica Boliviana San Pablo y NO DEBE SER COPIADO, PRESTADO total o parcialmente o utilizado para cualquier propósito sin el permiso por escrito. |

Acabado:
**Pulido**

A menos que indique lo contrario:
Tol. Lineal.: **±0.01**, Tol. Angular.: **0°05'**
Acabado Superficial: **0.8μm**

Escala de dibujo:
**5:1**

Peso Aprox.:
**0.001** Kg

Dibujo producido en de acuerdo
con.: **BS8888**

Metodo de proyeccion:
**Tercer Angulo**

Tamaño de hoja::
**A4**

**Universidad Católica Boliviana San Pablo**

Dueño Legal:

UNIVERSIDAD
**CATÓLICA**
BOLIVIANA
SANTA CRUZ

Título
# Cabeza del pistón

Dibujado por:
**L.J.A.A**

Fecha de dibujo
**09/09/2020**

Verificado / Aprobado por:
**D.R.A**

Fecha de verificación / aprobación:
**09/09/2020**

Número de parte
**5**

Número de Dibujo:
**7**

Hoja:
**7 de 8**

Revisión:

| 1 | 2 | 3 | 4 |
|---|---|---|---|

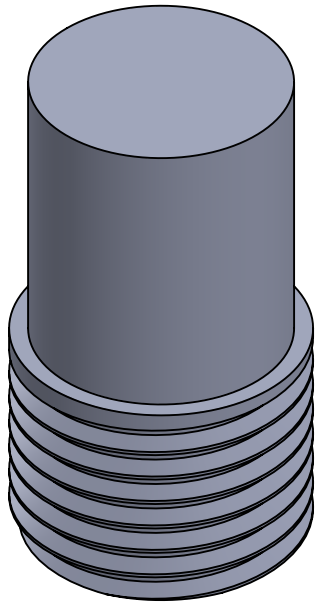

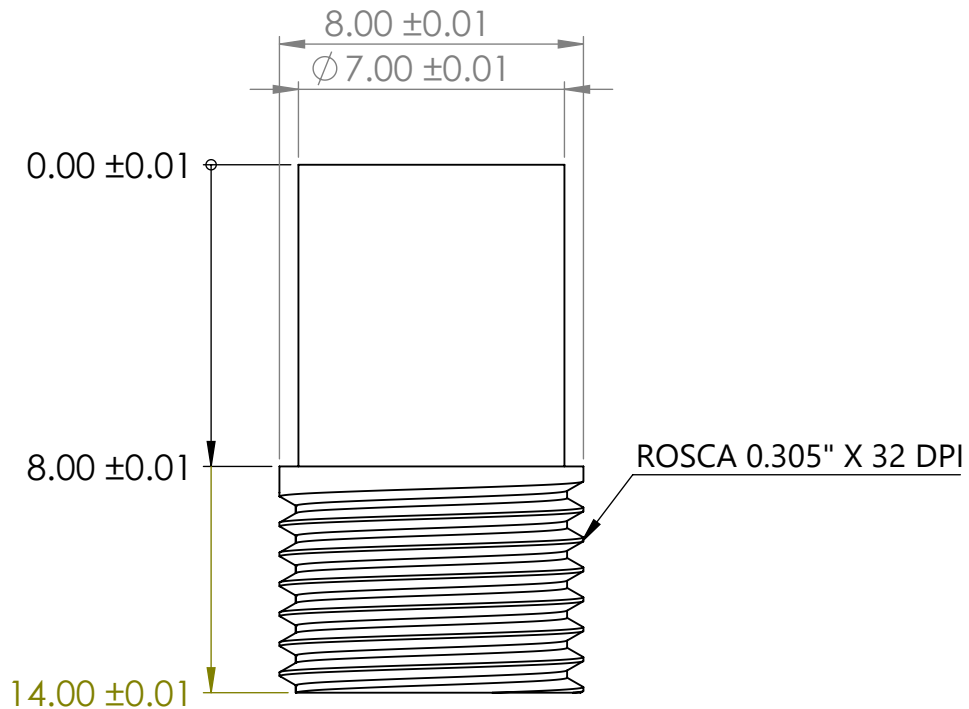

8.00 ±0.01
⌀ 7.00 ±0.01

0.00 ±0.01

8.00 ±0.01

14.00 ±0.01

ROSCA 0.305" X 32 DPI

| Material: **Aluminio 5052** | Este dibujo y cualquier información o material descriptivo expuesto en él son propiedad y con derechos de la Universidad Católica Boliviana San Pablo y NO DEBE SER COPIADO, PRESTADO total o parcialmente o utilizado para cualquier propósito sin el permiso por escrito. | Título **Valvula Schrader** | | |
|---|---|---|---|---|
| Acabado: **Pulido** | | | | |
| A menos que indique lo contrario: Tol. Lineal.: ±**0.01**, Tol. Angular.: **0°05'** Acabado Superficial: **0.8µm** | | Dibujado por: **L.J.A.A.** | Fecha de dibujo **09/09/2020** | |
| Escala de dibujo: **5:1** | **Universidad Católica Boliviana San Pablo** | Verificado / Aprobado por: **D.R.A.** | Fecha de verificación / aprobación: **09/09/2020** | |
| Peso Aprox.: **0.004** Kg | Dibujo producido en de acuerdo con:: **BS8888** | Dueño Legal: | Número de parte **6** | |
| Metodo de proyeccion: **Tercer Angulo** | Tamaño de hoja:: **A4** | | Número de Dibujo: **8** | Hoja: **8 of 8** | Revisión: |

UNIVERSIDAD CATÓLICA BOLIVIANA SANTA CRUZ

**ANEXO 2**

**Hojas de especificaciones tecnicas**

# Integrated Silicon Pressure Sensor

## On-Chip Signal Conditioned, Temperature Compensated and Calibrated

| **MPX5500 Series** |
| :---: |
| 0 to 500 kPa (0 to 72.5 psi) |
| 0.2 to 4.7 V Output |

The MPX5500 series piezoresistive transducer is a state-of-the-art monolithic silicon pressure sensor designed for a wide range of applications, but particularly those employing a microcontroller or microprocessor with A/D inputs. This patented, single element transducer combines advanced micromachining techniques, thin-film metallization, and bipolar processing to provide an accurate, high level analog output signal that is proportional to the applied pressure.

## Features

• 2.5% Maximum Error over 0° to 85°C
• Ideally suited for Microprocessor or Microcontroller-Based Systems
• Patented Silicon Shear Stress Strain Gauge
• Durable Epoxy Unibody Element
• Available in Differential and Gauge Configurations

## Operating Characteristics

Table 1. Operating Characteristics $V_S$ = 5.0 Vdc,

| Characteristic | | Symbol | Min | Typ | Max | Unit |
|---|---|---|---|---|---|---|
| | | | | | | |
| Pressure Range [1] | | $P_{OP}$ | 0 | — | 500 | kPa |
| Supply Voltage [2] | | $V_S$ | 4.75 | 5.0 | 5.25 | Vdc |
| Supply Current | | Io | — | 7.0 | 10 | mAdc |
| Zero Pressure Offset[3] | (0 to 85°C) | $V_{off}$ | 0.088 | 0.20 | 0.313 | Vdc |
| Full Scale Output[4] | (0 to 85°C) | $V_{FSO}$ | 4.587 | 4.70 | 4.813 | Vdc |
| Full Scale Span[5] | (0 to 85°C) | $V_{FSS}$ | — | 4.50 | — | Vdc |
| Accuracy[6] | (0 to 85°C) | — | — | — | ±2.5 | %$V_{FSS}$ |
| Sensitivity | | V/P | — | 9.0 | — | mV/kPa |
| Response Time[7] | | $t_R$ | — | 1.0 | — | ms |
| Output Source Current at Full Scale Output | | Io+ | — | 0.1 | — | mAdc |
| Warm-Up Time[8] | | — | — | 20 | — | ms |

# Maximum Ratings

**Table 2. Maximum Ratings** [1]

| Rating | Symbol | Value | Unit |
|---|---|---|---|
| Maximum Pressure [2] (P2 ≤ 1 Atmosphere) | $P1_{max}$ | 2000 | kPa |
| Storage Temperature | $T_{stg}$ | -40 to +125 | °C |
| Operating Temperature | $T_A$ | -40 to +125 | °C |

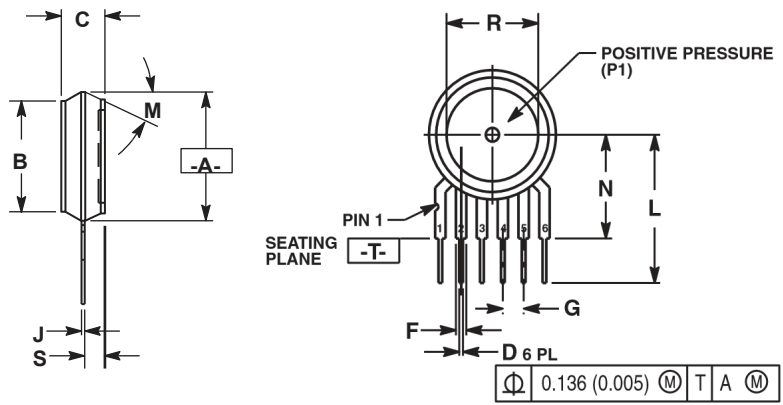

NOTES:
1.
2.
3.

# Hoja de especificaciones técnicas
## cDAQ 9174
## NI CompactDAQ Four-Slot USB Chassis

SPECIFICATIONS

NI cDAQ™-9174

**NI CompactDAQ Four-Slot USB Chassis**

These specifications are for the National Instruments CompactDAQ 9174 chassis only. These specifications are typical at 25 °C unless otherwise noted. For the C Series I/O module specifications, refer to the documentation for the C Series I/O module you are using.

## Analog Input

Input FIFO size .................................................. 127 samples per slot

Maximum sample rate[1] ......................................... Determined by the C Series I/O module or modules

Timing accuracy[2] ............................................. 50 ppm of sample rate

Timing resolution[3] ........................................... 12.5 ns

Number of channels supported .......................... Determined by the C Series I/O module or modules

## Analog Output

Number of channels supported Hardware-timed task

    Onboard regeneration ......................... 16

    Non-regeneration ............................... Determined by the C Series I/O module or modules

Non-hardware-timed task      Determined by the C Series I/O module or modules

Maximum update rate

    Onboard regeneration ................................. 1.6 MS/s (multi-channel, aggregate)

    Non-regeneration ...................................... Determined by the C Series I/O module or modules

Timing accuracy ................................................. 50 ppm of sample rate

Timing resolution ............................................... 12.5 ns

Output FIFO size

Onboard regeneration ................................ 8,191 samples shared among channels used

Non-regeneration ...................................... 127 samples per slot

AO waveform modes........................................ Non-periodic waveform, periodic waveform
regeneration mode from onboard memory, periodic waveform
regeneration from host buffer including dynamic update

**General-Purpose Counters/Timers**

Number of counters/timers................................. 4

Resolution ........................................................ 32 bits

Counter measurements...................................... Edge counting, pulse, semi-period, period,
two-edge separation, pulse width

Position measurements...................................... X1, X2, X4 quadrature encoding with
Channel Z reloading; two-pulse encoding

Output applications ........................................... Pulse, pulse train with dynamic updates,
frequency division, equivalent time sampling Internal base clocks    80
MHz, 20 MHz, 100 kHz

External base clock frequency .......................... 0 to 20 MHz

Base clock accuracy .......................................... 50 ppm

Output frequency ............................................... 0 to 20 MHz

. Inputs ............................................................. Gate, Source, HW_Arm, Aux, A, B, Z,
Up_Down

Routing options for inputs ................................ Any module PFI, analog trigger, many internal
signals

. FIFO ............................................................... Dedicated 127-sample FIFO

**Module I/O States**

At power-on..................................................... Module-dependent. Refer to the documentation
for each C Series I/O module.

**Note** The chassis may revert the input/output of the modules to their power-on state when the USB cable is removed.

**Power Requirements**

**Caution** You must use a National Electric Code (NEC) Class 2 power source with the NI cDAQ-9174 chassis.

**Note** Some C Series I/O modules have additional power requirements. For more information about C Series I/O module power requirements, refer to the documentation for each C Series I/O module.

**Note** Sleep mode for C Series I/O modules is not supported in the NI cDAQ-9174.

............. ............................9 to 30 V
Input voltage range

Maximum required input 15 W
..............................................

Power input connector ...................................... 2 positions 3.5 mm pitch pluggable screw
terminal with screw locks similar to Sauro CTMH020F8-0N001

Power input mating connector ........................... Sauro CTF020V8, Phoenix Contact 1714977,
or equivalent

Power consumption from USB, ........................ 500 μA maximum
4.10 to 5.25 V

DATASHEET

# NI 9219

4 AI, 100 S/s/ch Simultaneous, Universal Measurements

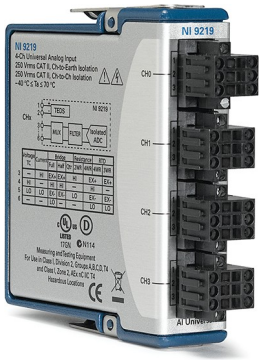

The NI 9219 is a universal C Series module designed for multipurpose testing in any NI CompactDAQ or CompactRIO chassis. With the NI 9219, you can measure several signals from sensors such as strain gages, RTDs, thermocouples, load cells, and other powered sensors. The channels are individually selectable, so you can perform a different measurement type on each of the four channels. Measurement ranges differ for each type of measurement and include up to ±60 V for voltage and ±25 mA for current.

**NI 9219 Circuitry**

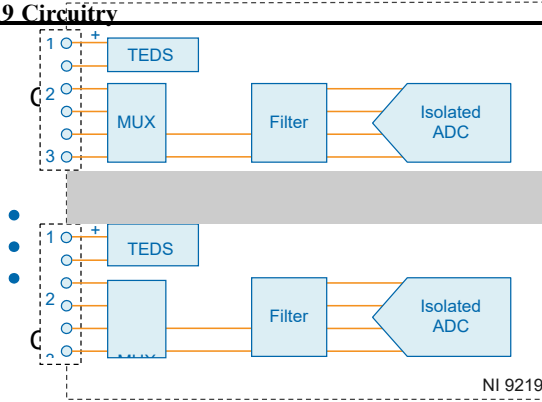

**Voltage Circuitry**

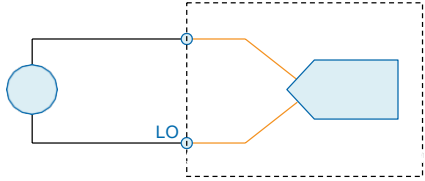

**Full-Bridge Circuitry**

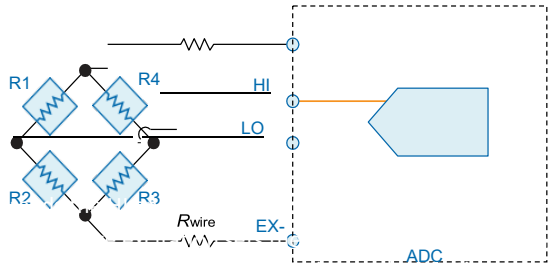

The following specifications are typical for the range -40 °C to 70 °C unless otherwise noted.

**Caution** Do not operate the NI 9219 in a manner not specified in this document. Product misuse can result in a hazard. You can compromise the safety protection built into the product if the product is damaged in any way. If the product is damaged, return it to NI for repair.

**Input Characteristics**

| | |
|---|---|
| Number of channels | 4 analog input channels |
| ADC resolution | 24 bits |
| Type of ADC | Delta-sigma (with analog prefiltering) |
| Sampling mode | Simultaneous |
| Type of TEDS supported | IEEE 1451.4 TEDS Class 2 (Interface) |

**Table 1.** Input Ranges

| Measurement Type | Nominal Range(s) | Actual Range(s) |
|---|---|---|
| Voltage | ±60 V, ±15 V, ±4 V, ±1 V, ±125 mV | ±60 V, ±15 V, ±4 V, ±1 V, ±125 mV |
| Current | ±25 mA | ±25 mA |
| Thermocouple | ±125 mV | ±125 mV |
| 4-Wire and 2-Wire Resistance | 10 kΩ, 1 kΩ | 10.5 kΩ, 1.05 kΩ |
| 4-Wire and 3-Wire RTD | Pt 1000, Pt 100 | 5.05 kΩ, 505 Ω |

| Quarter-Bridge | 350 Ω, 120 Ω | 390 Ω, 150 Ω |
|---|---|---|
| Half-Bridge | ±500 mV/V | ±500 mV/V |
| Full-Bridge | ±62.5 mV/V, ±7.8 mV/V | ±62.5 mV/V, ±7.8125 mV/V |

**Table 1. Input Ranges (Continued)**

| Measurement Type | Nominal Range(s) | Actual Range(s) |
|---|---|---|
| Digital In | — | 0 V to 60 V |
| Open Contact | — | 1.05 kΩ |

**Table 2. Accuracy**

| Measurement Type | Range | Gain Error (Percent of Reading) | Offset Error (ppm of Range) |
|---|---|---|---|
| | | Typical (25 °C ±5 °C), Maximum (-40 °C to 70 °C) | |
| Voltage | ±60 V | ±0.3, ±0.4 | ±20, ±50 |
| | ±15 V | ±0.3, ±0.4 | ±60, ±180 |
| | ±4 V | ±0.3, ±0.4 | ±240, ±720 |
| | ±1 V | ±0.1, ±0.18 | ±15, ±45 |
| Voltage/Thermocouple | ±125 mV | ±0.1, ±0.18 | ±120, ±360 |
| Current | ±25 mA | ±0.1, ±0.6 | ±30, ±100 |
| 4-Wire and 2-Wire[1] Resistance | 10 kΩ | ±0.1, ±0.5 | ±120, ±320 |
| | 1 kΩ | ±0.1, ±0.5 | ±1200, ±3200 |
| 4-Wire and 3-Wire RTD | Pt 1000 | ±0.1, ±0.5 | ±240, ±640 |
| | Pt 100 | ±0.1, ±0.5 | ±2400, ±6400 |
| Quarter-Bridge | 350 Ω | ±0.1, ±0.5 | ±2400, ±6400 |
| | 120 Ω | ±0.1, ±0.5 | ±2400, ±6400 |
| Half-Bridge | ±500 mV/V | ±0.03, ±0.07 | ±300, ±450 |
| Full-Bridge | ±62.5 mV/V | ±0.03, ±0.08 | ±300, ±1000 |
| | ±7.8 mV/V | ±0.03, ±0.08 | ±2200, ±8000 |

**USER GUIDE AND SPECIFICATIONS**

# NI myRIO-1900

The National Instruments myRIO-1900 is a portable reconfigurable I/O (RIO) device that students can use to design control, robotics, and mechatronics systems. This document contains pinouts, connectivity information, dimensions, mounting instructions, and specifications for the NI myRIO-1900.

**Figure 1.** NI myRIO-1900

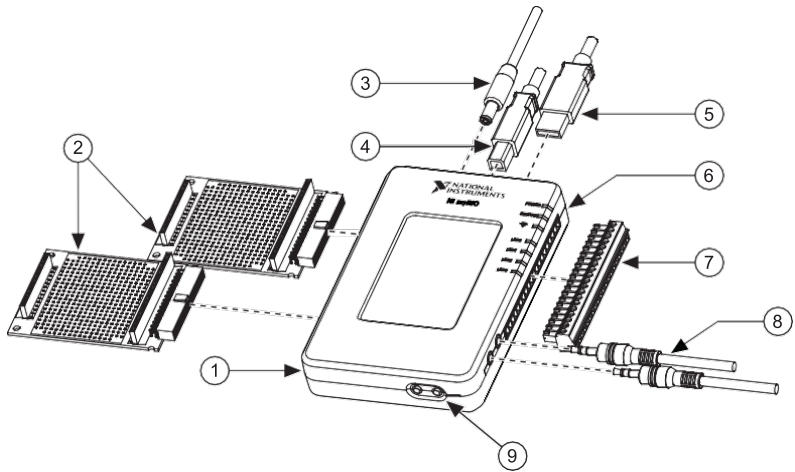

1   NI myRIO-1900

**Specifications**
The following specifications are typical for the 0 to 40 °C operating tempreature range unless otherwise noted.
**Processor**

Processor type......................................................... linx Z-7010

Processor speed................................................ 667 MHz

Processor cores ................................................ 2
**Memory**

Nonvolatile memory ........................................ 512 MB

DDR3 memory................................................ 256 MB

DDR3 clock frequency............................ 533 MHz

DDR3 data bus width.............................. 16 bits

**Wireless Characteristics**

Radio mode ....................................................... IEEE 802.11 b,g,n

Frequency band ................................................. ISM 2.4 GHz

Channel width ................................................... 20 MHz

Channels........................................................... USA 1 to 11, International 1 to 13

TX power ....................................................... +10 dBm max (10 mW)

Outdoor range .................................................. Up to 150 m (line of sight)

Antenna directivity........................................... Omnidirectional

Security ......................................................... WPA, WPA2, WPA2-Enterprise

**USB Ports**

USB host port.................................................. USB 2.0 Hi-Speed

USB device port .............................................. USB 2.0 Hi-Speed

**Analog Input**

Aggregate sample rate ...................................... 500 kS/s

Resolution ...................................................... 12 bits

Overvoltage protection ..................................... ±16 V

MXP connectors

Configuration ........................................... Four single-ended channels per connector

Input impedance ....................................... >500 kΩ acquiring at 500 kS/s

1 MΩ powered on and idle

4.7 kΩ powered off Recommended source

impedance ............................................... 3 kΩ or less

Nominal range........................................... V to +5 V

Absolute accuracy ..................................... ±50 mV

Bandwidth .............................................. >300 kHz

MSP connector

Configuration ........................................... Two differential channels

Input impedance ....................................... Up to 100 nA leakage powered on;
4.7 kΩ powered off Nominal range    ±10 V

Working voltage

(signal + common mode)........................... ±10 V of AGND

Absolute accuracy ................................... ±200 mV

Bandwidth ............................................... 20 kHz minimum, >50 kHz typical

Audio input

Configuration ........................................... One stereo input consisting of two AC-coupled,
single-ended channels Input impedance

............................................................... 10 kΩ at DC

Nominal range........................................... ±2.5 V

Bandwidth ............................................... 2 Hz to >20 kHz

**Analog Output**

Aggregate maximum update rates

All AO channels on MXP connectors 345 kS/s

All AO channels on MSP connector

and audio output channels ........................ 345 kS/s

Resolution ........................................................ 12 bits

Overload protection .......................................... ±16 V

Startup voltage .................................................. 0 V after FPGA initialization

MXP connectors

Configuration .......................................... Two single-ended channels per connector

Range ...................................................... V to +5 V

Absolute accuracy ......................................... mV

Current drive ........................................... 3 mA

Slew rate .................................................. 0.3 V/µs

MSP connector

Configuration .......................................... Two single-ended channels

Range ..................................................... ±10 V

Absolute accuracy .................................... ±200 mV

Current drive ........................................... 2 mA

Slew rate ................................................. 2 V/µs

Audio output

Configuration .......................................... One stereo output consisting of
two AC-coupled, single-ended channels Output impedance 100 Ω

in series with 22 µF

Bandwidth .............................................. 70 Hz to >50 kHz into 32 Ω load;
2 Hz to >50 kHz into high-impedance load

**Digital I/O**

Number of lines

MXP connectors            2 ports of 16 DIO lines (one port per connector);

one UART.RX and one UART.TX line per connector

MSP connector ........................................ 1 port of 8 DIO lines

Direction control ............................................... Each DIO line individually programmable as
input or output

Logic level ........................................................ 5 V compatible LVTTL input; 3.3 V LVTTL
output

Input logic levels

Input low voltage, $V_{IL}$ .................................. V min; 0.8 V max

Input high voltage, $V_{IH}$ .................................. V min; 5.25 V max

Output logic levels

Output high voltage, $V_{OH}$

sourcing 4 mA        V min; 3.465 V max

Output low voltage, $V_{OL}$

sinking 4 mA .............................................. V min; 0.4 V max

Minimum pulse width ......................................20 ns

Maximum frequencies for secondary digital functions SPI        4 MHz

PWM ........................................................ 100 kHz

Quadrature encoder input ......................... 100 kHz

I²C ......................................................... 400 kHz

### UART lines

Maximum baud rate.................................. 230,400 bps

Data bits .................................................. 5, 6, 7, 8

Stop bits .................................................. 1, 2

Parity ...................................................... Odd, Even, Mark, Space

Flow control ............................................ XON/XOFF

**Accelerometer**

Number of axes ....................................... 3

Range ...................................................... ±8 g

Resolution ............................................... 12 bits

Sample rate.............................................. 800 S/s

Noise ....................................................... 3.9 mg$_{rms}$ typical at 25 °C

**Power Output**

### +5 V power output

Output voltage................................................. V to 5.25 V

Maximum current on each connector ....... 100 mA

### +3.3 V power output

Output voltage................................................. V to 3.6 V

Maximum current on each connector ....... 150 mA

### +15 power output

Output voltage..........................................+15 V to +16 V

Maximum current .................................... 32 mA (16 mA during startup)

- ### 15 V power output

Output voltage..........................................-15 V to -16 V

Maximum current .................................... 32 mA (16 mA during startup)

### Maximum combined power from +15 V

### and -15 V power output     500 mW

**Power Requirements**

NI myRIO-1900 requires a power supply connected to the power connector. Power supply voltage

range................................................................. 6 to 16 VDC

Maximum power consumption ......................... 14 W

Typical idle power consumption ...................... 2.6 W

## LOW CAPACITY SINGLE POINT LOAD CELL

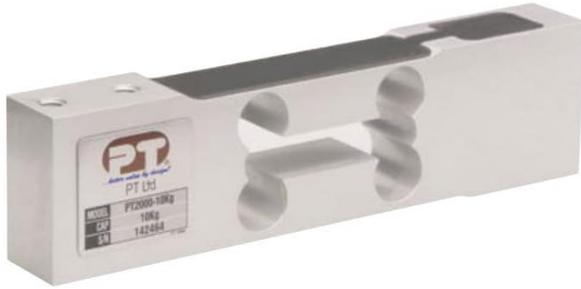

*Our most popular single point model, used in high numbers in shop counter retail and bench scales.*

*The PT2000 is available in 9 capacities from 5kg to 100kg. Industry standard design allows interchange with many manufacturers models.*

*A popular cell for retail scale manufacture because of its generous platform size of 400mm x 400mm, also used extensively in belt conveyer and check weighing applications.*

*Protected with SURESEAL to IP66 the PT2000 is standard with overload protection facility built-in.*

*Marine grade anodised and designed to have low sensitivity to off-center loading it's a perfect load cell selection for the volume scale manufacturer.*

## APPLICATIONS

- Low cost retail scales
- Low cost bench scales
- Hopper scales & net weighing scales

## FEATURES

- Marine grade anodised
- Protected with **SURESEAL**™
- Industry standard interchange design
- Generous platform sizes up to 400mm x 400mm
- Overload protection facility

## Specifications

*Note: All specifications are a maximum, as a % (±) of full load, unless otherwise stated.*

| | | | |
|---|---|---|---|
| Nominal Capacity | 5kg ~ 100kg | Ultimate Load | 300% of Rated Capacity |
| Signal Output at Capacity | 2mV/V ± 10% | Input Impedance | 425Ω ± 15Ω |
| Linearity Error | < 0.020% FSO | Output Impedance | 350Ω ± 3Ω |
| Non-Repeatability | < 0.010% FSO | Insulation Impedance | > 5000 MΩ at 100V DC |
| Combined Error | < 0.025% FSO | Excitation Voltage (Recommended) | 5 ~ 12V AC/DC |
| Hysteresis | < 0.015% FSO | Excitation Voltage (Maximum) | 15V AC/DC |
| Creep/Zero Return (30 mins) | < 0.030% / 0.020% FSO | Eccentric Loading (effect/cm) | < 0.0074% FSO |
| Zero Balance | < 3.000% Capacity | Deflection at Rated Capacity | < 0.4mm |
| Temperature Effect on Span/10°C | < 0.010% FSO | Storage Temperature Range | -50 ~ 70°C |
| Temperature Effect on Zero/10°C | < 0.015% Capacity | Cable Type | 4mm, Screened, PVC Sheath |
| Compensated Temperature Range | -10 ~ 40°C | | 4 Core x 0.09mm² (28 AWG) |
| Operating Temperature Range | -30 ~ 70°C | Cable Length | 2 Metres |
| Service Load | 100% of Rated Capacity | Material | Aluminium |
| Safe Load | 150% of Rated Capacity | Finish | Marine Anodised |